\documentclass{article}

\usepackage{microtype}
\usepackage{graphicx}
\usepackage[frozencache=true,cachedir=minted-cache]{minted}
\usepackage{wrapfig}
\usepackage{bm}
\usepackage{booktabs} 

\usepackage{xcolor}
\usepackage{hyperref}

\newcommand{\cmt}[1]{}

\usepackage[accepted]{icml2023}

\usepackage{amsmath}
\usepackage{amssymb}
\usepackage{mathtools}
\usepackage{amsthm}

\usepackage[capitalize,noabbrev]{cleveref}

\usepackage{enumerate}
\usepackage{tikz}
\usepackage{subcaption}
\usepackage{xcolor}
\hypersetup{
    colorlinks=true,
    urlcolor=magenta,
}

\DeclareMathOperator*{\argmin}{arg\,min}

\usetikzlibrary{bayesnet, calc, arrows.meta, bending}
\usetikzlibrary{decorations.pathreplacing,angles,quotes,shapes.multipart, calligraphy}

\makeatletter

\newcommand{\Rmnum}[1]{\expandafter\@slowromancap\romannumeral #1@}

\makeatother

\theoremstyle{plain}

\theoremstyle{definition}

\theoremstyle{remark}

\usepackage[textsize=tiny]{todonotes}

\icmltitlerunning{High Fidelity Image Counterfactuals with Probabilistic Causal Models}

\begin{document}

\twocolumn[
\icmltitle{High Fidelity Image Counterfactuals with Probabilistic Causal Models}



\icmlsetsymbol{equal}{*}

\begin{icmlauthorlist}
\icmlauthor{Fabio De Sousa Ribeiro}{yyy}
\icmlauthor{Tian Xia}{yyy}
\icmlauthor{Miguel Monteiro}{yyy}
\icmlauthor{Nick Pawlowski}{comp}
\icmlauthor{Ben Glocker}{yyy}
\end{icmlauthorlist}

\icmlaffiliation{yyy}{Imperial College London}
\icmlaffiliation{comp}{Microsoft Research Cambridge, UK}
\icmlcorrespondingauthor{Fabio De Sousa Ribeiro}{f.de-sousa-ribeiro@imperial.ac.uk}

\icmlkeywords{deep generative models, causality, causal inference, counterfactual inference, causal effect estimation, variational inference, variational autoencoder, medical imaging}

\vskip 0.3in
]



\printAffiliationsAndNotice{}  

\begin{abstract}
We present a general causal generative modelling framework for accurate estimation of high fidelity image counterfactuals with deep structural causal models. Estimation of interventional and counterfactual queries for high-dimensional structured variables, such as images, remains a challenging task. We leverage ideas from \textit{causal mediation} analysis and advances in generative modelling to design new deep causal mechanisms for structured variables in causal models. Our experiments demonstrate that our proposed mechanisms are capable of accurate \textit{abduction} and estimation of \textit{direct}, \textit{indirect} and \textit{total effects} as measured by axiomatic \textit{soundness} of counterfactuals. 
\end{abstract}

\section{Introduction}
\label{sec:introduction}
Many real-world challenges still prevent the adoption of Deep Learning (DL) systems in safety-critical settings~\cite{d2022underspecification}. 
It has been argued that such obstacles arise partly from a purely statistical treatment of predictive modelling, wherein notions of \textit{causality} are not taken into account~\cite{pearl2009causality,Bengio2013,kusner2017counterfactual,peters2017elements}. Consequently, research on causality and representation learning has garnered significant interest~\cite{scholkopf2021toward,scholkopf2022causality}. 

Scientific inquiry is invariably motivated by causal questions: ``\textit{how effective is $X$ in preventing $Y$?}", or ``\textit{what would have happened to $Y$ had $X$ been $x$?}". Such questions cannot be answered using statistical tools alone~\cite{pearl2009causality}. As such, a mathematical framework is required to precisely express and answer such questions using observed data. A \textit{causal model} represents our assumptions about how nature assigns values to variables of interest in a system. The relationships between variables in a causal model are directed from cause to effect, and intervening on a cause ought to change the effect and not the other way around. The goal is to leverage causal models to estimate the causal effect of actions, even in hypothetical (counterfactual) scenarios. 

The ability to generate plausible counterfactuals has wide scientific applicability and is particularly valuable in fields like medical imaging, wherein data are scarce and underrepresentation of subgroups is prevalent~\cite{pawlowski2020deep,castro2020causality,seyyed2020chexclusion,glocker2023algorithmic}. Suppose we are granted access to medical imaging data alongside reliable meta-data of the respective patients, e.g. annotations of their protected attributes. In such cases, if we can make sensible medically-informed causal assumptions about the underlying data generating process, we may be able to construct a causal model which better reflects reality. Furthermore, we argue that the ability to answer counterfactual queries like "\textit{why?}'' and "\textit{what if..?}" expressed in the language of causality could greatly benefit several other important areas: (i) explainability~\cite{wachter2017counterfactual,mothilal2020explaining}, e.g. through causal mediation effects as studied herein; (ii) data augmentation, e.g. mitigating data scarcity and underrepresentation of subgroups~\cite{Kaushik2020Learning,xia2022adversarial}; (iii) robustness, to e.g. spurious correlations~\cite{simon1954spurious,balashankar2021can}, and (iv) fairness notions in both observed and counterfactual outcomes~\cite{kusner2017counterfactual,zhang2018fairness}. Despite recent progress, accurate estimation of interventional and counterfactual queries for high-dimensional structured variables (e.g. images) remains an open problem~\cite{pawlowski2020deep,yang2021causalvae,schut2021generating,sanchez2021diffusion}.

Our research bolsters an ongoing effort to combine causality and deep representation learning~\cite{Bengio2013,scholkopf2021toward}. However, few previous works have attempted to fulfil all three rungs of Pearl’s \textit{ladder of causation}~\cite{pearl2009causality}: association ($\mathcal{L}_1$); intervention ($\mathcal{L}_2$) and counterfactuals ($\mathcal{L}_3$) in a principled manner using deep models. Notable exceptions include Deep Structural Causal Models (DSCMs)~\cite{pawlowski2020deep} and Neural Causal Models (NCMs)~\cite{xia2021causal,xia2023neural}, both of which our research builds upon. Contrary to preceding studies, our main focus is on exploring the practical limits and possibilities of estimating and empirically evaluating high-fidelity image counterfactuals of real-world data. For this purpose, we introduce a specific system and method. 

Our main contributions can be summarised as follows:
\begin{enumerate}[{(i)}]
    \item We present a general causal generative modelling framework for producing high-fidelity image counterfactuals with Markovian probabilistic causal models;
    \item Inspired by causal mediation analysis, our proposed deep causal mechanisms can plausibly estimate direct, indirect, and total treatment effects on high-dimensional structured variables (i.e. images);
    \item We demonstrate the \textit{soundness} of our counterfactuals by evaluating axiomatic properties that must hold true in all causal models: \textit{effectiveness} and \textit{composition}. 
\end{enumerate}
\section{Background}
\label{background}
\subsection{Structural Causal Models}
\label{scms}
A Structural Causal Model (SCM)~\cite{pearl2009causality,peters2017elements} is a triple $\mathcal{M} \coloneqq \langle X, U, F \rangle$ consisting of two sets of variables, $X = \{x_1,\dots,x_N\}$ and $U = \{u_1,\dots,u_N\}$, and a set of functions $F = \{f_1,\dots,f_N\}$. The value of each variable $x_k \in X$ is a function of its direct cause(s) $\mathbf{pa}_k \subseteq X \setminus \{x_k\}$, and an exogenous noise variable $u_k \in U$:
\begin{align}
&x_k \coloneqq f_k(\mathbf{pa}_k, u_k), &k=1,\dots,N.    
\end{align}
The variables in $X$ are called \textit{endogenous} since they are caused by the variables in the model $X \cup U$, whereas variables in $U$ are \textit{exogenous} as they are caused by factors which are external to the model. The functions in $F$ are known as \textit{structural assignments} or \textit{causal mechanisms}. A \textit{causal world} is a pair $\langle \mathcal{M}, \mathbf{u} \rangle$ where $\mathbf{u}$ is a realization of the exogenous variables $U$, and a \textit{probabilistic causal model} $\langle \mathcal{M}, P(U) \rangle$ is a distribution over causal worlds.

\paragraph{Observational Distribution.}  If the structural assignments are acyclic, the SCM can be represented by a Directed Acyclic Graph (DAG) with edges pointing from causes to effects. If the exogenous variables are jointly independent $P(U) = \prod_{k=1}^N P(u_k)$, the model is called \textit{Markovian}. 
Every Markovian causal model induces a unique joint observational distribution over the endogenous variables: $P_{\mathcal{M}}(X) = P_{\mathcal{M}}(x_1,\dots,x_N) = \prod_{k=1}^N P(x_k \mid \mathbf{pa}_k)$, satisfying the causal Markov condition; that each variable is independent of its nondescendants given its direct causes. 

\paragraph{Interventional Distribution.} SCMs can predict the causal effects of actions by performing \textit{interventions} on the endogenous variables. Interventions answer questions like ``\textit{what would $x_k$ be if $X \setminus \{x_k\}$ had been fixed to certain values?}". An intervention is the action of replacing one or several of the structural assignments using the \textit{do}-operator. A \textit{hard} intervention replaces $f_k$ by setting $x_k$ to some constant $c$, denoted as \textit{do}$(x_k \coloneqq c)$ or \textit{do}$(c)$. A \textit{soft} intervention is more general and can consist of replacing $f_k$ by some new mechanism, e.g. \textit{do}$(x_k \coloneqq \widetilde{f}_k(\mathbf{pa}_k, \widetilde{u}_k))$~\cite{peters2017elements}. Intervening on an SCM $\mathcal{M}$ by \textit{do}$(c)$ induces a submodel $\mathcal{M}_{c}$. The entailed distribution of $\mathcal{M}_c$ is called an \textit{interventional distribution} $P_{\mathcal{M}_{c}}(X \mid \textit{do}(c))$, and it is generally different from the observational distribution entailed by $\mathcal{M}$.

\paragraph{Counterfactuals.} SCMs further enable us to consider hypothetical scenarios and answer counterfactual questions like: ``\textit{given that we observed $X$, what would $x_k$ have been had $X \setminus \{x_k\}$ been fixed to certain values?}". Counterfactuals are the result of interventions in the context of a particular observation of $X$. Computing counterfactuals involves the following three-step procedure~\cite{pearl2009causality}:
\begin{enumerate}[(i)]
    \item \textbf{Abduction:} Update $P(U)$ given observed evidence, i.e infer the posterior noise distribution $P(U \mid X)$.
    \item \textbf{Action:} Perform an intervention, e.g. \textit{do}$(\widetilde{x}_k \coloneqq c)$, to obtain the modified submodel $\mathcal{M}_{c}$.
    \item \textbf{Prediction:} Use the model $\langle \mathcal{M}_{c}$, $P(U \mid X) \rangle$ to compute the probability of a counterfactual.
\end{enumerate}
\subsection{Hierarchical Latent Variable Models}
A Hierarchical Latent Variable Model (HLVM) defines a generative model for data $\mathbf{x}$ using a prior over $L$ layers of hierarchical latent variables $\{\mathbf{z}_i\}_{i=1}^L$, factorizing as:
\begin{align}
    p(\mathbf{x}, \mathbf{z}_{1:L}) = p(\mathbf{x} \mid \mathbf{z}_{1:L}) p(\mathbf{z}_L)\prod_{i=1}^{L-1} p(\mathbf{z}_i \mid \mathbf{z}_{>i}).
\end{align}
Hierarchical Variational Autoencoders (HVAEs)~\cite{kingma2016improved,sonderby2016ladder,burda2015importance} extend standard VAEs~\cite{kingma2013auto,rezende2014stochastic} to $L>1$. HVAEs train a hierarchical generative model $p_\theta(\mathbf{x}, \mathbf{z}_{1:L})$, by introducing a variational inference model $q_\phi(\mathbf{z}_{1:L} \mid \mathbf{x})$ and maximizing the Evidence Lower Bound (ELBO) on the marginal log-likelihood of the data:
\begin{align}
    \log p_\theta & (\mathbf{x})  \geq 
    \mathbb{E}_{q_\phi(\mathbf{z}_{1:L}|\mathbf{x})} [\log p_\theta(\mathbf{x} \mid \mathbf{z}_{1:L})] \\& \nonumber - D_{\mathrm{KL}}(q_\phi(\mathbf{z}_{1:L} \mid \mathbf{x}) \parallel p_\theta(\mathbf{z}_{1:L})) = \mathcal{L}(\theta, \phi;\mathbf{x}).
\end{align}
The goal is to optimize the ELBO via the trainable parameters $\theta$ and $\phi$ such that the marginal $p_\theta(\mathbf{x})$ is close to a given data distribution $p_\mathrm{data}(\mathbf{x})$. \citet{sonderby2016ladder} 
proposed the Ladder VAE, featuring a \textit{top-down} inference model:
\begin{align}
q_\phi(\mathbf{z}_{1:L} \mid \mathbf{x}) = q_\phi(\mathbf{z}_L \mid \mathbf{x})\prod_{i=1}^{L-1} q_\phi(\mathbf{z}_{i} \mid \mathbf{z}_{>i}, \mathbf{x}),
\end{align}
which infers the latent variables in the same top-down order as the generative model, rather than in the standard reverse generative order (\textit{bottom-up} inference). 
More recently, this top-down inference structure has featured in much deeper state-of-the-art HVAEs~\cite{maaloe2019biva,vahdat2020nvae,child2020very,shu2022bit}.
\begin{figure}[!t]
    \centering
    \hfill
    \begin{subfigure}{.39\columnwidth}
        \centering       
        \begin{tikzpicture}[thick,scale=0.75, every node/.style={scale=0.75}]
            \node[obs] (x) {$\mathbf{x}$};
            \node[det, above=20pt of x] (d1) {$\mathbf{d}_1$};
            \node[det, above=20pt of d1] (d2) {$\mathbf{d}_2$};
            \node[det, above=20pt of d2] (d3) {$\mathbf{d}_3$};
            \node[latent, right=25pt of d1] (z1) {$\mathbf{z}_1$};
            \node[latent, right=25pt of d2] (z2) {$\mathbf{z}_2$};
            \node[latent, right=25pt of d3] (z3) {$\mathbf{z}_3$};
            \node[det, right=15pt of z2] (h2) {$\mathbf{h}_2$};
            \node[det, right=15pt of z3] (h3) {$\mathbf{h}_3$};
    
            \node[above=20pt of h3] (hl) {$\mathbf{h}_\mathrm{init}$};
            
            \edge[-{Latex[scale=1.0]}]{x}{d1}
            \edge[-{Latex[scale=1.0]}]{d1}{d2} 
            \edge[-{Latex[scale=1.0]}]{d2}{d3} 
            \edge[-{Latex[scale=1.0]}]{z2}{h2}
            \edge[-{Latex[scale=1.0]}]{z3}{h3}
            \edge[-{Latex[scale=1.0]}, dotted]{h2}{z1}
            \edge[-{Latex[scale=1.0]}, dotted]{h3}{z2}
            \edge[-{Latex[scale=1.0]}, red]{h3}{h2}
            
            \draw[-{Latex[scale=1.0]}, dotted] (d1) -- (z1) coordinate[midway] (d1z1);
            \draw[-{Latex[scale=1.0]}, dotted] (d2) -- (z2) coordinate[midway] (d2z2);
            \node[obs, above=11pt of d1z1] (pa1) {$\mathbf{pa}_\mathbf{x}$};
            \draw[-{Circle[scale=.8]}] (pa1) edge[out=-90, in=90] (d1z1);
    
            \draw[-{Latex[scale=1.0]}, dotted] (d3) -- (z3) coordinate[midway] (d3z3);
            \node[obs, above=11pt of d2z2] (pa2) {$\mathbf{pa}_\mathbf{x}$};
            \draw[-{Circle[scale=.8]}] (pa2) edge[out=-90, in=90] (d2z2);
            \draw[-{Circle[scale=.8]}] (pa2) edge[out=90, in=-90] (d3z3);
            
            \edge[-{Latex[scale=1.0]}, dotted]{hl}{z3}
            \edge[-{Latex[scale=1.0]}, red]{hl}{h3}
        \end{tikzpicture}
        \caption{Inference}
        \label{inf_model}
    \end{subfigure}
    \hfill
    \begin{subfigure}{.29\columnwidth}
        \centering
        \begin{tikzpicture}[thick,scale=0.75, every node/.style={scale=0.75}]
            \node[] (hl) {$\mathbf{h}_\mathrm{init}$};
            \node[det, below=20pt of hl] (h3) {$\mathbf{h}_3$};
            \node[det, below=20pt of h3] (h2) {$\mathbf{h}_2$};
            \node[det, below=20pt of h2] (h1) {$\mathbf{h}_1$};
            \node[obs, below=20pt of h1] (x) {$\mathbf{x}$};
            
            \node[latent, right=25pt of h1] (z1) {$\mathbf{z}_1$};
            \node[latent, right=25pt of h2] (z2) {$\mathbf{z}_2$};
            \node[latent, right=25pt of h3] (z3) {$\mathbf{z}_3$};
    
            \edge[-{Latex[scale=1.0]}]{h1}{x}
            \edge[-{Latex[scale=1.0]}, red]{h2}{h1}
            \edge[-{Latex[scale=1.0]}, red]{h3}{h2}
            \edge[-{Latex[scale=1.0]}, red]{hl}{h3}
            \edge[-{Latex[scale=1.0]}]{hl}{z3}
    
            \draw[-{Latex[scale=1.0]}] (z1) -- (h1) coordinate[midway] (z1h1);
            \draw[-{Latex[scale=1.0]}] (z2) -- (h2) coordinate[midway] (z2h2);
            \draw[-{Latex[scale=1.0]}] (z3) -- (h3) coordinate[midway] (z3h3);
    
            \edge[-{Latex[scale=1.0]}]{z3}{z2}
            \edge[-{Latex[scale=1.0]}]{z2}{z1}
            
            \node[obs, below=11pt of z3h3] (pa2) {$\mathbf{pa}_\mathbf{x}$};
            \draw[-{Circle[scale=.8]}] (pa2) edge[out=90, in=-90] (z3h3);
            \draw[-{Circle[scale=.8]}] (pa2) edge[out=-90, in=90] (z2h2);
            \node[obs, below=11pt of z2h2] (pa2) {$\mathbf{pa}_\mathbf{x}$};
            \draw[-{Circle[scale=.8]}] (pa2) edge[out=270, in=90] (z1h1);
        \end{tikzpicture}
    \caption{Generative \#1}
    \end{subfigure}
    \hfill
    \begin{subfigure}{.29\columnwidth}
        \centering
        \begin{tikzpicture}[thick,scale=0.75, every node/.style={scale=0.75}]
        \node[] (hl) {$\mathbf{h}_\mathrm{init}$};
        \node[det, below=20pt of hl] (h3) {$\mathbf{h}_3$};
        \node[det, below=20pt of h3] (h2) {$\mathbf{h}_2$};
        \node[det, below=20pt of h2] (h1) {$\mathbf{h}_1$};
        \node[obs, below=20pt of h1] (x) {$\mathbf{x}$};
        
        \node[latent, right=25pt of h1] (z1) {$\mathbf{z}_1$};
        \node[latent, right=25pt of h2] (z2) {$\mathbf{z}_2$};
        \node[latent, right=25pt of h3] (z3) {$\mathbf{z}_3$};
    
        \edge[-{Latex[scale=1.0]}]{h1}{x}
        \edge[-{Latex[scale=1.0]}, red]{h2}{h1}
        \edge[-{Latex[scale=1.0]}, red]{h3}{h2}
        \edge[-{Latex[scale=1.0]}, red]{hl}{h3}
    

        \draw[-{Latex[scale=1.0]}] (z3) -- (h3) coordinate[midway, xshift=0pt, yshift=0pt] (z3h3);
        \draw[-{Latex[scale=1.0]}] (z2) -- (h2) coordinate[midway, xshift=0pt, yshift=0pt] (z2h2);
        \draw[-{Latex[scale=1.0]}] (z1) -- (h1) coordinate[midway, xshift=0pt, yshift=0pt] (z1h1);
        
        \edge[-{Latex[scale=1.0]}]{h3}{z2}
        \edge[-{Latex[scale=1.0]}]{h2}{z1}
    
        \draw[-{Latex[scale=1.0]}] (hl) -- (z3) coordinate[midway, xshift=-3pt, yshift=3pt] (hlz3);
        \node[obs, above=3.5pt of z3, xshift=0pt] (paz3) {$\mathbf{pa}_\mathbf{x}$};
        \draw[-{Circle[scale=.8]}] (paz3) -- (hlz3);
        \draw[-{Circle[scale=.8]}] (paz3) edge[out=185, in=90] (z3h3);
        
        \draw[dashed] (h3) -- (z2) coordinate[midway] (h3z2);
        \node[obs, above=3.5pt of z2, xshift=0pt] (paz2) {$\mathbf{pa}_\mathbf{x}$};
        \draw[-{Circle[scale=.8]}] (paz2) -- (h3z2);
        \draw[-{Circle[scale=.8]}] (paz2) edge[out=185, in=90] (z2h2);
    
        \draw[dashed] (h2) -- (z1) coordinate[midway] (h2z1);
        \node[obs, above=3.5pt of z1] (paz1) {$\mathbf{pa}_\mathbf{x}$};
        \draw[-{Circle[scale=.8]}] (paz1) -- (h2z1);
        \draw[-{Circle[scale=.8]}] (paz1) edge[out=185, in=90] (z1h1);
        \end{tikzpicture}
        \caption{Generative \#2}
        \label{chvae_c}
    \end{subfigure}
    \hfill
    \caption{Proposed conditional HVAE (3 layer example). (a) Top-down inference model. (b) Proposed conditional generative model with an \textit{exogenous} prior. (c) Alternative to (b) with a conditional prior. Red arrows denote shared layers between the inference (a) and generative (b-c) models. Circles are stochastic nodes, diamonds are deterministic, and shaded are observed. Black circles denote conditioning. 
    }
    \label{chvae}
\end{figure}
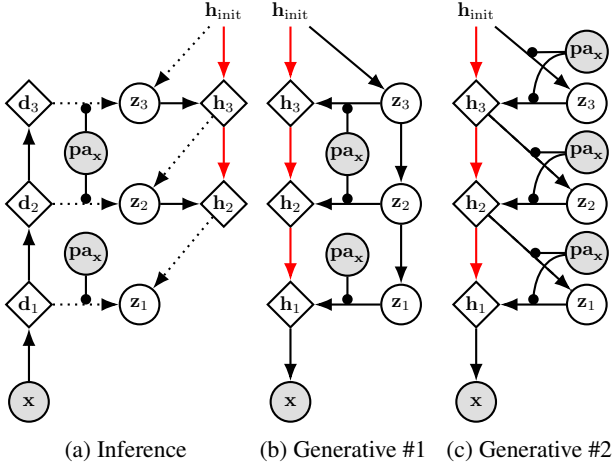
\section{Methods}
In general, we assume a probabilistic Markovian SCM of data $\mathcal{M} \coloneqq \langle X, U, F, P(U)\rangle$, in which an endogenous high-dimensional structured variable $\mathbf{x} \in X$ (e.g. an image) is caused by $K$ lower dimensional endogenous parent variables $\mathbf{pa}_\mathbf{x} \subseteq X \setminus \{\mathbf{x}\}$ (e.g. attributes). Ancestors of $\mathbf{x}$: $\mathbf{an}_{\mathbf{x}} \supseteq \mathbf{pa}_\mathbf{x}$, are \textit{not} assumed to be independent, so we learn their mechanisms from observed data~(further details in Appendix~\ref{appendix: Invertible Mechanisms for Attributes}). The set of mechanisms in $F$ are learned using deep learning components inspired by DSCMs~\cite{pawlowski2020deep} and NCMs~\cite{xia2021causal,xia2023neural}.
\subsection{Deep Mechanisms for Structured Variables}
\label{subsec:Deep Mechanisms for Structured Variables}
The goal is to learn a mechanism for a high-dimensional structured variable, $\mathbf{x} \coloneqq f_\theta(\mathbf{pa}_\mathbf{x}, \mathbf{u}_\mathbf{x})$, which we can invert to abduct the exogenous noise: $\mathbf{u}_\mathbf{x} = f^{-1}_\theta(\mathbf{pa}_\mathbf{x}, \mathbf{x})$. 
\citet{pawlowski2020deep} proposed a VAE setup in which the mechanism for $\mathbf{x}$ is separated into an invertible $h(\cdot)$ and a non-invertible component $g_\theta(\cdot)$ (decoder): $\mathbf{x} \coloneqq f_\theta(\mathbf{pa}_\mathbf{x}, \mathbf{u}_\mathbf{x}) = h(\boldsymbol{\epsilon}; g_\theta(\mathbf{z},\mathbf{pa}_{\mathbf{x}}))$, representing a factored exogenous noise decomposition: $p(\mathbf{u}_{\mathbf{x}}) = p_\theta(\mathbf{z})p(\boldsymbol{\epsilon})$.
The invertible mechanism $h(\cdot)$ is a reparameterization of $g_\theta(\cdot)$'s output mean and variance: $\mathbf{x} = \mu(\mathbf{z}, \mathbf{pa}_\mathbf{x}) + \sigma(\mathbf{z}, \mathbf{pa}_\mathbf{x}) \odot \boldsymbol{\epsilon}$, $\boldsymbol{\epsilon} \sim \mathcal{N}(\mathbf{0},\mathbf{I})$. The exogenous noise $\mathbf{u}_{\mathbf{x}}$ is then (non-deterministically) abducted via variational inference: $p(\mathbf{u}_\mathbf{x} | \mathbf{x}, \mathbf{pa}_\mathbf{x}) \approx q_\phi(\mathbf{z} | \mathbf{x},\mathbf{pa}_\mathbf{x}) \delta(
\boldsymbol{\epsilon} - h^{-1}(\mathbf{x};g_\theta(\mathbf{z},\mathbf{pa}_{\mathbf{x}}))
)$.

In practice, \citet{pawlowski2020deep} used VAEs with limited capacity and \textit{near-deterministic} likelihoods $\sigma^{2} \rightarrow{} 0$ for low resolution data. Although this near-deterministic design choice was motivated by optimization difficulties, there is an alternative nontrivial explanation for its practical success. Notably, \citet{nielsen2020survae} argued that deterministic VAEs optimize an exact log-likelihood like normalizing flows, stating that the VAE encoder inverts the decoder (\textit{self-consistency}). \citet{NEURIPS2022_4eb91efe} recently proved VAE self-consistency in the near-deterministic regime. As such, \citet{pawlowski2020deep}'s near-deterministic VAE setup incidentally emulates a normalizing flow and attempts to deterministically abduct $\mathbf{x}$'s exogenous noise, which also partly explains the poor random sample quality achieved (see Figure~\ref{fig:scales} in Appendix~\ref{app:Supplementary Methods}). 
To address this and generate plausible high-fidelity image counterfactuals, a powerful generative causal mechanism capable of accurate abduction is required.

We propose two deep causal mechanisms based on HVAEs. The first mechanism (Section~\ref{subsec: Exogenous Prior HVAE}) is designed to be directly compatible with standard DSCMs. The second mechanism (Section~\ref{subsec: Hierarchical Latent Mediator}) involves an alternative causal model which is inspired by \textit{causal mediation} analysis. Notably, our HVAE mechanisms are \textit{not} trained in the near-deterministic regime and therefore induce a distribution over causal worlds in their associated probabilistic SCMs $\langle \mathcal{M}, P(U) \rangle$. 
\subsection{Conditional HVAE with an Exogenous Prior}
\label{subsec: Exogenous Prior HVAE}
\begin{figure*}[t!]
    \centering
    \hfill
    \begin{subfigure}{.32\textwidth}
        \centering       
        \begin{tikzpicture}[thick]
            \node[latent] (z) 
            {$\mathbf{z}_{1:L}$};
            \node[obs, below=of z] (x) 
            {$\mathbf{x}$};
            \node[obs, left=of z] (pa) 
            {$\mathbf{pa}_\mathbf{x}$};
            \node[latent, below=of pa] (e) 
            {$\boldsymbol{\epsilon}$};
            \node[right=of z] (phi) 
            {${\boldsymbol\phi}$};
            \node[right=of x] (theta) 
            {${\boldsymbol\theta}$};
            
            \plate {xplate} {(pa)(e)
            (x)} {$N$};
            
            \draw[{Latex[open,scale=1.0]}-{Latex[scale=1.0]}] (e) -- (x) node[midway, below] {$h$} coordinate[midway] (ex);
            \draw[-Circle] (z) edge[out=225, in=90] (ex);
            \draw[-Circle] (pa) edge[out=-45, in=90] (ex);
            \edge[-{Latex[scale=1.0]}]{theta}{z}
            \edge[-{Latex[scale=1.0]}]{theta}{x}
            \edge[-{Latex[scale=1.0]}]{pa}{x}
            \draw[densely dashed,-{Latex[scale=1.0]}] (phi) edge[out=180, in=0] (z);
            \draw[densely dashed,-{Latex[scale=1.0]}] (x) edge[out=90, in=-90] (z);
        \end{tikzpicture}
    \caption{Causal Mechanism for $\mathbf{x}$}
    \label{fig:plate_hscm}
    \end{subfigure}
    \hfill
    \begin{subfigure}{.32\textwidth}
        \centering
        \begin{tikzpicture}[thick]
            \node[obs] (x) {$\mathbf{x}$};
            \node[latent, right=of x, yshift=20pt] (e) {$\boldsymbol{\epsilon}$};
            \node[latent, right=77pt of x] (cfx) {$\widetilde{\mathbf{x}}$};
            \node[latent, above=5pt of e] (z) {$\mathbf{z}_{1:L}$};
            \node[obs, above=of x] (pa) {$\mathbf{pa}_\mathbf{x}$};
            \node[latent, above=of cfx] (cfpa) {$\widetilde{\mathbf{pa}}_\mathbf{x}$};
            \node[latent, right=of pa, yshift=1cm, xshift=-2.5pt] (u) {$\mathbf{U}_\mathbf{pa}$};
            \node[right=of x, yshift=-3pt] (ux) {$\mathbf{U}_\mathbf{x}$};
            
            \edge[-{Latex[scale=1.0]}]{u}{pa}
            \edge[-{Latex[scale=1.0]}]{u}{cfpa}
            
            \plate[inner sep=1.5pt, yshift=3pt] {zeplate} {(z)(e)}{};
            
            \edge[-{Latex[scale=1.0]}]{pa}{x}
            \edge[-{Latex[scale=1.0]}]{cfpa}{cfx}
            
            \draw[{Latex[open,scale=1.0]}-{Latex[scale=1.0]}] (e) -- (x) node[midway, below] {} coordinate[midway] (ex);
            
            \draw[{Latex[scale=1.0]}-{Latex[open,scale=1.0]}] (cfx) -- (e) node[midway, below] {} coordinate[midway] (cfxe);
            
            \draw[-Circle] (z) edge[out=225, in=90] (ex);
            \draw[-Circle] (z) edge[out=-45, in=90] (cfxe);

            \draw[-Circle] (pa) edge[out=-45, in=90] (ex);
            \draw[-Circle] (cfpa) edge[out=-135, in=90] (cfxe);
        \end{tikzpicture}
        \caption{Exogenous Prior: $p_\theta(\mathbf{z}_{1:L})$}
        \label{fig:twin_net}
    \end{subfigure}
    \hfill
    \begin{subfigure}{.32\textwidth}
        \centering
        \begin{tikzpicture}[thick]
            \node[obs] (x) {$\mathbf{x}$};
            \node[latent, right=35pt of x, yshift=0pt] (e) {$\mathbf{U}_\mathbf{x}$};
            \node[latent, right=35pt of e, yshift=-0pt] (cfx) {$\widetilde{\mathbf{x}}$};            
            \node[obs, above=of x] (pa) {$\mathbf{pa}_\mathbf{x}$};
            \node[latent, right=35pt of pa,
            yshift=1cm, xshift=-2.5pt] (u) {$\mathbf{U}_\mathbf{pa}$};
            \node[latent, below=2.5pt of pa, xshift=30pt] (z) {$\mathbf{z}_{1:L}$};
            
            \node[latent, above=of cfx] (cfpa) {$\widetilde{\mathbf{pa}}_\mathbf{x}$};
            \node[latent, below=2.5pt of cfpa, xshift=-30pt] (cfz) {$\widetilde{\mathbf{z}}_{1:L}$};

            \node[latent, right=5pt of z, yshift=22pt] (uz) {$\mathbf{U}_\mathbf{z}$};
            
            \edge[-{Latex[scale=1.0]}]{z}{x}
            \edge[-{Latex[scale=1.0]}]{uz}{z}
            \edge[-{Latex[scale=1.0]}]{uz}{cfz}
            
            \edge[-{Latex[scale=1.0]}]{u}{pa}
            \edge[-{Latex[scale=1.0]}]{u}{cfpa}
            
            \edge[-{Latex[scale=1.0]}]{pa}{x}
            \edge[-{Latex[scale=1.0]}]{cfpa}{cfx}
            \edge[-{Latex[scale=1.0]}]{pa}{z}

            \edge[-{Latex[scale=1.0]}]{e}{x}
            \edge[-{Latex[scale=1.0]}]{e}{cfx}

            \edge[-{Latex[scale=1.0]}]{cfpa}{cfz}
            \edge[-{Latex[scale=1.0]}]{cfz}{cfx}
        \end{tikzpicture}
        \caption{Latent Mediator: $p_\theta(\mathbf{z}_{1:L} \mid \mathbf{pa}_\mathbf{x})$}
        \label{fig:twin_net_sm}
    \end{subfigure}
    \hfill
    \caption{Graphical model of $\mathbf{x}$'s HVAE mechanism (a), and twin network representations of the deep SCMs proposed (b-c).
    (a) The parameters $\boldsymbol{\theta}$ and $\boldsymbol{\phi}$ pertain to the generative and inference models respectively. The mechanism $h(\cdot)$ is invertible and conditioned on $(\mathbf{z}_{1:L}, \mathbf{pa}_{\mathbf{x}})$ using, e.g. reparameterisation. (b) Markovian SCM of (a), with an unconditional prior over $\mathbf{z}_{1:L}$, where $\mathbf{z}_{1:L}$ is part of $\mathbf{x}$'s exogenous noise $\mathbf{U}_\mathbf{x}$. (c) Latent mediator model which is the result of a conditional prior. Here $\mathbf{z}_{1:L}$ becomes a latent mediator since it's no longer exogenous due to the dependence on endogenous variables $\mathbf{pa}_\mathbf{x}$.
    }
\end{figure*}
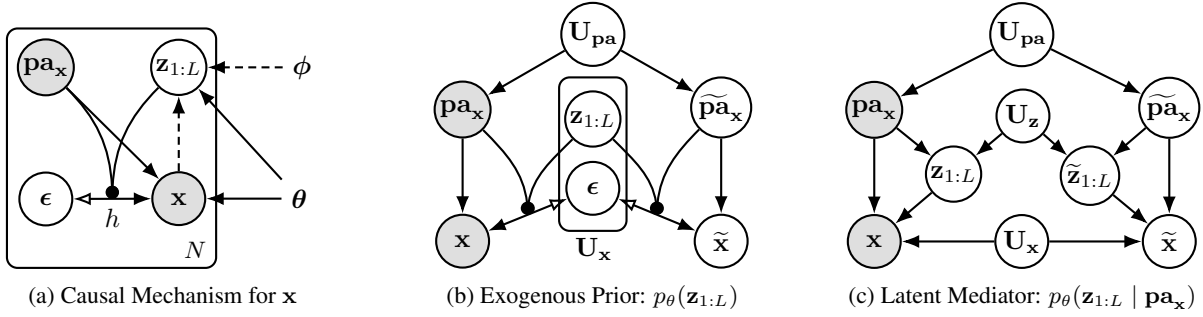

In DSCMs, the VAE's latent code $\mathbf{z}$ is defined as part of the \textit{exogenous} noise for $\mathbf{x}$, so the associated prior $p(\mathbf{z})$ must be \textit{unconditional} due to the underlying Markovian SCM. However, in state-of-the-art HVAEs~\cite{vahdat2020nvae,child2020very}, the prior is not fixed as $\mathcal{N}(\mathbf{0},\mathbf{I})$ but is learned from data. Therefore, some modifications to the generative model are needed to enable sampling conditioned on $\mathbf{pa}_\mathbf{x}$ while keeping the prior exogenous. Conditioning on (counterfactual) parents $\widetilde{\mathbf{pa}}_\mathbf{x}$ is required to generate counterfactuals $\widetilde{\mathbf{x}}$. As shown in Figure~\ref{chvae}, we propose a simple conditional HVAE structure that decouples the prior from the conditioning on $\mathbf{pa}_{\mathbf{x}}$ whilst retaining conditional sampling capability. The generative model is: $p_\theta(\mathbf{x}, \mathbf{z}_{1:L} \mid \mathbf{pa}_{\mathbf{x}}) =$
\begin{align}
    p_\theta(\mathbf{x} \mid \mathbf{z}_{1:L}, \mathbf{pa}_{\mathbf{x}})p_\theta(\mathbf{z}_L)\prod_{i=1}^{L-1} p_\theta(\mathbf{z}_i \mid \mathbf{z}_{>i}),
\end{align}
where we introduce $\mathbf{z}_i$ and $\mathbf{pa}_\mathbf{x}$ into each layer of the \textit{top-down} hierarchy via a parameterized function $\mathbf{f}_i^{\omega}$ as:
\begin{align}
    &\mathbf{h}_i = \mathbf{h}_{i+1} + \mathbf{f}^\omega_i(\mathbf{z}_i, \mathbf{pa}_\mathbf{x}), &\mathbf{z}_i \sim p_\theta(\mathbf{z}_i \mid \mathbf{z}_{>i}),
\end{align}
for $i = L-1,\dots,1$. Note that the initial $\mathbf{h}_\mathrm{init}$ is a vector of learned parameters, and $\mathbf{h}_1$ is the size of $\mathbf{x}$. With this conditioning structure, the prior $p_\theta(\mathbf{z}_{1:L})$ becomes independent of $\mathbf{pa}_\mathbf{x}$ (exogenous), but the likelihood is not: $p_\theta(\mathbf{x} \mid \mathbf{z}_{1:L}, \mathbf{pa}_{\mathbf{x}}) = \mathcal{N}(\mathbf{x} \mid \boldsymbol{\mu}_\theta(\mathbf{h}_1), \boldsymbol{\sigma}_\theta(\mathbf{h}_1))$, allowing us to retain conditional sampling capabilities as required.
Figures~\ref{fig:plate_hscm} and~\ref{fig:twin_net} depict the resulting causal mechanism for $\mathbf{x}$ and the associated SCM's twin network representation.
\paragraph{Discretized Likelihood \& Counterfactuals} Figure~\ref{fig:plate_hscm} depicts the causal mechanism for $\mathbf{x}$. Here we make the observation that since $p(\boldsymbol{\epsilon})$ is a Dirac delta distribution with no learned parameters of its own, training with the invertible mechanism $h$ as in \citet{pawlowski2020deep} is not strictly necessary. That is, rather than using a \textit{change-of-variables} to evaluate the conditional density of $\mathbf{x}$ at $\boldsymbol{\epsilon} = h^{-1}(\mathbf{x};g_\theta (\mathbf{z}_{1:L}, \mathbf{pa}_{\mathbf{x}}))$ during training:
\begin{align}
    p(\mathbf{x} \mid \mathbf{z}_{1:L}, \mathbf{pa}_{\mathbf{x}}) = p_{\mathcal{N}(\mathbf{0},\mathbf{I})}(\boldsymbol{\epsilon}) \left|\mathrm{det}J_{h}(\boldsymbol{\epsilon})
    \right|^{-1},
\end{align}
which requires dequantization of the input data, we can train using more stable likelihoods (e.g. discretized Gaussian~\cite{ho2020denoising}), and infer $\boldsymbol{\epsilon}$ for counterfactuals \textit{only}. Formally, since we assume a Gaussian observational distribution for $\mathbf{x}$, sampling from it entails: 
$\mathbf{x} = \boldsymbol{\mu} + \boldsymbol{\sigma} \odot \boldsymbol{\epsilon}$, $\boldsymbol{\epsilon} \sim \mathcal{N}(\mathbf{0},\mathbf{I})$. Thus, sampling from the counterfactual distribution involves the same $\boldsymbol{\epsilon}$ noise: $\widetilde{\mathbf{x}} = \widetilde{\boldsymbol{\mu}} + \widetilde{\boldsymbol{\sigma}} \odot \boldsymbol{\epsilon}$, where $\boldsymbol{\epsilon} = (\mathbf{x} - \boldsymbol{\mu}) \oslash \boldsymbol{\sigma}$. Here $(\boldsymbol{\mu}, \boldsymbol{\sigma})$ are per pixel mean/std. outputs of the decoder $g_\theta(\mathbf{z}_{1:L}, \mathbf{pa}_{\mathbf{x}})$, and similarly $(\widetilde{\boldsymbol{\mu}}, \widetilde{\boldsymbol{\sigma}}) = g_\theta(\mathbf{z}_{1:L}, \widetilde{\mathbf{pa}}_{\mathbf{x}})$. 
%
\subsection{Hierarchical Latent Mediator Model}
\label{subsec: Hierarchical Latent Mediator}
In Markovian SCMs, all causal effects are identifiable from observed data~\cite{pearl2009causality}, which motivated our setup in Section~\ref{subsec: Exogenous Prior HVAE}. However, when a VAE's prior $p(\mathbf{z})$ is unconditional (exogenous), the VAE is unidentifiable in the general case~\cite{locatello2019challenging}. This means we are not guaranteed to recover the true parameters of the generative model given infinite data. Using such a VAE for $\mathbf{x}$'s mechanism may affect the abduction step in our associated SCM since there can be multiple solutions for $\mathbf{z}$ which yield the same likelihood $p_\theta(\mathbf{x})$. \citet{khemakhem2020variational} showed that VAE identifiability can be established up to equivalence permutation if the prior is conditioned on additionally observed variables~\cite{hyvarinen2019nonlinear}. In our HVAE mechanism for $\mathbf{x}$, this amounts to having a prior on $\mathbf{z}_{1:L}$ conditioned on the endogenous causes of $\mathbf{x}$ as: $p_\theta(\mathbf{z}_{1:L} \mid \mathbf{pa}_\mathbf{x}) =$
\begin{align}
      p_\theta(\mathbf{z}_{L} \mid \mathbf{pa}_\mathbf{x}) \prod_{i=1}^{L-1} p_\theta(\mathbf{z}_{i} \mid \mathbf{z}_{>i}, \mathbf{pa}_\mathbf{x}).
\end{align}
The resulting generative model can be seen in Figures~\ref{chvae_c} \& \ref{chvae2}, and the new associated SCM is shown in Figure~\ref{fig:twin_net_sm}.

This model differs from the one in Section~\ref{subsec: Exogenous Prior HVAE} due to $\mathbf{z}_{1:L}$'s dependence on $\mathbf{pa}_{\mathbf{x}}$: i.e. the role of $\mathbf{z}_{1:L}$ has shifted from being part of $\mathbf{x}$'s exogenous noise, to being a latent \textit{mediator} we must infer. Despite the conditional prior on $\mathbf{z}_{1:L}$, we show that this model is still Markovian, as we have jointly independent exogenous noise variables: $p(\mathbf{U}) = p(\mathbf{U}_\mathbf{x})\prod_{k=1}^Kp(\mathbf{U}_{\mathbf{pa}_k})\prod_{i=1}^{L}p(\mathbf{U}_{\mathbf{z}_i})$. To compute counterfactuals $\widetilde{\mathbf{x}}$, we must now infer the counterfactual mediator $\widetilde{\mathbf{z}}_{1:L}$. If we somehow have access to true counterfactuals $\widetilde{\mathbf{x}}$, the counterfactual mediator could be inferred directly via: $\widetilde{\mathbf{z}}_{1:L} \sim q_\phi(\widetilde{\mathbf{z}}_{1:L} \mid \widetilde{\mathbf{x}}, \widetilde{\mathbf{pa}}_\mathbf{x})$, where $\widetilde{\mathbf{z}}_{1:L}$ is sampled using the same noise $\mathbf{U}_{\mathbf{z}_{1:L}}$ used for sampling $\mathbf{z}_{1:L}$. In most cases, we do not know $\widetilde{\mathbf{x}}$ so we must rely on approximations. We propose to first infer the \textit{factual} mediator $\mathbf{z}_{1:L}$ consistent with $(\mathbf{x}, \mathbf{pa}_\mathbf{x})$ in the anticausal direction as:
\begin{align}
    \mathbf{x} \coloneqq& \ f_\theta(\mathbf{z}_{1:L}, \mathbf{pa}_{\mathbf{x}}, \mathbf{U}_{\mathbf{x}})
    \approx p_\theta(\mathbf{x} \mid \mathbf{z}_{1:L}, \mathbf{pa}_\mathbf{x})
    \\[2pt]
    \label{inv_abd}
    \mathbf{z}_{1:L} =& \ f_\theta^{-1}(\mathbf{x}, \mathbf{pa}_{\mathbf{x}}, \mathbf{U}_\mathbf{x}) \approx q_\phi(\mathbf{z}_{1:L} \mid \mathbf{x}, \mathbf{pa}_\mathbf{x}),
\end{align}
where $\mathbf{U}_{\mathbf{x}} \sim \mathcal{N}(\mathbf{0},\mathbf{I})$. As shown in equation~\eqref{inv_abd}, this approximately inverts $\mathbf{x}$'s mechanism (decoder) w.r.t. the mediator. Recall that the optimal VAE encoder inverts the decoder (\textit{self-consistency}~\cite{NEURIPS2022_4eb91efe}). Since each inferred $\mathbf{z}_i \sim q_\phi(\mathbf{z}_i \mid \mathbf{z}_{>i}, \mathbf{x}, \mathbf{pa}_\mathbf{x})$ is Gaussian distributed, we can invert the reparameterized sampling to abduct the exogenous noise $\mathbf{U}_{\mathbf{z}_i} \sim \mathcal{N}(\mathbf{0},\mathbf{I})$ at each layer:
\begin{align}
    \label{eq:uz_abduction}
    \mathbf{U}_{\mathbf{z}_i} = (\mathbf{z}_i - \boldsymbol{\mu}_i^q(\mathbf{z}_{>i}, \mathbf{x}, \mathbf{pa}_\mathbf{x})) \oslash \boldsymbol{\sigma}_i^q(\mathbf{z}_{>i}, \mathbf{x}, \mathbf{pa}_\mathbf{x}).
\end{align}
The same abducted exogenous noise components are then used to sample the respective counterfactual mediator $\widetilde{\mathbf{z}}_{1:L}$. To help preserve the identity of observations $(\mathbf{x}, \mathbf{pa}_\mathbf{x})$ in their inferred counterfactuals $\widetilde{\mathbf{x}}$, we found it beneficial to construct a mixture distribution of the counterfactual prior and the factual posterior as: $r(\widetilde{\mathbf{z}}_i \mid \widetilde{\mathbf{z}}_{>i}, \mathbf{x}, \mathbf{pa}_\mathbf{x}, \widetilde{\mathbf{pa}}_\mathbf{x})=$
\begin{align}
    &\pi p_\theta(\widetilde{\mathbf{z}}_i \mid \widetilde{\mathbf{z}}_{>i}, \widetilde{\mathbf{pa}}_\mathbf{x})
    + (1-\pi) q_\phi(\mathbf{z}_i \mid \mathbf{z}_{>i}, \mathbf{x}, \mathbf{pa}_{\mathbf{x}}),
\end{align}
where $0 \leq \pi \leq 1$. We then sample each $\widetilde{\mathbf{z}}_i \sim r(\widetilde{\mathbf{z}}_i \mid \widetilde{\mathbf{z}}_{>i}, \mathbf{x}, \mathbf{pa}_\mathbf{x}, \widetilde{\mathbf{pa}}_\mathbf{x})$ using the (abducted) noise from Eq.~\eqref{eq:uz_abduction}:
\begin{align}
    \widetilde{\mathbf{z}}_i = \boldsymbol{\mu}_i^r(\widetilde{\mathbf{z}}_{>i}, \widetilde{\mathbf{pa}}_\mathbf{x}) + \boldsymbol{\sigma}_i^r(\widetilde{\mathbf{z}}_{>i}, \widetilde{\mathbf{pa}}_\mathbf{x}) \odot \mathbf{U}_{\mathbf{z}_i}.    
\end{align}
This way each underlying mechanism $\widetilde{\mathbf{z}}_i \coloneqq f(\widetilde{\mathbf{pa}}_{\mathbf{z}_i}, \mathbf{U}_{\mathbf{z}_i})$, with $\widetilde{\mathbf{pa}}_{\mathbf{z}_i}=\{ \widetilde{\mathbf{z}}_{>i},\widetilde{\mathbf{pa}}_\mathbf{x}\}$, is approximated by the mixture distribution rather than only by the counterfactual prior. Finally, to sample counterfactuals $\widetilde{\mathbf{x}}$ given $(\mathbf{x}, \mathbf{pa}_\mathbf{x})$ and $\widetilde{\mathbf{pa}}$ (e.g. result of an intervention $do(\mathbf{pa} \coloneqq \widetilde{\mathbf{pa}})$) we have:
\begin{align}
    \mathbf{U}_\mathbf{x} &= (\mathbf{x} - \boldsymbol{\mu}(\mathbf{z}_{1:L}, \mathbf{pa}_\mathbf{x})) \oslash \boldsymbol{\sigma}(\mathbf{z}_{1:L}, \mathbf{pa}_\mathbf{x}),
    \\[2pt]
    \widetilde{\mathbf{x}} &= \boldsymbol{\mu}(\widetilde{\mathbf{z}}_{1:L}, \widetilde{\mathbf{pa}}_\mathbf{x}) + \boldsymbol{\sigma}(\widetilde{\mathbf{z}}_{1:L}, \widetilde{\mathbf{pa}}_\mathbf{x}) \odot \mathbf{U}_\mathbf{x},
\end{align}
where $\boldsymbol{\mu}(\cdot)$, $\boldsymbol{\sigma}(\cdot)$ are the decoder's output mean and std. Here $\mathbf{U}_\mathbf{x}$ is the (only) exogenous noise for $\mathbf{x}$; it assumes a similar role to $\boldsymbol{\epsilon}$ in the exogenous prior model of Section~\ref{subsec: Exogenous Prior HVAE}.
\paragraph{Direct, Indirect \& Total Effects.} The proposed latent mediator model allows us to compute causal effects w.r.t. the parents $\mathbf{pa}_\mathbf{x}$ and the mediator $\mathbf{z}_{1:L}$ separately. Let $g(\mathbf{pa}_\mathbf{x}, \mathbf{z}_{1:L})$ denote the output of our generative model for $\mathbf{x}$; the following causal quantities can be computed:
\begin{align}
    \mathrm{DE}_\mathbf{x}(\widetilde{\mathbf{pa}}) &= g(\widetilde{\mathbf{pa}}_\mathbf{x}, \mathbf{z}_{1:L}) - g({\mathbf{pa}}_\mathbf{x}, \mathbf{z}_{1:L})
    \\
    \mathrm{IE}_\mathbf{x}(\widetilde{\mathbf{z}}_{1:L}) &= g({\mathbf{pa}}_\mathbf{x}, \widetilde{\mathbf{z}}_{1:L}) - g({\mathbf{pa}}_\mathbf{x}, \mathbf{z}_{1:L})
    \\
    \mathrm{TE}_\mathbf{x}(\widetilde{\mathbf{pa}}, \widetilde{\mathbf{z}}_{1:L}) &= g(\widetilde{\mathbf{pa}}_\mathbf{x}, \widetilde{\mathbf{z}}_{1:L}) - g({\mathbf{pa}}_\mathbf{x}, \mathbf{z}_{1:L}),
\end{align}
which are known as the Direct (DE), Indirect (IE), and Total Effects (TE) in causal mediation analysis and epidemiology~\cite{robins1992identifiability,10.5555/2074022.2074073}. For example, $\widetilde{\mathbf{x}} = g(\mathbf{pa}_\mathbf{x}, \widetilde{\mathbf{z}}_{1:L})$ is the counterfactual outcome of $\mathbf{x}$ given the observed parents $\mathbf{pa}_\mathbf{x}$ and the (counterfactual) mediator we would have observed $\widetilde{\mathbf{z}}_{1:L}$ had the parents been $\widetilde{\mathbf{pa}}_{\mathbf{x}}$. This is known as a \textit{cross-world} or \textit{apriori} counterfactual. We argue that the above causal quantities could be useful for offering causal explanations of outcomes when applied to high-dimensional structured variables such as images. 
%
\subsection{Ignored Counterfactual Conditioning}
\label{subsec:ignored}
A primary issue with conditional generative models is that they are free to ignore conditioning $c$ by finding a solution satisfying $p_\theta(\mathbf{x}|c) = p_\theta(\mathbf{x})$~\cite{chen2016infogan}. 
In our case, the decoder may not learn to disentangle the effect of the exogenous noise $\mathbf{z}_{1:L}$ and the parents $\mathbf{pa}_\mathbf{x}$ on the output. 
This also affects what we call \textit{counterfactual conditioning}, i.e. the act of conditioning the generative model on the counterfactual parents $\widetilde{\mathbf{pa}}_{\mathbf{x}}$, holding $\mathbf{x}$'s abducted noise $\mathbf{z}_{1:L}$ fixed, to generate a counterfactual $\widetilde{\mathbf{x}}$. We find that counterfactual conditioning can be ignored, even when observational conditioning is not (e.g. in random conditional sampling). To mitigate this problem, we propose an information theory inspired strategy for enforcing counterfactual conditioning.

\paragraph{Counterfactual Training.} Counterfactuals $\widetilde{\mathbf{x}}$ should obey counterfactual conditioning on $\widetilde{\mathbf{pa}}_k$ (e.g. result of an intervention $do(\mathbf{pa}_{k} \coloneqq \widetilde{\mathbf{pa}}_{k})$) by manifesting semantically meaningful changes from $\mathbf{x}$. Thus, the Mutual Information (MI) between a counterfactual $\widetilde{\mathbf{x}}$ and $\widetilde{\mathbf{pa}}_k$ should be: $I(\widetilde{\mathbf{pa}}_k;\widetilde{\mathbf{x}}) > 0$. Maximizing this MI term directly is intractable, but we can use a variational technique~\cite{agakov2004algorithm} to lower bound it as: $I(\widetilde{\mathbf{pa}}_k;\widetilde{\mathbf{x}})=$
\begin{align}
 &\mathbb{E}_{p(\widetilde{\mathbf{pa}}_k,\widetilde{\mathbf{x}})}\left[\log \left(\frac{p(\widetilde{\mathbf{pa}}_k \mid \widetilde{\mathbf{x}})}{p(\widetilde{\mathbf{pa}}_k)} \cdot \frac{q_\psi(\widetilde{\mathbf{pa}}_k \mid \widetilde{\mathbf{x}})}{q_\psi(\widetilde{\mathbf{pa}}_k \mid \widetilde{\mathbf{x}})} \right) \right] 
    \\[4pt] 
    = & \ \mathbb{E}_{p(\widetilde{\mathbf{pa}}_k,\widetilde{\mathbf{x}})}
    \left[\log \frac{q_\psi(\widetilde{\mathbf{pa}}_k \mid \widetilde{\mathbf{x}})}{p(\mathbf{pa}_k)} \right] 
    \\[2pt]   \nonumber & \qquad + \mathbb{E}_{p(\widetilde{\mathbf{x}})}D_{\mathrm{KL}}(p(\widetilde{\mathbf{pa}}_k \mid \widetilde{\mathbf{x}}) \parallel q_\psi(\widetilde{\mathbf{pa}}_k \mid \widetilde{\mathbf{x}}))
    \\[4pt] \geq & \ \mathbb{E}_{p(\widetilde{\mathbf{pa}}_k,\widetilde{\mathbf{x}})}\left[\log q_\psi(\widetilde{\mathbf{pa}}_k \mid \widetilde{\mathbf{x}}) \right] + H(\widetilde{\mathbf{pa}}_k),
    \label{eq:mi}
\end{align}
where $q_\psi(\widetilde{\mathbf{pa}}_k \mid \widetilde{\mathbf{x}})$ is a learned variational distribution for approximating $p(\widetilde{\mathbf{pa}}_k \mid \widetilde{\mathbf{x}})$. This MI bound motivates the optimization of a probabilistic predictor $q_\psi(\widetilde{\mathbf{pa}}_k \mid \widetilde{\mathbf{x}})$ for each parent, as a way to enforce counterfactual conditioning.

In practice, we optimize the following modified objective with $H(\widetilde{\mathbf{pa}}_k)$ held constant. We perform random interventions on $\mathbf{pa}_{\mathbf{x}}$ by sampling each parent independently from its marginal distribution $\widetilde{\mathbf{pa}}_k \sim p(\mathbf{pa}_k)$, and maximize the log-likelihood of the probabilistic predictors given a sampled counterfactual $\widetilde{\mathbf{x}}$ from the counterfactual distribution:
\begin{align}
    &\max_{P_\mathcal{M}, q_{\boldsymbol{\psi}}}
    \mathbb{E}_{
    p_{\mathrm{data}} (\mathbf{x}, \mathbf{pa}_{\mathbf{x}})}\left[-\mathcal{L}_{\mathrm{CT}}(\mathcal{M}; \mathbf{x}, \mathbf{pa}_{\mathbf{x}})\right],
    \label{eq:cf_train}
\end{align}
where the counterfactual loss is: $\mathcal{L}_{\mathrm{CT}}(\mathcal{M}; \mathbf{x}, \mathbf{pa}_{\mathbf{x}}) =$
\begin{align}
    -\sum_{k=1}^K \mathop{\mathop{\mathbb{E}}_{\widetilde{\mathbf{pa}}_k \sim p(\mathbf{pa}_k),}}_{\widetilde{\mathbf{x}} \sim P_{\mathcal{M}}(\widetilde{\mathbf{x}} \mid do(\widetilde{\mathbf{pa}}_k), \mathbf{x})}\left[
    \log q_{\boldsymbol{\psi}_k}(\widetilde{\mathbf{pa}}_k \mid  \widetilde{\mathbf{x}}) \right].
\end{align}
Recall that all the mechanisms in our SCM $\mathcal{M}$ have optimizable parameters, i.e. the generative and inference parameters $\{\theta, \phi \}$ pertaining to $\mathbf{x}$'s HVAE mechanism, and $\boldsymbol{\omega}$ denoting the parameters of all other mechanisms. The objective in equation~\eqref{eq:cf_train} can be optimized by a variant of the \textit{Wake-Sleep} algorithm~\cite{hinton1995wake}, alternating between optimizing the parameters of the SCM mechanisms $\{\theta, \phi, \boldsymbol{\omega}\}$ with the parent predictors parameters $\boldsymbol{\psi}$ fixed and vice-versa. In practice, we found it more effective to pre-train all the SCM mechanisms and parent predictors on observational data first, yielding $\{\hat{\theta}, \hat{\phi}, \hat{\boldsymbol{\omega}}, \hat{\boldsymbol{\psi}\}}$. Then, optimize equation~\eqref{eq:cf_train} by fine-tuning $\mathbf{x}$'s mechanism only, i.e. updating the HVAE's parameters $\{\hat{\theta}, \hat{\phi}\}$ with $\{\hat{\boldsymbol{\omega}}, \hat{\boldsymbol{\psi}}\}$ fixed.

\begin{figure*}[!t]
    \centering
    \hfill
    \begin{subfigure}{.72\textwidth}
        \centering
        \includegraphics[trim={0 0 0 0}, clip, width=\textwidth]{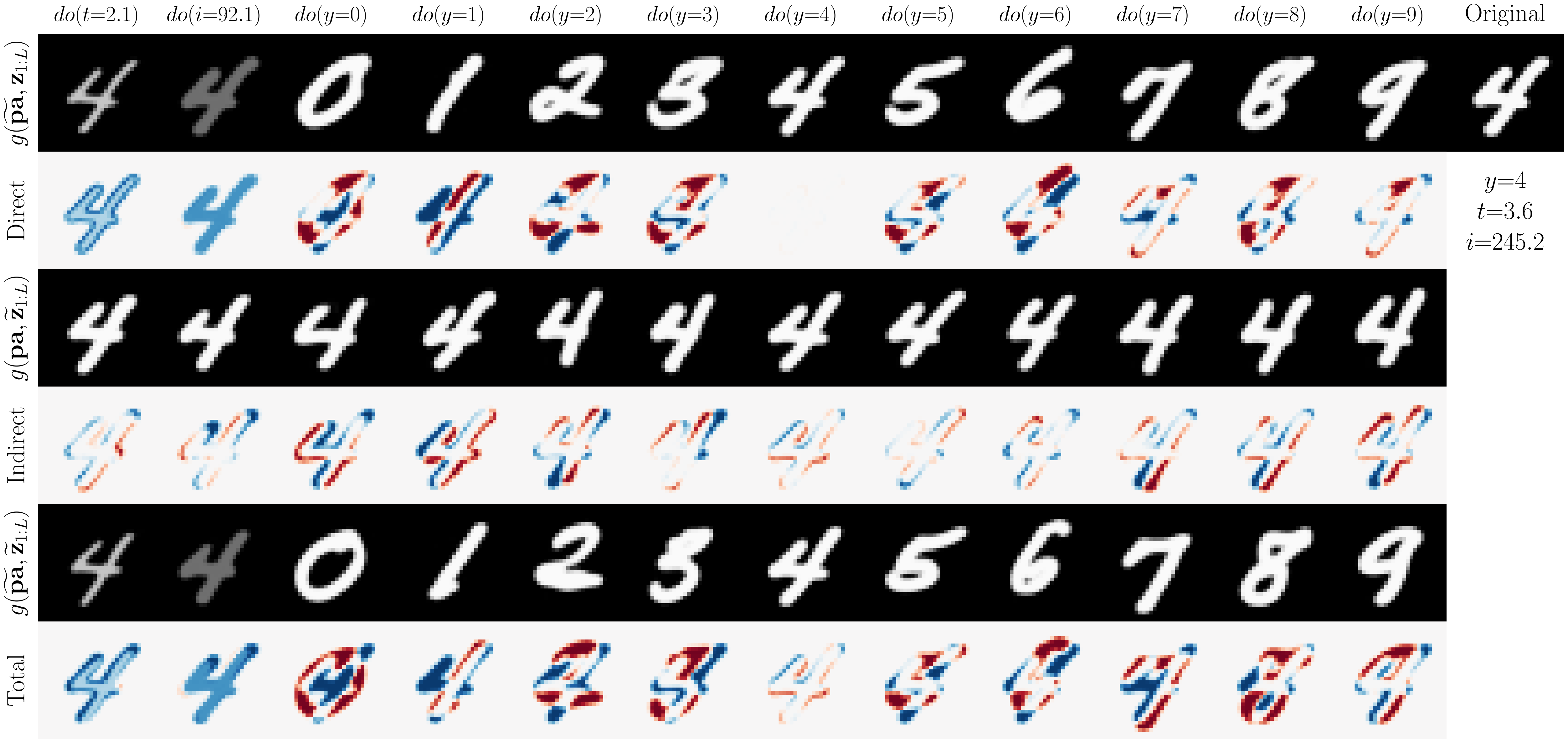} 
    \end{subfigure}
    \hfill
    \begin{subfigure}{.27\textwidth}
        \centering
        \begin{tikzpicture}[thick]
            \node[obs] (x) {$\mathbf{x}$};
            \node[obs, left=15pt of x] (y) {$y$};
            \node[obs, left=15pt of y] (i) {$i$};
            \node[obs, left=15pt of i] (t) {$t$};
            \node[latent, xshift=-18pt, above=25pt of y] (z) {$\mathbf{z}_{1:L}$};

            \node[latent, right=21pt of z] (uz) {$\mathbf{U}_{\mathbf{z}}$};
            \node[latent, below=20pt of t] (ut) {$\mathbf{U}_t$};
            \node[latent, below=20pt of i] (ui) {$\mathbf{U}_i$};
            \node[latent, below=20pt of y] (uy) {$\mathbf{U}_y$};
            \node[latent, below=20pt of x] (ux) {$\mathbf{U}_\mathbf{x}$};

            \edge[-{Latex[scale=1.0]}]{y}{x}
            \edge[-{Latex[scale=1.0]}]{t}{i}

            \edge[-{Latex[scale=1.0]}]{y}{z}
            \edge[-{Latex[scale=1.0]}]{t}{z}
            \edge[-{Latex[scale=1.0]}]{i}{z}
            \edge[-{Latex[scale=1.0]}]{z}{x}

            \draw [-{Latex[scale=1.0]}] (i) to [out=-45,in=-135] (x);
            \draw [-{Latex[scale=1.0]}] (t) to [out=-45,in=-135] (x);

            \draw [-{Latex[scale=1.0]}] (ux) to [out=90,in=-90] (x);
            \draw [-{Latex[scale=1.0]}] (ut) to [out=90,in=-90] (t);
            \draw [-{Latex[scale=1.0]}] (ui) to [out=90,in=-90] (i);
            \draw [-{Latex[scale=1.0]}] (uy) to [out=90,in=-90] (y);
            \draw [-{Latex[scale=1.0]}] (uz) to [out=-180,in=0] (z);
        \end{tikzpicture}
        \caption{Latent mediator SCM for Morpho-MNIST. Observed variables in the graph: image $(\mathbf{x})$, digit $(y)$, stroke thickness $(t)$ and pixel intensity $(i)$.}
        \label{mediator_morphomnist}
    \end{subfigure}
    \caption{Morpho-MNIST counterfactuals from our latent mediator SCM. \textit{Direct}, \textit{indirect} and \textit{total} causal effects of interventions are shown (red: increase; blue: decrease). Recall that $\widetilde{\mathbf{x}}_{\mathrm{IE}} \sim g(\mathbf{pa}_\mathbf{x}, \widetilde{\mathbf{z}}_{1:L})$ are \textit{cross-world} counterfactuals, i.e. the potential outcome of $\mathbf{x}$ given $\mathbf{pa}_\mathbf{x}$ and the (counterfactual) mediator we would have observed $\widetilde{\mathbf{z}}_{1:L}$ had $\mathbf{pa}_\mathbf{x}$ been $\widetilde{\mathbf{pa}}_\mathbf{x}$.
    }
    \hfill
\end{figure*}
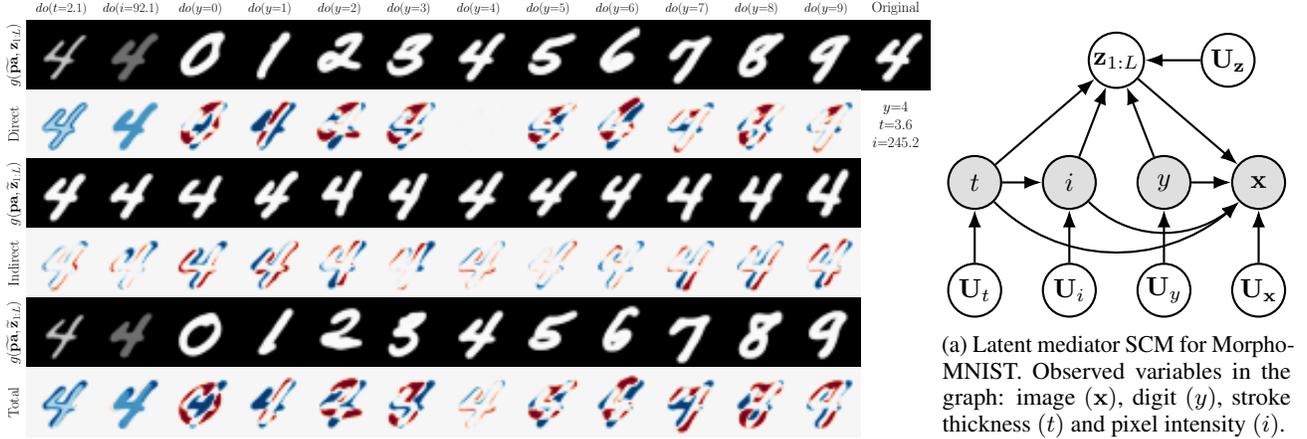
\paragraph{Constrained Counterfactual Training.} One issue with fine-tuning $\mathbf{x}$'s HVAE mechanism with counterfactual training is that the original performance on observational data can deteriorate as we update $\{\hat{\theta}, \hat{\phi}\}$. To mitigate this, we propose to reframe counterfactual training as a Lagrangian optimization problem, using the differential multiplier method~\cite{platt1987constrained}. Our proposed constraint is the pre-trained HVAE's negative ELBO (free energy $\mathcal{F}_{\mathrm{FE}}$) averaged over the observational data: $c \coloneqq \mathbb{E}_{p_{\mathrm{data}}(\mathbf{x}, \mathbf{pa}_\mathbf{x})}[\mathcal{F}_\mathrm{FE}(\hat{\theta}, \hat{\phi}; \mathbf{x}, \mathbf{pa}_\mathbf{x})]$, which should not increase during counterfactual training. Formally, the constrained counterfactual optimization problem is
\begin{align}
    \argmin_{\theta, \phi}& \ 
    \mathbb{E}_{
    p_{\mathrm{data}} (\mathbf{x}, \mathbf{pa}_{\mathbf{x}})}\left[
    \mathcal{L}_\mathrm{CT}(\mathcal{M};\mathbf{x}, \mathbf{pa}_{\mathbf{x}})\right]
    \\\mathrm{s.t.}& \ \mathcal{F}_{\mathrm{FE}}(\theta, \phi; \mathbf{x}, \mathbf{pa}_\mathbf{x}) \leq c,
\end{align}
rewritten as the Lagrangian: $\mathcal{L}_{\mathrm{Lg}}(\theta, \phi, \lambda; \mathbf{x}, \mathbf{pa}_{\mathbf{x}}) =$ 
\begin{align}
    \label{eq:lagrangian}
    \mathcal{L}_{\mathrm{CT}}(\mathcal{M}; \mathbf{x}, \mathbf{pa}_{\mathbf{x}}) - \lambda(c - \mathcal{F}_{\mathrm{FE}}(\theta, \phi; \mathbf{x}, \mathbf{pa}_\mathbf{x})).
\end{align}
Optimizing this Lagrangian 
involves performing gradient descent on the HVAE's parameters $\theta$ and $\phi$, and gradient \textit{ascent} on the Lagrange multiplier $\lambda$. 
The intended effect is to fine-tune $\mathbf{x}$'s HVAE mechanism to improve counterfactual conditioning by maximizing $I(\widetilde{\mathbf{pa}}_k;\widetilde{\mathbf{x}})$, without degrading the original performance on observational data.
\begin{table*}[!t]
    \centering
    \footnotesize
    \caption{Quantitative evaluation of generated counterfactuals for the Morpho-MNIST test set. Identical learned SCM mechanisms for $t$, $i$, and $y$ are used in all cases to enable direct comparison of causal mechanisms for $\mathbf{x}$. `Baseline' refers to the original DSCM~\cite{pawlowski2020deep}, $\text{EXO}_{\text{DE}}$ denotes our exogenous prior HVAE, whereas $\text{MED}_{\text{DE}}$ and $\text{MED}_{\text{TE}}$ refer to direct and total effect counterfactuals from our latent mediator model respectively. `bpd' denotes bits per dimension, i.e \textit{distortion} of null-interventions plus the \textit{rate} ($D_{\text{KL}}(q\parallel p)$). `$\sim$ Prior' denotes random samples from the model directly above. Accompanying table reporting standard deviations of multiple runs and extra ablations are shown in Appendix~\ref{app:Morpho-MNIST}.
    } 
    \label{tab:morphomnist}
    \begin{tabular}{lcc|cccc|cccc|cccc}
        \toprule
        & & &\multicolumn{4}{c}{\textsc{Thickness} $(t)$ MAE $\downarrow$} & \multicolumn{4}{c}{\textsc{Intensity} $(i)$ MAE $\downarrow$} & \multicolumn{4}{c}{\textsc{Digit} $(y)$ Acc. (\%) $\uparrow$} \\[4pt]   
        \textsc{Method} & $\beta$ & bpd $\downarrow$ & $do(t)$ & $do(i)$ & $do(y)$ & mix & $do(t)$ & $do(i)$ & $do(y)$ & mix & $do(t)$ & $do(i)$ & $do(y)$ & mix \\        
        \midrule
        Baseline & 1 & {2.04} & .112 \cmt{(1e-3)} & .178 \cmt{(1e-3)} & .175 \cmt{(2e-3)} & {.177} \cmt{(1e-3)} & 8.31 \cmt{(1e-3)} & 8.10 \cmt{(2e-2)} & 10.4 \cmt{(3e-2)} & {9.61} \cmt{(4e-3)} & 99.20 \cmt{(4e-4)} & 99.08 \cmt{(8e-4)} & {83.18} \cmt{(7e-4)} & {89.54} \cmt{(2e-3)} \\
        \ $\sim$ Prior & 1 & N/A & .193 \cmt{(3e-3)} & .225 \cmt{(5e-4)} & .191 \cmt{(1e-3)} & .209 \cmt{(2e-3)} & 10.5 \cmt{(5e-2)} & 11.1 \cmt{(3e-2)} & 10.6 \cmt{(9e-3)} & 10.8 \cmt{(1e-2)} & 82.75 \cmt{(3e-3)} & 81.10 \cmt{(2e-3)} & 82.62 \cmt{(3e-3)} & 81.99 \cmt{(3e-3)} \\
        Baseline & 3 & 2.17 & .126 \cmt{(2e-3)} & .185 \cmt{(1e-3)} & .149 \cmt{(8e-4)} & .171 \cmt{(1e-3)} & 14.1 \cmt{(4e-2)} & 15.5 \cmt{(1e-2)} & 15.1 \cmt{(8e-2)} & 15.6 \cmt{(0.13)} & 99.47 \cmt{(6e-4)} & 99.34 \cmt{(5e-4)} & 97.89 \cmt{(1e-3)} & 98.34 \cmt{(1e-3)} \\
        \midrule
        $\text{EXO}_{\text{DE}}$ & 1 & {.674} & .125 \cmt{(9e-4)} & .140 \cmt{(2e-4)}  & .149 \cmt{(6e-4)} & {.148} \cmt{(2e-3)} & 1.78 \cmt{(2e-2)} & 2.08 \cmt{(3e-2)} & 1.87 \cmt{(1e-2)} & {2.24} \cmt{(4e-2)} & 99.31
        \cmt{(5e-4)} & 98.88 \cmt{(5e-4)} & {99.49} \cmt{(3e-4)} & {99.23} \cmt{(3e-4)}
        \\
        \ $\sim$ Prior & 1 & N/A & .178 \cmt{(2e-3)} & .192 \cmt{(9e-4)} & .175 \cmt{(2e-3)} & .186 \cmt{(8e-5)} & 2.18 \cmt{(1e-2)} & 3.08 \cmt{(3e-2)} & 2.20 \cmt{(1e-2)} & 2.74 \cmt{(9e-3)} & 98.30 \cmt{(3e-4)} & 97.68 \cmt{(1e-3)} & 98.49 \cmt{(1e-3)} & 97.95 \cmt{(3e-4)} \\
        $\text{EXO}_{\text{DE}}$ & 3 & .942 & .129 \cmt{(1e-3)} & .133 \cmt{(1e-3)}  & .142 \cmt{(2e-4)} & .139 \cmt{(2e-3)} & 1.83 \cmt{(1e-2)} & 2.70 \cmt{(4e-2)} & 1.77 \cmt{(6e-3)} & {2.32} \cmt{(1e-2)} & 99.46
        \cmt{(1e-4)} & 99.01 \cmt{(6e-4)} & {99.73} \cmt{(3e-4)} & {99.34} \cmt{(1e-3)}
        \\
        \midrule
        $\text{MED}_{\text{DE}}$ & 1 & {.682} & .125 \cmt{(6e-4)} & .137 \cmt{(4e-5)} & .157 \cmt{(1e-3)} & {.149} \cmt{(2e-3)} & 1.65 \cmt{(2e-2)} & 1.48 \cmt{(2e-2)} & 1.80 \cmt{(1e-2)} & {1.89} \cmt{(3e-4)} & 99.38
        \cmt{(3e-4)} & 98.73 \cmt{(6e-4)} & {99.47} \cmt{(4e-5)} & {99.09} \cmt{(3e-4)}
        \\
        $\text{MED}_{\text{TE}}$ & 1 & {.682} & .141 \cmt{(2e-4)} & .153 \cmt{(6e-4)} & .146 \cmt{(1e-3)} & {.150} \cmt{(7e-4)} & 1.72 \cmt{(2e-2)} & 2.17 \cmt{(1e-2)} & 1.78 \cmt{(5e-3)} & {2.01} \cmt{(3e-3)} & 99.75 \cmt{(4e-4)} & 99.30
        \cmt{(5e-4)} & {99.68} \cmt{(4e-4)} & {99.41} \cmt{(3e-4)}
        \\
        $\text{MED}_{\text{DE}}$ & 3 & .941 & .133 \cmt{(8e-4)} & .146 \cmt{(3e-4)} & .139 \cmt{(1e-3)} & .145 \cmt{(4e-4)} & 1.94 \cmt{(9e-3)} & 2.71 \cmt{(2e-2)} & 1.94 \cmt{(9e-3)} & 2.45 \cmt{(3e-2)} & 99.45 \cmt{(1e-4)} & 99.15 \cmt{(2e-4)} & {99.62} \cmt{(8e-4)} & {99.40} \cmt{(2e-4)}
        \\
        $\text{MED}_{\text{TE}}$ & 3 & .941 & .130 \cmt{(1e-3)} & .141 \cmt{(8e-4)} & .135 \cmt{(7e-4)} & .138 \cmt{(1e-3)} & {2.10} \cmt{(9e-3)} & 3.11 \cmt{(2e-2)} & 2.13 \cmt{(6e-3)} & {2.69} \cmt{(4e-2)} & 99.85 \cmt{(5e-4)} & 99.65
        \cmt{(4e-4)} & {99.79} \cmt{(4e-4)} & {99.71} \cmt{(1e-4)}
        \\
        \bottomrule
    \end{tabular}
\end{table*}
\begin{figure*}[!t]
    \centering
    \hfill
    \begin{subfigure}{.74\textwidth}
        \centering
        \includegraphics[trim={0 0 0 0}, clip, width=\textwidth]{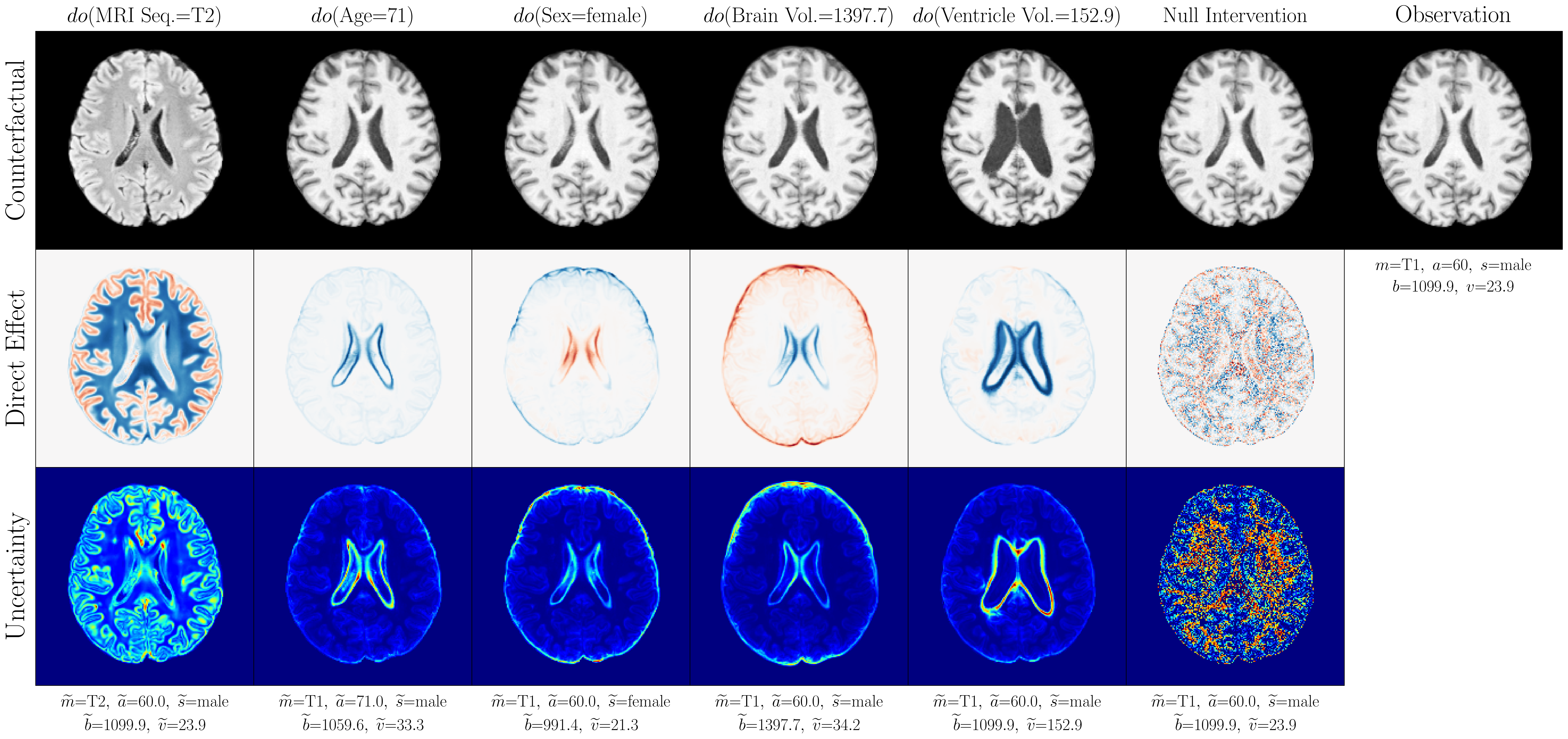}
    \end{subfigure}
    \hfill
    \begin{subfigure}{.25\textwidth}
        \centering
        \begin{tikzpicture}[thick]
            \node[obs] (x) {$\mathbf{x}$};
            \node[obs, left=18pt of x] (m) {$m$};
            \node[obs, above=18pt of m] (v) {$v$};
            \node[obs, right=18pt of x] (s) {$s$};
            \node[obs, above=18pt of s] (b) {$b$};
            \node[obs, above=18pt of x] (a) {$a$};
            \node[latent, xshift=-12pt, below=15pt of x] (zx) {$\mathbf{z}_{1:L}$};
            \node[latent, xshift=12pt, below=15pt of x] (ux) {$\boldsymbol{\epsilon}$};
            \node[latent, xshift=0, below=15pt of m] (um) {$\mathbf{U}_m$};
            \node[latent, below=15pt of s] (us) {$\mathbf{U}_s$};
            \node[latent, above=15pt of v] (uv) {$\mathbf{U}_v$};
            \node[latent, above=15pt of b] (ub) {$\mathbf{U}_b$};
            \node[latent, above=15pt of a] (ua) {$\mathbf{U}_a$};

            \edge[-{Latex[scale=1.0]}]{m}{x}
            \edge[-{Latex[scale=1.0]}]{s}{x}
            \edge[-{Latex[scale=1.0]}]{s}{b}
            \edge[-{Latex[scale=1.0]}]{a}{v}
            \edge[-{Latex[scale=1.0]}]{v}{x}
            \edge[-{Latex[scale=1.0]}]{b}{x}
            \edge[-{Latex[scale=1.0]}]{a}{b}

           \draw [-{Latex[scale=1.0]}] (b) to [out=135,in=45] (v);

            \draw [-{Latex[scale=1.0]}] (um) to [out=90,in=-90] (m);
            \draw [-{Latex[scale=1.0]}] (us) to [out=90,in=-90] (s);
            \draw [-{Latex[scale=1.0]}] (ub) to [out=-90,in=90] (b);
            \draw [-{Latex[scale=1.0]}] (uv) to [out=-90,in=90] (v);
            \draw [-{Latex[scale=1.0]}] (ua) to [out=-90,in=90] (a);
            \draw [-{Latex[scale=1.0]}] (ux) to (x);
            \draw [-{Latex[scale=1.0]}] (zx) to (x);
        \end{tikzpicture}
        \caption{Deep SCM for UK Biobank. MRI Seq. $(m)$, age $(a)$, sex $(s)$, brain $(b)$ \& ventricle $(v)$ volume.}
        \label{scm_brains}
    \end{subfigure}
    \caption{Brain MRI counterfactuals from our deep SCM. Direct causal effect is shown (red: increase; blue: decrease).
    We observe qualitatively sharp $192{\times}192$ counterfactuals that manifest localized interventional changes according to the associated causal graph in (a), whilst preserving the identity of the observation. Counterfactual uncertainty is also shown.}
    \label{fig:brains}
    \hfill
\end{figure*}
\begin{figure*}[ht]
    \centering
    \hfill
    \begin{subfigure}{.72\textwidth}
        \centering
        \includegraphics[trim={0 5 0 0}, clip, width=\textwidth]{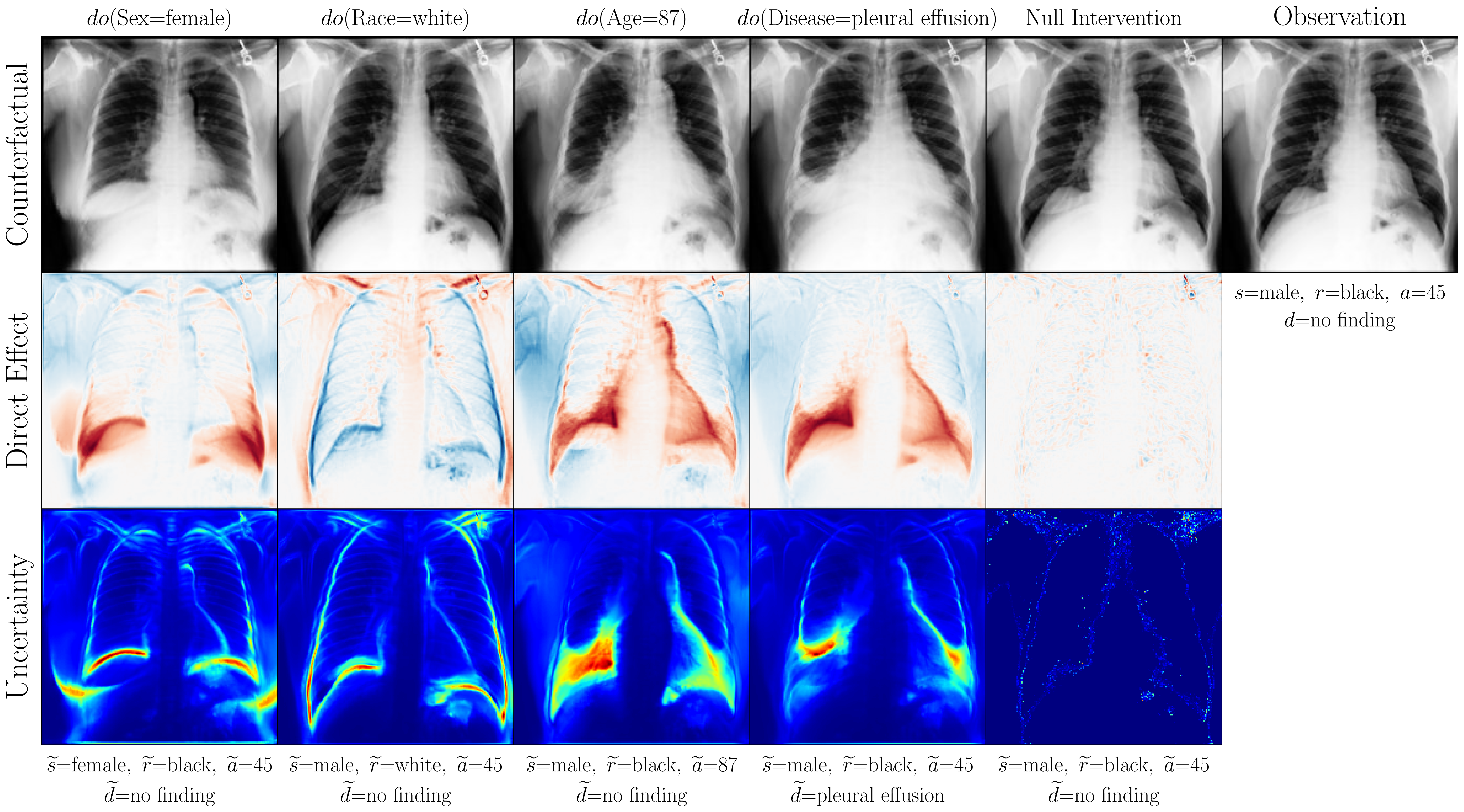}
    \end{subfigure}
    \hfill
    \begin{subfigure}{.25\textwidth}
        \centering
        \begin{tikzpicture}[thick]
            \node[obs] (x) {$\mathbf{x}$};
            \node[obs, above=18pt of x] (d) {$d$};
            \node[obs, right=18pt of d] (r) {$r$};
            \node[obs, left=18pt of d] (a) {$a$};
            \node[obs, right=18pt of x] (s) {$s$};

            \node[latent, xshift=0pt, left=18pt of x] (zx) {$\mathbf{z}_{1:L}$};
            \node[latent, xshift=0pt, below=15pt of x] (ux) {$\boldsymbol{\epsilon}$};
            \node[latent, below=15pt of s] (us) {$\mathbf{U}_s$};
            \node[latent, above=15pt of r] (ur) {$\mathbf{U}_r$};
            \node[latent, above=15pt of d] (ud) {$\mathbf{U}_d$};
            \node[latent, above=15pt of a] (ua) {$\mathbf{U}_a$};

            \edge[-{Latex[scale=1.0]}]{a}{x}
            \edge[-{Latex[scale=1.0]}]{a}{d}
            \edge[-{Latex[scale=1.0]}]{d}{x}
            \edge[-{Latex[scale=1.0]}]{r}{x}
            \edge[-{Latex[scale=1.0]}]{s}{x}

            \draw [-{Latex[scale=1.0]}] (us) to [out=90,in=-90] (s);
            \draw [-{Latex[scale=1.0]}] (ua) to [out=-90,in=90] (a);
            \draw [-{Latex[scale=1.0]}] (ur) to [out=-90,in=90] (r);
            \draw [-{Latex[scale=1.0]}] (ud) to [out=-90,in=90] (d);
            \draw [-{Latex[scale=1.0]}] (ux) to (x);
            \draw [-{Latex[scale=1.0]}] (zx) to (x);
        \end{tikzpicture}
        \caption{Deep SCM for MIMIC-CXR. The variables in the causal graph are: age $(a)$, sex $(s)$, race $(r)$, disease $(d)$ and chest x-ray $(\mathbf{x})$. The disease $d$ is pleural effusion.}
        \label{scm_chest}
    \end{subfigure}
    \caption{Chest X-ray counterfactuals from our deep SCM. Direct causal effect is shown (red: increase; blue: decrease). We observe qualitatively sharp $192{\times}192$ counterfactuals that manifest localized interventional changes according to the associated causal graph in (a), whilst preserving the identity of the observation. Counterfactual uncertainty is also shown.}
    \label{fig:chest}
\end{figure*}
\section{Experiments}
\label{sec:experiments}
We present 3 case studies on counterfactual inference of high-dimensional structured variables\footnote{\scriptsize\url{https://github.com/biomedia-mira/causal-gen}}. To quantitatively evaluate our deep SCMs, we measure \textit{effectiveness} and \textit{composition}, which are axiomatic properties of counterfactuals that hold true in all causal models~\cite{pearl2009causality,monteiro2023measuring}. Effectiveness is measured via the anticausal parent predictors from Section~\ref{subsec:ignored}, and composition is measured via the distortion of $\mathbf{x}$'s HVAE mechanism upon null-interventions. Please refer to Appendix~\ref{sec:Counterfactual Evaluation} for more details.
\subsection{Causal Mediation on Morpho-MNIST}
\label{subsec:mnist}
Causal mediation studies the extent to which the effect of a treatment is mediated by another variable in order to help explain why/how individuals respond to certain stimulus~\cite{imai2010identification}. To demonstrate this concept on structured variables, we extend the causal modelling scenario presented by~\citet{pawlowski2020deep} using the Morpho-MNIST~\cite{castro2019morpho} dataset. The dataset is generated from a known causal graph shown in Figure~\ref{mediator_morphomnist} and Appendix~\ref{app:Morpho-MNIST}, where we introduced an additional digit class variable $y$ to study discrete counterfactuals. 
We use normalizing flows to model the causal mechanisms of variables $t$, $i$ and $y$ as in Appendix~\ref{appendix: Invertible Mechanisms for Attributes}, and use the proposed HVAE-based mechanisms for $\mathbf{x}$. Figure~\ref{mediator_morphomnist} demonstrates our latent mediator model's ability to estimate the \textit{direct}, \textit{indirect} and \textit{total} causal effects of interventions. Notably, direct effect counterfactuals preserve the identity and modify only the parents $\mathbf{pa}_\mathbf{x}$, whereas indirect effect (\textit{cross-world}) counterfactuals preserve $\mathbf{pa}_{\mathbf{x}}$ whilst changing the style according to the counterfactual mediator $\widetilde{\mathbf{z}}_{1:L}$ we would have observed had $\mathbf{pa}_{\mathbf{x}}$ been $\widetilde{\mathbf{pa}}_{\mathbf{x}}$. Our total effect counterfactuals are a combination of direct and indirect effects, which agrees with causal mediation theory~\cite{robins1992identifiability,10.5555/2074022.2074073}.

Since the generative process is known, we can measure the quality of our counterfactual approximations using the ground truth mechanisms. For variable $y$, we used an accurate digit classifier with ${\simeq}99.5\%$ test acc. instead. Table~\ref{tab:morphomnist} reports counterfactual evaluation results from random interventions on each parent. We find that our exogenous prior and latent mediator HVAE $\mathbf{x}$ mechanisms perform similarly, and both outperform baselines~\cite{pawlowski2020deep} by a wide margin especially on digit (discrete) counterfactuals which are more challenging. Total effect counterfactuals ($\text{MED}_{\text{TE}}$) are generally more faithful to counterfactual conditioning than direct effect counterparts ($\text{MED}_{\text{DE}}$) but are more likely to deviate from the identity of observations.
\subsection{Brain Imaging Counterfactuals}
\label{subsec:brains}
To demonstrate our model's ability to produce faithful high-fidelity counterfactuals of real data, we extend our approach to a real-world scenario involving brain MRI scans from the UK Biobank~\cite{sudlow2015uk}. As before, we start with an assumed causal generative process involving a set of observed variables as shown in Figure~\ref{scm_brains}. The causal graph is medically informed and extends the scenario in~\citet{pawlowski2020deep} by: (i) introducing an additional MRI Sequence (T1/T2) binary variable $m$ to enable discrete counterfactuals; (ii) having $s\rightarrow \mathbf{x}$ directly. We used a scaled-up version of our exogenous prior HVAE as $\mathbf{x}$'s mechanism and used (conditional) normalizing flows for the other mechanisms (see Appendix~\ref{appendix: Invertible Mechanisms for Attributes}). As shown in Figure~\ref{fig:brains}, our deep SCM is capable of producing qualitatively sharp $192{\times}192$ counterfactuals with localised changes according to the intervened upon parent(s) and the associated causal graph. Importantly, the identity of subjects is well preserved in all cases including null-interventions (i.e. $do(\cdot)$ nothing). Table~\ref{tab:brains} shows the counterfactual effectiveness results from random interventions on each variable. We observed satisfactory initial counterfactual effectiveness and significant improvements of post counterfactual training, demonstrating the merit of the proposed approach. Please refer to Appendix~\ref{appsec:distribution over causal worlds} for notes on abduction uncertainty and~\ref{app:brains} for additional results.

\subsection{Chest X-ray Imaging Counterfactuals}
\label{sec:lung_image}
We further extend the proposed approach to the MIMIC-CXR dataset~\cite{johnson2019mimic} to demonstrate our model's ability to estimate high-fidelity counterfactuals of real chest X-ray images. This is motivated by the need for a better understanding of algorithmic bias and reported subgroup disparities \cite{bernhardt2022potential}. We begin with an assumed causal generative process of data involving the following observed variables: age ($a$), sex ($s$), race ($r$), disease ($d$), and chest X-ray image ($\mathbf{x}$). Notably, we assume that age $a$ causes disease $d$ (pleural effusion) which requires inference of \textit{discrete} counterfactuals from upstream interventions on age. For details on computing discrete counterfactuals and other experiments, please refer to Appendix~\ref{app:chest}. Following the general setup in Section~\ref{subsec:brains}, we used a scaled-up version of our exogenous prior HVAE for $\mathbf{x}$'s mechanism. We trained for relatively few iterations on MIMIC-CXR; ${\simeq}210$K.
The quantitative counterfactual evaluation results from random interventions on each variable are reported in Table~\ref{tab:brains}. We observed significant improvements in counterfactual \textit{effectiveness} post counterfactual training, particularly for \textit{race}, \textit{age} and \textit{disease} attributes. For extensive visual evaluation results please refer to Appendix~\ref{app:chest extra results}.
\begin{table*}[!t]
    \centering
    \footnotesize
    \caption{Quantitative evaluation of generated test set counterfactuals for UK Biobank (Top) and MIMIC-CXR (Bottom). Random interventions on each variable were performed and counterfactual \textit{effectiveness} was measured via attribute predictors. `mixed' denotes also randomizing the intervened upon variable. Results in brackets $(\uparrow \cdot), (\downarrow \cdot)$ are the absolute differences in \textit{effectiveness} between regular training and with fine-tuning $\mathbf{x}$'s mechanism using counterfactual training (see Appendix~\ref{app:Anticausal Predictors}).
    }
    \label{tab:brains}
    \begin{tabular}{ccccccc}
        \toprule
        \textsc{Intervention} & \textsc{Sex} $(s)$ & \textsc{MRI} $(m)$ & \textsc{Age} $(a)$ & \textsc{Brain Vol.} $(b)$ & \textsc{Ventricle Vol.} $(v)$ \\[2pt]
        \scriptsize{(UK Biobank)} & ROCAUC $\uparrow$ & ROCAUC $\uparrow$ & MAE (years) $\downarrow$ & MAE (ml) $\downarrow$ & MAE (ml) $\downarrow$ \\
        \midrule
        $do(s)$ & 0.9905 ({\color{green}$\uparrow$} \textbf{0.172}) & 1.0 (-) & 4.849 ({\color{green}$\downarrow$} 0.018) & 24.55 ({\color{green}$\downarrow$} 24.5) & 1.854 ({\color{green}$\downarrow$} 0.322) \\
        $do(m)$ & 0.9893 ({\color{green}$\uparrow$} 0.023) & 1.0 (-) & 4.846 (-) & 26.14 ({\color{green}$\downarrow$} 1.88) & 1.932 ({\color{green}$\downarrow$} 0.092) \\
        $do(a)$ & 0.9892 ({\color{green}$\uparrow$} 0.016) & 1.0 (-) & 4.937 ({\color{red}$\uparrow$} \textbf{0.004}) & 26.24 ({\color{green}$\downarrow$} 7.31) & 1.890 ({\color{green}$\downarrow$} 0.451) \\
        $do(b)$ & 0.9944 ({\color{green}$\uparrow$} 0.069) & 1.0 (-) & 5.059 ({\color{green}$\downarrow$} 0.032) & 25.49 ({\color{green}$\downarrow$} \textbf{38.6}) & 1.846 ({\color{green}$\downarrow$} 0.933) \\
        $do(v)$ & 0.9893 ({\color{green}$\uparrow$} 0.031) & 1.0 (-) & 6.045 ({\color{green}$\downarrow$} 0.102) & 25.69 ({\color{green}$\downarrow$} 3.22) & 1.826 ({\color{green}$\downarrow$} \textbf{2.115}) \\
        mixed & 0.9899 ({\color{green}$\uparrow$} 0.061) & 1.0 (-) & 5.128 ({\color{green}$\downarrow$} 0.046) & 25.41 ({\color{green}$\downarrow$} 15.1) & 1.900 ({\color{green}$\downarrow$} 0.822) \\
        \bottomrule
    \end{tabular}
\end{table*}
\begin{table*}[!t]
    \centering
    \footnotesize
    \label{tab:chest}
    \begin{tabular}{cccccc}
        \toprule
        \textsc{Intervention} & \textsc{Sex} $(s)$ & \textsc{Race} $(r)$ & \textsc{Age} $(a)$ & \textsc{Disease} $(d)$ \\[2pt]
        \scriptsize{\text{(MIMIC-CXR)}} & ROCAUC $\uparrow$ & ROCAUC $\uparrow$ & MAE (years) $\downarrow$ & ROCAUC $\uparrow$ \\
        \midrule
        $do(s)$ & 1.000 ({\color{green}$\uparrow$}\textbf{0.078}) & 0.839 ({\color{green}$\uparrow$}0.094)
        & 6.485 ({\color{green}$\downarrow$}0.198) 
        & 0.969 ({\color{green}$\uparrow$}0.038)\\
        $do(r)$
        &0.997 ({\color{green}$\uparrow$}0.002) 
        & 0.867 ({\color{green}$\uparrow$}\textbf{0.283}) 
        & 6.311 ({\color{green}$\downarrow$}0.115) 
        & 0.874 ({\color{green}$\uparrow$}0.008) \\
        $do(a)$ 
        & 0.997 ({\color{green}$\uparrow$}0.002) 
        & 0.807 ({\color{green}$\uparrow$}0.058)
        &  6.643
({\color{green}$\downarrow$}\textbf{3.426}) 
        & 0.916 
        ({\color{green}$\uparrow$}0.033)
        \\
        $do(d)$ 
        &  0.997 ({\color{green}$\uparrow$}0.001) 
        &  0.793
        ({\color{green}$\uparrow$}0.041) 
        & 6.568 ({\color{green}$\downarrow$}0.189) 
        & 0.982 
        ({\color{green}$\uparrow$}\textbf{0.258})
        \\
        mixed 
        &  0.998 ({\color{green}$\uparrow$}0.015)
        &  0.828
        ({\color{green}$\uparrow$}0.116)
        &  6.497
        ({\color{green}$\downarrow$}0.866)
        &  0.950 
         ({\color{green}$\uparrow$}0.076)
        \\
        \bottomrule
    \end{tabular}
\end{table*}
\section{Related Work}
Our work bolsters an ongoing effort to combine representation learning and causality~\cite{Bengio2013,scholkopf2021toward}. Causal representation learning is also closely linked to \textit{disentanglement}, where the goal is to uncover the true underlying (disentangled) generative factors of data~\cite{Higgins2017,locatello2019challenging,Kim2018,chen2018isolating}. 
Generative models such as VAEs~\cite{kingma2013auto}, GANs~\cite{goodfellow2020generative}, Normalizing Flows~\cite{tabak2010density,rezende2015variational} and Diffusion models~\cite{sohl2015deep,ho2020denoising,song2021scorebased} have become indispensable tools for causal representation learning. They have been leveraged for causal effect estimation~\cite{louizos2017causal,kocaoglu2017causalgan,tran2017implicit}, causal discovery~\cite{yang2021causalvae,sanchez2022diffusion,geffner2022deep}, and various other extensions have enabled modelling of conditional~\cite{trippe2018conditional,mirza2014conditional,sohn2015learning,dhariwal2021diffusion} and interventional distributions~\cite{kocaoglu2018causalgan,ke2019learning,xia2021causal,zevcevic2021relating}. However, few works have focused on fulfilling all three rungs of Pearl's \textit{ladder of causation}: (i) association; (ii) intervention; (iii) counterfactuals~\cite{pearl2009causality,bareinboim2022pearl} in a principled manner using deep models.

Our work is most closely related to DSCMs~\cite{pawlowski2020deep} and NCMs~\cite{xia2021causal, xia2023neural} in that we leverage deep learning components to learn causal mechanisms. However, our focus is on the practical estimation and evaluation of plausible high-fidelity image counterfactuals, whereas previous work mostly focused on theoretical and/or proof-of-concept low-resolution settings. \citet{sanchez2021diffusion,sanchez2022healthy} proposed Diffusion SCMs (Diff-SCMs) for high-fidelity counterfactuals, but considered only two-variable causal models. Our approach is inspired by recent identifiability results in deep generative models~\cite{khemakhem2020variational,hyvarinen2019nonlinear}, as well as modern HVAE architectures~\cite{vahdat2020nvae,child2020very} which \textit{are} readily amenable to explicit, identity-preserving abduction. Causal mediation analysis concepts like direct, indirect and total effects~\cite{robins1992identifiability,imai2010identification,10.5555/2074022.2074073} also guided our latent mediator SCM setup. Many image-to-image translation approaches~\cite{Isola2017,liu2017unsupervised,su2022dual,saharia2022palette,brooks2022instructpix2pix,preechakul2022diffusion} are also related to counterfactual inference, but only in an informal sense as they do not explicitly perform abduction, model interventions, nor use causal structure.
\section{Conclusion}
\label{sec:conclusion}
We present a pragmatic causal generative modelling framework for estimating high-fidelity image counterfactuals using deep SCMs. Our proposed deep causal mechanisms are inspired by recent identifiability results for deep generative models, as well as \textit{causal mediation analysis} theory. We show how to plausibly estimate \textit{direct}, \textit{indirect}, and \textit{total} causal effects on high-dimensional structured variables such as images, and provide \textit{abduction} uncertainty estimates. We quantify the \textit{soundness} of our counterfactuals by evaluating axiomatic properties that hold true in all causal models: i.e. \textit{effectiveness} and \textit{composition}. We believe the ability to generate plausible counterfactuals could greatly benefit several important areas: (i) \textit{explainability}, e.g. through causal mediation effects as studied here; (ii) \textit{data augmentation}, e.g. mitigating data scarcity and underrepresentation of subgroups; (iii) \textit{robustness}, to e.g. spurious correlations. Our work contributes primarily to the empirical and theoretical advancement of counterfactual inference models -- valuable extensions for future work include demonstrating the advantage of using counterfactuals in the aforementioned areas.

\paragraph{Limitations.} This work considers only Markovian SCMs, wherein all causal effects are identifiable from observed data under the assumption of no unobserved confounding. Markovianity is a common assumption in academic literature but may be too restrictive in some real-world scenarios. We take a pragmatic empirical approach to counterfactual evaluation by measuring their \textit{axiomatic soundness} rather than being bound by a lack of theoretical identifiability in the limit of infinite data. Nonetheless, extensions to Semi- and/or Non-Markovian settings would boost the practicality of our approach, but this is highly non-trivial for structured variables. Further, we stress that any conclusions drawn using our approach are strictly dependent on the correctness of the assumed SCM. We urge practitioners to carefully consider the ethical implications of their modelling assumptions when applying this framework in real-world settings.

\paragraph{Acknowledgements.} This project has received funding from the ERC under the EU's Horizon 2020 research and innovation programme (grant No. 757173). B.G. is grateful for the support from the Royal Academy of Engineering as part of his Kheiron Medical Technologies / RAEng Research Chair in Safe Deployment of Medical Imaging AI.




\bibliography{bibfile}
\bibliographystyle{icml2023}

\onecolumn
\newpage
\twocolumn
\appendix
\onecolumn
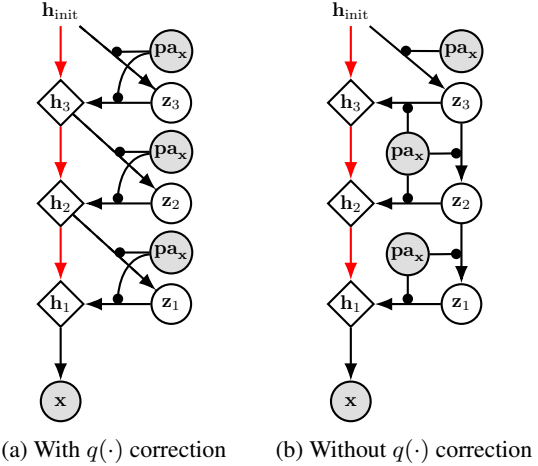
\begin{wrapfigure}[25]{r}{0.45\textwidth}
    \vspace{-23pt}
    \centering
    \hfill
    \begin{subfigure}{.22\textwidth}
        \centering
        \begin{tikzpicture}[thick,scale=0.75, every node/.style={scale=0.75}]
\node[] (hl) {$\mathbf{h}_\mathrm{init}$};
        \node[det, below=20pt of hl] (h3) {$\mathbf{h}_3$};
        \node[det, below=20pt of h3] (h2) {$\mathbf{h}_2$};
        \node[det, below=20pt of h2] (h1) {$\mathbf{h}_1$};
        \node[obs, below=20pt of h1] (x) {$\mathbf{x}$};
        
        \node[latent, right=25pt of h1] (z1) {$\mathbf{z}_1$};
        \node[latent, right=25pt of h2] (z2) {$\mathbf{z}_2$};
        \node[latent, right=25pt of h3] (z3) {$\mathbf{z}_3$};
    
        \edge[-{Latex[scale=1.0]}]{h1}{x}
        \edge[-{Latex[scale=1.0]}, red]{h2}{h1}
        \edge[-{Latex[scale=1.0]}, red]{h3}{h2}
        \edge[-{Latex[scale=1.0]}, red]{hl}{h3}
        \edge[-{Latex[scale=1.0]}]{hl}{z3}

        \draw[-{Latex[scale=1.0]}] (z3) -- (h3) coordinate[midway, xshift=0pt, yshift=0pt] (z3h3);
        \draw[-{Latex[scale=1.0]}] (z2) -- (h2) coordinate[midway, xshift=0pt, yshift=0pt] (z2h2);
        \draw[-{Latex[scale=1.0]}] (z1) -- (h1) coordinate[midway, xshift=0pt, yshift=0pt] (z1h1);
        
        \edge[-{Latex[scale=1.0]}]{h3}{z2}
        \edge[-{Latex[scale=1.0]}]{h2}{z1}
    
        \draw[dashed] (hl) -- (z3) coordinate[midway, xshift=-3pt, yshift=3pt] (hlz3);
        \node[obs, above=3.5pt of z3, xshift=0pt] (paz3) {$\mathbf{pa}_\mathbf{x}$};
        \draw[-{Circle[scale=.8]}] (paz3) -- (hlz3);
        \draw[-{Circle[scale=.8]}] (paz3) edge[out=185, in=90] (z3h3);
        
        \draw[dashed] (h3) -- (z2) coordinate[midway] (h3z2);
        \node[obs, above=3.5pt of z2, xshift=0pt] (paz2) {$\mathbf{pa}_\mathbf{x}$};
        \draw[-{Circle[scale=.8]}] (paz2) -- (h3z2);
        \draw[-{Circle[scale=.8]}] (paz2) edge[out=185, in=90] (z2h2);
    
        \draw[dashed] (h2) -- (z1) coordinate[midway] (h2z1);
        \node[obs, above=3.5pt of z1] (paz1) {$\mathbf{pa}_\mathbf{x}$};
        \draw[-{Circle[scale=.8]}] (paz1) -- (h2z1);
        \draw[-{Circle[scale=.8]}] (paz1) edge[out=185, in=90] (z1h1);
        \end{tikzpicture}
    \caption{With $q(\cdot)$ correction}
    \end{subfigure}
    \hfill
    \begin{subfigure}{.22\textwidth}
        \centering
        \begin{tikzpicture}[thick,scale=0.75, every node/.style={scale=0.75}]
        \node[] (hl) {$\mathbf{h}_\mathrm{init}$};
        \node[det, below=20pt of hl] (h3) {$\mathbf{h}_3$};
        \node[det, below=20pt of h3] (h2) {$\mathbf{h}_2$};
        \node[det, below=20pt of h2] (h1) {$\mathbf{h}_1$};
        \node[obs, below=20pt of h1] (x) {$\mathbf{x}$};
        
        \node[latent, right=25pt of h1] (z1) {$\mathbf{z}_1$};
        \node[latent, right=25pt of h2] (z2) {$\mathbf{z}_2$};
        \node[latent, right=25pt of h3] (z3) {$\mathbf{z}_3$};
    
        \edge[-{Latex[scale=1.0]}]{h1}{x}
        \edge[-{Latex[scale=1.0]}, red]{h2}{h1}
        \edge[-{Latex[scale=1.0]}, red]{h3}{h2}
        \edge[-{Latex[scale=1.0]}, red]{hl}{h3}

        \edge[-{Latex[scale=1.0]}]{z3}{h3}
        \edge[-{Latex[scale=1.0]}]{z2}{h2}
        \edge[-{Latex[scale=1.0]}]{z1}{h1}

        \draw[-{Latex[scale=1.0]}] (z3) -- (z2) coordinate[midway, xshift=0pt, yshift=0pt] (z3z2);
        \draw[-{Latex[scale=1.0]}] (z2) -- (z1) coordinate[midway, xshift=0pt, yshift=0pt] (z2z1);
        
        \draw[-{Latex[scale=1.0]}] (hl) -- (z3) coordinate[midway, xshift=-3.5pt, yshift=3.5pt] (hlz3);
        \node[obs, above=3.5pt of z3, xshift=0pt] (paz3) {$\mathbf{pa}_\mathbf{x}$};
        \draw[-{Circle[scale=.8]}] (paz3) -- (hlz3);

        \node[obs, below=11pt of z3h3] (pa2) {$\mathbf{pa}_\mathbf{x}$};
        
        \draw[-{Circle[scale=.8]}] (pa2) edge[out=90, in=-90] (z3h3);
        \draw[-{Circle[scale=.8]}] (pa2) edge[out=-90, in=90] (z2h2);
        \node[obs, below=11pt of z2h2] (pa1) {$\mathbf{pa}_\mathbf{x}$};
        \draw[-{Circle[scale=.8]}] (pa1) edge[out=270, in=90] (z1h1);

        \draw[-{Circle[scale=.8]}] (pa2) edge[out=0, in=0] (z3z2);
        \draw[-{Circle[scale=.8]}] (pa1) edge[out=0, in=0] (z2z1);
        \end{tikzpicture}
        \caption{Without $q(\cdot)$ correction}
        \label{chvae_c2}
    \end{subfigure}
    \hfill
    \caption{Conditional HVAE generative model structures for the proposed latent mediator model. The associated inference model is the same as in Figure~\ref{inf_model}. (a) Like the ladder VAE~\cite{sonderby2016ladder}, each conditional prior distribution $p_\theta(\mathbf{z}_i \mid \mathbf{z}_{>i})$ at each layer $i$ is corrected by a data-dependent likelihood contribution from the posterior of the layer above via $\mathbf{h}_{i+1}$. 
    Unlike (a), the conditional prior distributions at each layer of (b) do \textit{not} depend on the posterior, but each posterior still depends on the prior from the layer above.}
    \label{chvae2}
\end{wrapfigure}

\section{Supplementary Methods}
\label{app:Supplementary Methods}
\subsection{Invertible Mechanisms for Attributes}
\label{appendix: Invertible Mechanisms for Attributes}
Attributes which are \textit{ancestors} of the image $\mathbf{x}$, $\mathbf{an}_{\mathbf{x}}= \{a_1,\dots,a_{N-1}\} \supseteq \mathbf{pa}_\mathbf{x}$, are generally \textit{not} assumed to be independent, so we learn their structural assignments from data. To enable tractable abduction for $\mathbf{an}_\mathbf{x}$, we learn invertible mechanisms $f_k$ using conditional normalizing flows~\cite{trippe2018conditional} as suggested by \citet{pawlowski2020deep}. Each attribute's mechanism $a_k \coloneqq f_k(u_k; \mathbf{pa}_k)$ is a conditional flow, where $a_k$ is expressed as a parameterised function of $\mathbf{pa}_k$ and samples from a base distribution $u_k \sim p_u(u_k)$. The conditional density is given by 
\begin{align}
    p(a_k \mid \mathbf{pa}_k) = p_{u}(u_k)
    \left|\mathrm{det}J_{f_k}(u_k)
    \right|^{-1},
\end{align}
where $u_k = f_k^{-1}(a_k;\mathbf{pa}_k)$, and $J_{f_k}(u_k)$ is the Jacobian matrix of all partial derivatives of $f_k$ with respect to $u_k$. The base distribution for the exogenous noise is typically assumed to be Gaussian, which may be restrictive. Moreover, we note that $u_k$ here is not strictly latent (unobserved) as described in SCM theory, since knowing $(a_k, \mathbf{pa}_k)$ uniquely determines $u_k$. A counterfactual attribute $\widetilde{a}_k$ is given by forwarding the mechanism using its counterfactual parents and the abducted exogenous noise: $\widetilde{a}_k \coloneqq f_k(u_k; \widetilde{\mathbf{pa}}_k)$. In practice, we use standard Gaussians as base distributions for the exogenous noise and leverage available PyTorch~\cite{paszke2019pytorch} \& Pyro~\cite{bingham2019pyro} implementations.
\begin{wrapfigure}[14]{r}{0.35\textwidth}
    \vspace{-20pt}
    \centering
    \includegraphics[width=.35\textwidth]{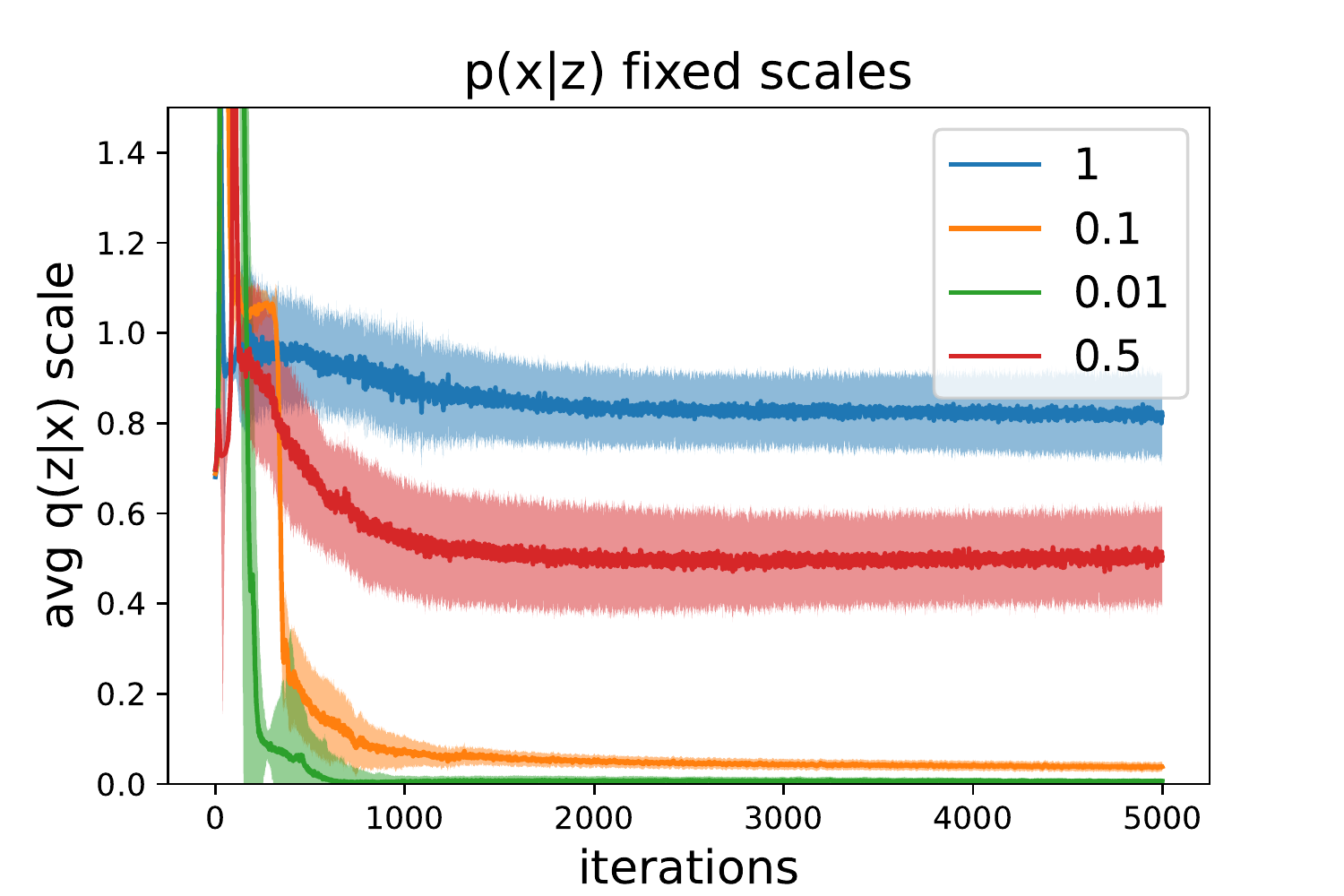}
    \caption{Showing a (1 stochastic layer) VAE collapsing to a deterministic AE throughout training on Morpho-MNIST (3 random seeds used). We observe that the marginal posterior scale tends towards zero when the image (Gaussian) likelihood scale is fixed in the near-deterministic regime. This observation is consistent with the VAE \textit{self-consistency} claims of~\citet{nielsen2020survae} and \citet{NEURIPS2022_4eb91efe}.
    }
    \label{fig:scales}
\end{wrapfigure}
\subsection{Distribution over Causal Worlds}
\label{appsec:distribution over causal worlds}
As explained in Section~\ref{subsec:Deep Mechanisms for Structured Variables}, the original DSCM~\cite{pawlowski2020deep} framework's VAE-based causal mechanism for $\mathbf{x}$ was trained in the near-deterministic regime, thereby incidentally attempting to deterministically abduct $\mathbf{x}$'s exogenous like a normalizing flow. Consequently, the model struggled to: (i) produce realistic random samples from the SCM; (ii) represent abduction uncertainty; (iii) induce a distribution over causal worlds. Our proposed HVAE-based deep causal mechanisms address these issues. 

The counterfactual distribution of $\widetilde{\mathbf{x}}$: $P_{\mathcal{M}}(\widetilde{\mathbf{x}} \mid do(\widetilde{\mathbf{pa}}_k), \mathbf{x})$, is our distribution of interest associated with the modified probabilistic SCM $\langle \mathcal{M}_{do(\widetilde{\mathbf{pa}}_k)}, P(U | X) \rangle$, after the three-step procedure (Section~\ref{scms}). The prior and posterior distributions over the exogenous noise variables (with an exogenous prior HVAE mechanism for $\mathbf{x}$) are given by
\begin{align}
    P(U) &= \underbrace{p_\theta(\mathbf{z}_{1:L})p(\boldsymbol{\epsilon})}_{p(\mathbf{u}_\mathbf{x})}\prod_{i=1}^{N-1}p(\mathbf{u}_i),
    \\
    P(U | X) &\approx \underbrace{q_\phi(\mathbf{z}_{1:L} \mid \mathbf{x}, \mathbf{pa}_{\mathbf{x}})\delta(\boldsymbol{\epsilon} \mid \mathbf{x}, \mathbf{pa}_\mathbf{x})}_{\approx p(\mathbf{u}_\mathbf{x} \mid \mathbf{x}, \mathbf{pa}_\mathbf{x})}
    \prod_{i=1}^{N-1}\delta(\mathbf{u}_i \mid a_i, \mathbf{pa}_i). 
\end{align} 
Since abduction is non-deterministic in this model, we can sample different realizations of the exogenous variables from $P(U|X)$ by sampling from the HVAE encoder $q_\phi(\mathbf{z}_{1:L}| \mathbf{x}, \mathbf{pa}_{\mathbf{x}})$, thereby inducing a distribution over causal worlds and yielding varied counterfactuals of $\mathbf{x}$. Note that the Delta distributed $\delta(\cdot)$ exogenous variable posteriors are a result of deterministic abduction (e.g. from inverting a normalizing flow mechanism). Furthermore, we can calculate the first and second moments of the counterfactual distribution:
\begin{align}
    &\boldsymbol{\mu}_{\widetilde{\mathbf{x}}} = \mathbb{E}_{P(U|X)}\big[P_{\mathcal{M}}(\widetilde{\mathbf{x}} \mid do(\widetilde{\mathbf{pa}}_k), \mathbf{x})\big],
    \quad
    &\boldsymbol{\sigma}_{\widetilde{\mathbf{x}}} = \mathrm{Var}_{P(U|X)}\big[P_{\mathcal{M}}(\widetilde{\mathbf{x}} \mid do(\widetilde{\mathbf{pa}}_k), \mathbf{x})\big]^{\frac{1}{2}},
\end{align}
where $\boldsymbol{\mu}_{\widetilde{\mathbf{x}}}$ can be interpreted as the most likely counterfactual of $\mathbf{x}$ and $\boldsymbol{\sigma}_{\widetilde{\mathbf{x}}}$ as a measure of \textit{counterfactual uncertainty}.
%
\subsection{Latent Mediator Architectures}
\label{subsec:Latent Mediator Architectures}
As shown in Figure~\ref{chvae_c2}, we can alter the conditional generative model structure of the latent mediator model, such that the conditional prior distributions no longer receive data-dependent corrections from previous layer posteriors as in the Ladder VAE~\cite{sonderby2016ladder}. We find that this architecture (Figure~\ref{chvae_c2}) is less prone to ignored counterfactual conditioning, especially when trained with parent \textit{conditioning dropout} (see comparative results in Table~\ref{tab:morphomnist_app}). Parent conditioning dropout consists of randomly selecting when $\mathbf{pa}_{\mathbf{x}}$ is merged into the downstream. We can either drop the $\mathbf{pa}_{\mathbf{x}}$ merge connections between each $\mathbf{h}_i$ and $\mathbf{z}_i$ (deterministic path) or between each $\mathbf{z}_i$ and $\mathbf{z}_{i-1}$ (stochastic path), whilst holding the other fixed. Parent conditioning dropout is somewhat reminiscent of classifier-free guidance~\cite{ho2022classifier} in diffusion models but the application and motivations here are different; to prevent the model from prioritising one conditioning path over the other and improve counterfactual conditioning on $\widetilde{\mathbf{pa}}_{\mathbf{x}}$ in the forward model with the abducted noise $\mathbf{z}_{1:L}$ fixed.
%
%
\begin{table*}[!t]
    \centering
    \footnotesize
    \caption{Comparing counterfactual effectiveness of random interventions using our latent mediator model ($\beta{=}1$) when trained with and without conditioning dropout (CD) as described in Section~\ref{subsec:Latent Mediator Architectures}. We find that \textit{effectiveness} of digit counterfactuals improves significantly with CD at the cost of slightly worse composition (higher bits per dimension).} 
    \label{tab:morphomnist_app}
    \begin{tabular}{lcc|cccc|cccc|cccc}
        \toprule
        & & &\multicolumn{4}{c}{\textsc{Thickness} $(t)$ MAE $\downarrow$} & \multicolumn{4}{c}{\textsc{Intensity} $(i)$ MAE $\downarrow$} & \multicolumn{4}{c}{\textsc{Digit} $(y)$ Acc. (\%) $\uparrow$} \\[4pt]     
        \textsc{Method} & CD & bpd $\downarrow$ & $do(t)$ & $do(i)$ & $do(y)$ & mix & $do(t)$ & $do(i)$ & $do(y)$ & mix & $do(t)$ & $do(i)$ & $do(y)$ & mix \\        
        \midrule
        $\text{MED}_{\text{DE}}$ & N & .676 & .127 \cmt{(1e-3)} & .133 \cmt{(2e-4)} & .252 \cmt{(3e-4)} & .202 \cmt{(9e-4)} & 1.70 \cmt{(6e-3)} & 2.04 \cmt{(0.129)} & 1.85 \cmt{(8e-3)} & 2.17 \cmt{(1e-2)} & 99.30
        \cmt{(1e-3)} & 99.06 \cmt{(2e-3)} & 81.07 \cmt{(2e-4)} & 88.37 \cmt{(2e-4)}
        \\
        $\text{MED}_{\text{TE}}$ & N & .676 & .162 \cmt{(2e-3)} & .168 \cmt{(1e-3)} & .225 \cmt{(9e-4)} & .200 \cmt{(7e-4)} & 1.73 \cmt{(1e-2)} & 2.60 \cmt{(1e-2)} & 1.79 \cmt{(4e-3)} & 2.22 \cmt{(3e-3)} & 99.74 \cmt{(2e-4)} & 99.36
        \cmt{(4e-4)} & 94.28 \cmt{(1e-3)} & 95.87 \cmt{(1e-3)}
        \\
        \midrule
        $\text{MED}_{\text{DE}}$ & Y & .682 & .125 \cmt{(6e-4)} & .137 \cmt{(4e-5)} & .157 \cmt{(1e-3)} & .149 \cmt{(2e-3)} & 1.65 \cmt{(2e-2)} & 1.48 \cmt{(2e-2)} & 1.80 \cmt{(1e-2)} & 1.89 \cmt{(3e-4)} & 99.38
        \cmt{(3e-4)} & 98.73 \cmt{(6e-4)} & 99.47 \cmt{(4e-5)} & 99.09 \cmt{(3e-4)}
        \\
        $\text{MED}_{\text{TE}}$ & Y & .682 & .141 \cmt{(2e-4)} & .153 \cmt{(6e-4)} & .146 \cmt{(1e-3)} & .150 \cmt{(7e-4)} & 1.72 \cmt{(2e-2)} & 2.17 \cmt{(1e-2)} & 1.78 \cmt{(5e-3)} & 2.01 \cmt{(3e-3)} & 99.75 \cmt{(4e-4)} & 99.30
        \cmt{(5e-4)} & 99.68 \cmt{(4e-4)} & 99.41 \cmt{(3e-4)}
        \\
        \bottomrule
    \end{tabular}
\end{table*}
\section{Axiomatic Counterfactual Evaluation}
\label{sec:Counterfactual Evaluation}
In order to quantitatively evaluate our approximate counterfactual inference models, we measure the axiomatic properties of counterfactuals: (i) composition; (ii) effectiveness; (iii) reversibility~\cite{pearl2009causality,monteiro2023measuring}, which hold true in all causal models. The \textit{soundness}~\cite{Galles1998} and \textit{completeness}~\cite{Halpern1998} theorems state that composition, effectiveness and reversibility are the necessary and sufficient properties of counterfactuals in any causal model. The three axiomatic properties of counterfactuals can be summarised as follows:
\begin{enumerate}[(i)]
    \item \textbf{Composition:} Intervening on a variable to have a value it would have had without our intervention will not affect the other variables in the system;
    \item \textbf{Effectiveness:} Intervening on a variable to have a specific value will cause the variable to take on that value; 
    \item \textbf{Reversibility:} Precludes multiple solutions due to feedback loops, and follows directly from composition in recursive systems such as DAGs. Refer for~\cite{pearl2009causality} for further details on non-recursive systems.
\end{enumerate}
Following the counterfactual evaluation framework proposed by~\citet{monteiro2023measuring}, we measure counterfactual \textit{effectiveness} using a `pseudo-oracle' function's accuracy/error (i.e. calculated from our parent predictors) and measure composition via the distortion induced by $\mathbf{x}$'s mechanism from (repeated) null-interventions. In the case of a HVAE-based causal mechanism for $\mathbf{x}$, composition can be understood as reconstructing the input given observed parents, and reversibility as the act of cycling back between factual and counterfactual parent interventions. In both cases, distance metrics can be used to measure differences between counterfactual and factual images (e.g. $L1$ image distance per-pixel).
\begin{figure}
    \centering
    \includegraphics[width=\textwidth]{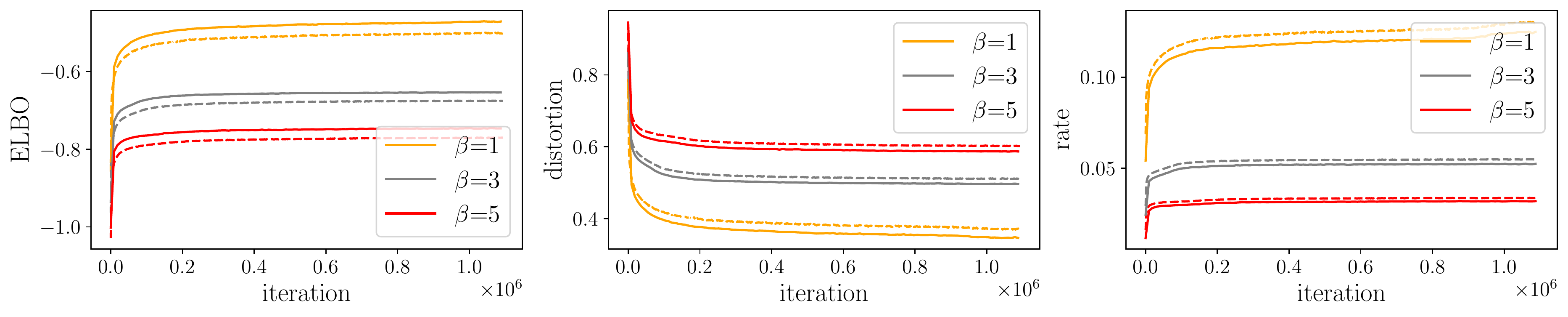}
    \includegraphics[width=\textwidth]{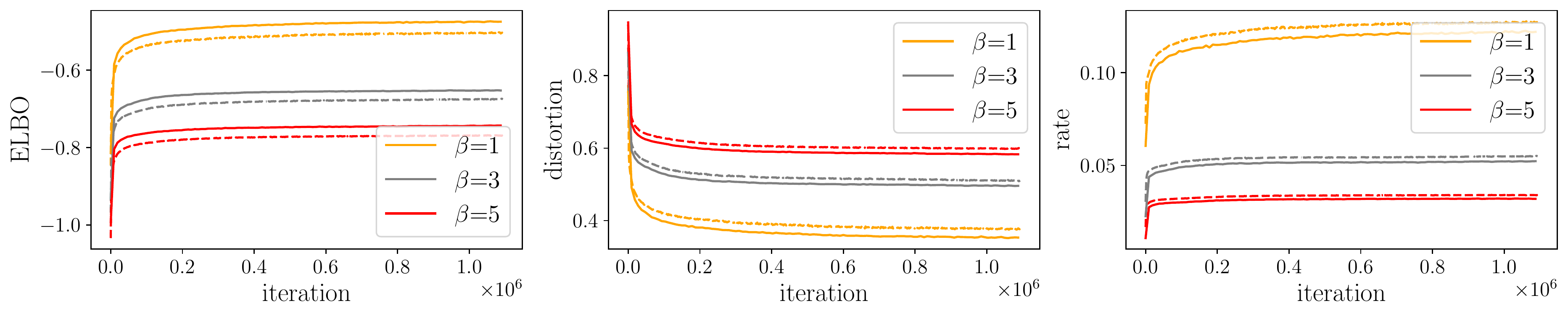}
    \caption{Evidence lower bound, distortion and rate throughout training on Morpho-MNIST (shown in nats per dimension). Top row corresponds to our exogenous prior HVAE, and the bottom row to the HVAE for our latent mediator model. Identical architectures were used in both cases, leading to similar results despite the change in conditioning structure. Dashed and solid lines correspond to training and validation respectively.}
    \label{fig:morpho_elbo}
\end{figure}
\begin{figure}
    \centering
    \includegraphics[width=\textwidth]{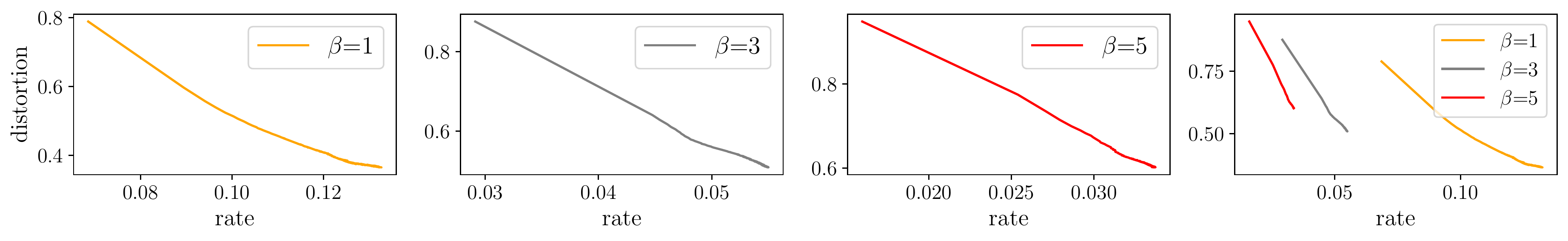}
    \includegraphics[width=\textwidth]{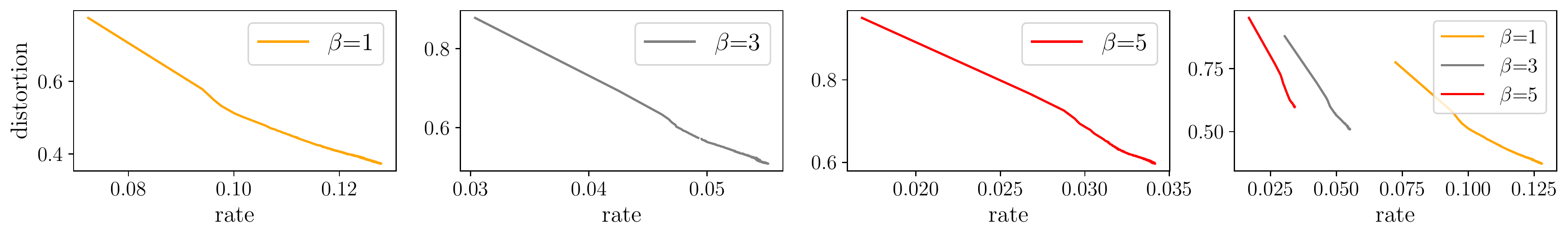}
    \caption{Rate-distortion curves for our exogenous prior HVAE (Top) and the latent mediator model's HVAE (Bottom) trained on Morpho-MNIST (shown in nats per dimension). For both models, we can see that increasing $\beta$ reduces the rate whilst increasing distortion, encouraging the model to ignore imperceptible details and focus more on semantic compression. This helps explain why increasing $\beta$ can improve counterfactual conditioning, since the model is discouraged from focusing mostly on improving distortion (image likelihood) and finding a solution $p_\theta(\mathbf{x}) = p_\theta(\mathbf{x}|c)$ by ignoring conditioning $c$.}
    \label{fig:morpho_rate_distortion}
\end{figure}
\begin{table*}[!t]
    \centering
    \footnotesize
    \caption{Additional quantitative evaluation of generated counterfactuals for the Morpho-MNIST test set. Supplementary to Table~\ref{tab:morphomnist} in the main text and includes further ablations on the $\beta$ penalty used during training.
    } 
    \label{tab:morphomnist_2}
    \begin{tabular}{lcc|cccc|cccc|cccc}
        \toprule
        & & &\multicolumn{4}{c}{\textsc{Thickness} $(t)$ MAE $\downarrow$} & \multicolumn{4}{c}{\textsc{Intensity} $(i)$ MAE $\downarrow$} & \multicolumn{4}{c}{\textsc{Digit} $(y)$ Acc. (\%) $\uparrow$} \\[4pt]     
        \textsc{Method} & $\beta$ & bpd $\downarrow$ & $do(t)$ & $do(i)$ & $do(y)$ & mix & $do(t)$ & $do(i)$ & $do(y)$ & mix & $do(t)$ & $do(i)$ & $do(y)$ & mix \\        
        \midrule
        Baseline & 3 & 2.17 & .126 \cmt{(2e-3)} & .185 \cmt{(1e-3)} & .149 \cmt{(8e-4)} & .171 \cmt{(1e-3)} & 14.1 \cmt{(4e-2)} & 15.5 \cmt{(1e-2)} & 15.1 \cmt{(8e-2)} & 15.6 \cmt{(0.13)} & 99.47 \cmt{(6e-4)} & 99.34 \cmt{(5e-4)} & 97.89 \cmt{(1e-3)} & 98.34 \cmt{(1e-3)} \\
        \ $\sim$ Prior & 3 & N/A & .174 \cmt{(9e-4)} & .222 \cmt{(1e-3)} & .173 \cmt{(2e-3)} & .201 \cmt{(1e-3)} & 15.4 \cmt{(3e-2)} & 17.1 \cmt{(0.146)} & 15.3 \cmt{(6e-2)} & 16.4 \cmt{(0.122)} & 96.21 \cmt{(1e-3)} & 96.01 \cmt{(5e-5)} & 96.39 \cmt{(1e-3)} & 96.27 \cmt{(1e-3)} \\
        \midrule
        $\text{EXO}_{\text{DE}}$ & 5 & 1.08 & .137 \cmt{(1e-3)} & .158 \cmt{(4e-4)} & .137 \cmt{(8e-4)} & .149 \cmt{(4e-4)} & 2.91 \cmt{(1e-2)} & 4.09 \cmt{(1e-2)} & 2.82 \cmt{(7e-3)} & 3.59 \cmt{(6e-2)} & 99.62
        \cmt{(2e-4)} & 99.28 \cmt{(1e-4)} & 99.82 \cmt{(3e-4)} & 99.49 \cmt{(1e-4)}
        \\
        \midrule
        $\text{MED}_{\text{DE}}$ & 5 & 1.07 & .139 \cmt{(4e-4)} & .149 \cmt{(6e-4)} & .140 \cmt{(9e-4)} & .145 \cmt{(5e-4)} & 2.66 \cmt{(1e-2)} & 4.28 \cmt{(4e-2)} & 2.61 \cmt{(2e-2)} & 3.57 \cmt{(8e-2)} & 99.61
        \cmt{(2e-4)} & 99.26 \cmt{(4e-4)} & 99.76 \cmt{(7e-4)} & 99.52 \cmt{(8e-5)}
        \\
        $\text{MED}_{\text{TE}}$ & 5 & 1.07 & .126 \cmt{(1e-3)} & .141 \cmt{(4e-4)} & .127 \cmt{(8e-4)} & .134 \cmt{(6e-4)} & 2.96 \cmt{(2e-2)} & 4.87 \cmt{(1e-2)} & 2.94 \cmt{(1e-2)} & 4.04 \cmt{(9e-3)} & 99.86 \cmt{(3e-4)} & 99.60
        \cmt{(3e-4)} & 99.82 \cmt{(2e-4)} & 99.66 \cmt{(4e-4)}
        \\
        \bottomrule
    \end{tabular}
\end{table*}
\begin{table*}[!t]
    \centering
    \footnotesize
    \caption{Quantitative evaluation of generated counterfactuals for the Morpho-MNIST test set. Sister table of results in Tables~\ref{tab:morphomnist} \&~\ref{tab:morphomnist_2} in the main text reporting the standard deviations of results from three random seed runs.
    } 
    \label{tab:morphomnist_std}
    \begin{tabular}{lcc|cccc|cccc|cccc}
        \toprule
        & & &\multicolumn{4}{c}{\textsc{Thickness} $(t)$ MAE $\downarrow$} & \multicolumn{4}{c}{\textsc{Intensity} $(i)$ MAE $\downarrow$} & \multicolumn{4}{c}{\textsc{Digit} $(y)$ Acc. (\%) $\uparrow$} \\[4pt]   
        \textsc{Method} & $\beta$ & bpd $\downarrow$ & $do(t)$ & $do(i)$ & $do(y)$ & mix & $do(t)$ & $do(i)$ & $do(y)$ & mix & $do(t)$ & $do(i)$ & $do(y)$ & mix \\        
        \midrule
        Baseline & 1 & 2.04 & 1e-3 & 1e-3 & 2e-3 & 1e-3 & 1e-3 & 2e-2 & 3e-2 & 4e-3 & 4e-4 & 8e-4 & 7e-4 & 2e-3 \\
        \ $\sim$ Prior & 1 & N/A & 3e-3 & 5e-4 & 1e-3 & 2e-3 & 5e-2 & 3e-2 & 9e-3 & 1e-2 & 3e-3 & 2e-3 & 3e-3 & 3e-3 \\
        Baseline & 3 & 2.17 & 2e-3 & 1e-3 & 8e-4 & 1e-3 & 4e-2 & 1e-2 & 8e-2 & 0.13 & 6e-4 & 5e-4 & 1e-3 & 1e-3 \\
        \ $\sim$ Prior & 3 & N/A & 9e-4 & 1e-3 & 2e-3 & 1e-3 & 3e-2 & 0.146 & 6e-2 & 0.122 & 1e-3 & 5e-5 & 1e-3 & 1e-3 \\
        \midrule
        $\text{EXO}_{\text{DE}}$ & 1 & .674 & 9e-4 & 2e-4  & 6e-4 & 2e-3 & 2e-2 & 3e-2 & 1e-2 & 4e-2 & 5e-4 & 5e-4 & 3e-4 & 3e-4
        \\
        \ $\sim$ Prior & 1 & N/A & 2e-3 & 9e-4 & 2e-3 & 8e-5 & 1e-2 & 3e-2 & 1e-2 & 9e-3 & 3e-4 & 1e-3 & 1e-3 & 3e-4 \\
        $\text{EXO}_{\text{DE}}$ & 3 & .942 & 1e-3 & 1e-3  & 2e-4 & 2e-3 & 1e-2 & 4e-2 & 6e-3 & 1e-2 & 1e-4 & 6e-4 & 3e-4 & 1e-3
        \\
        \midrule
        $\text{MED}_{\text{DE}}$ & 1 & .682 & 6e-4 & 4e-5 & 1e-3 & 2e-3 & 2e-2 & 2e-2 & 1e-2 & 3e-4 & 3e-4 & 6e-4 & 4e-5 & 3e-4
        \\
        $\text{MED}_{\text{TE}}$ & 1 & .682 & 2e-4 & 6e-4 & 1e-3 & 7e-4 & 2e-2 & 1e-2 & 5e-3 & 3e-3 & 4e-4 & 5e-4 & 4e-4 & 3e-4
        \\
        $\text{MED}_{\text{DE}}$ & 3 & .941 & 8e-4 & 2e-4 & 1e-3 & 4e-4 & 9e-3 & 2e-2 & 9e-3 & 3e-2 & 1e-4 & 2e-4 & 8e-4 & 2e-4
        \\
        $\text{MED}_{\text{TE}}$ & 3 & .941 & 3e-4 & 8e-4 & 9e-4 & 2e-3 & 1e-2 & 4e-2 & 3e-2 & 4e-2 & 4e-4 & 1e-3 & 6e-4 & 1e-3
        \\
        \bottomrule
    \end{tabular}
\end{table*}
\section{Morpho-MNIST}
\label{app:Morpho-MNIST}
\subsection{Dataset Details}
For our Morpho-MNIST experiments, we construct a similar scenario to \citet{pawlowski2020deep} in which a dataset is generated according to the following known structural causal model:
\begin{align}
    y &\coloneqq f_y(u_y), 
    \qquad &&u_y \sim \text{MNIST} 
    \\
    t &\coloneqq f_t(u_t) = 0.5 + u_t, 
    \qquad &&u_t \sim \mathrm{Gamma}(10,5) 
    \\
    i &\coloneqq f_i(t, u_i) = 191 \cdot \sigma(0.5u_i + 2t - 5), 
    \qquad &&u_i \sim \mathcal{N}(0,1) 
    \\
    \mathbf{x} &\coloneqq f_x(i,t,y,u_x) = \text{Set}_{i}(i,y,\text{Set}_{t}(t,y,u_x)), 
    \qquad &&u_x \sim \text{MNIST}, 
\end{align}
The $\text{Set}_{i}(\cdot)$ and $\text{Set}_{t}(\cdot)$ are morphological operations that act on an image and set its intensity $i$ and thickness $t$. We've introduced the categorical variable $y$ for digit class, to increase the complexity of the learning problem and extend counterfactual inference to the discrete case. The resulting dataset follows the original MNIST dataset splits.
\subsection{Experiment Setup}
Our deep SCMs are implemented in Pyro and Pytorch. Unlike \citet{pawlowski2020deep}, we train the causal mechanisms (normalizing flows) for all variables except the image $\mathbf{x}$ concurrently in Pyro, whereas $\mathbf{x}$'s causal mechanism is trained separately in Pytorch. Training $\mathbf{x}$'s HVAE mechanism separately from the flow mechanisms allows us to compare different versions of the $\mathbf{x}$'s mechanism fairly while keeping the rest of the SCM's mechanisms fixed. Once all the SCM components are trained they are combined into a single PyTorch module for counterfactual training and inference. 

\begin{wrapfigure}[22]{r}{0.4\textwidth}
    \vspace{-0pt}
    \centering
    \includegraphics[width=.4\textwidth]{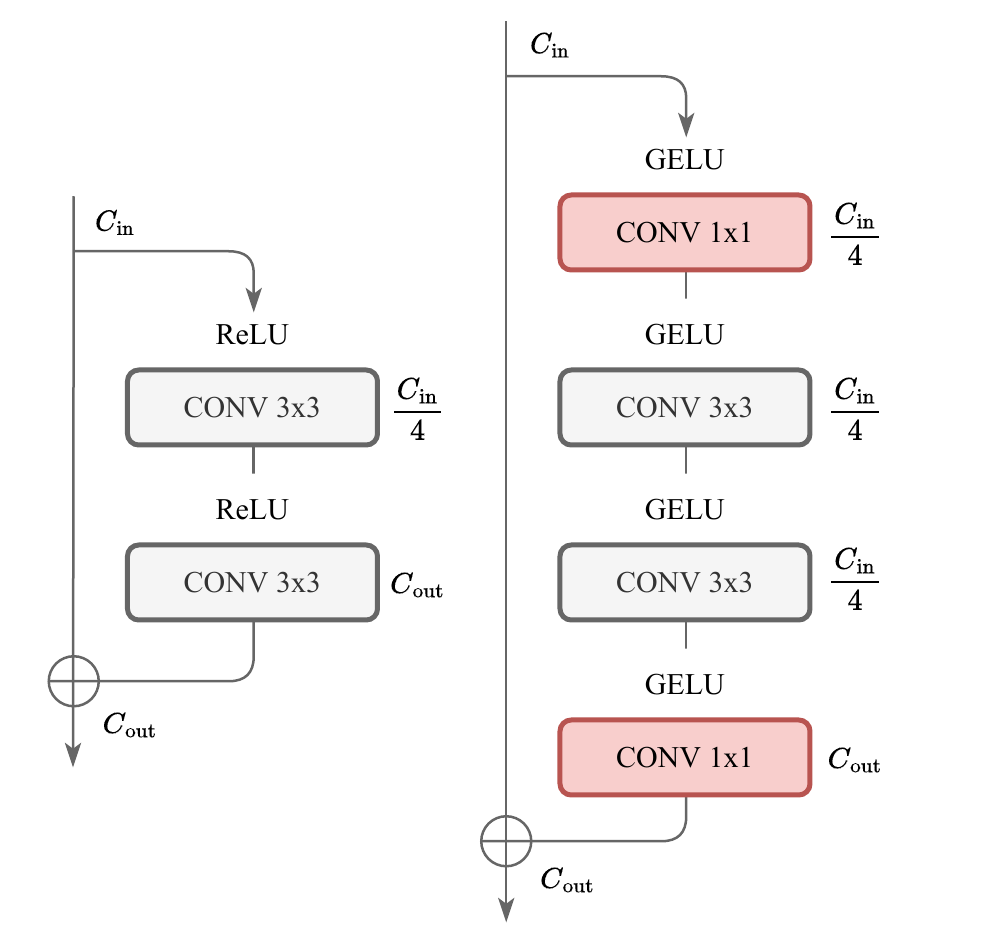}
    \caption{The residual blocks used to build our HVAEs. The block on the RHS tends to perform better but requires more GPU memory to run, so we used the LHS one for the Brain MRI dataset.}
    \label{fig:resblocks}
\end{wrapfigure}
\paragraph{Architecture.} For the experiments on the Morpho-MNIST dataset, we built upon the general setup of the very deep VAE (VDVAE) from~\citet{child2020very} and introduced structural modifications to accommodate both parent conditioning and abduction in our exogenous prior and latent mediator models described in the text. The architecture itself is largely based on the ResNet-VAE of~\cite{kingma2016improved} but contains many more layers of stochastic latent variables. The prior
and posterior are diagonal Gaussian distributions and the model is trained end-to-end by optimizing
the variational bound on the log-likelihood (ELBO)~\cite{kingma2013auto, kingma2016improved, maaloe2019biva}. Both our exogenous prior and latent mediator HVAEs for Morpho-MNIST have $20$ stochastic latent variables spanning 5 resolution scales up to the $32{\times}32$ input resolution: $\{1^2, 4^2, 8^2, 16^2, 32^2\}$. Each resolution scale contains $4$ inverted residual blocks (Figure~\ref{fig:resblocks}), and each latent variable has $16$ channels. We use variable widths per resolution of: $\{16,32,64,128,256\}$, and the total trainable parameter count is $\simeq$2M. For downsampling we use average pooling layers and for upsampling we use nearest neighbour interpolation followed by convolution. In order to condition our HVAEs, we expand and concatenate $\mathbf{pa}_\mathbf{x}$ with the latent variables $\mathbf{z}_i$ at each layer of the hierarchy in the locations specified in Figures~\ref{chvae} and~\ref{chvae2}. The resulting tensor is then merged into the downstream via a $1{\times}1$ convolution.
\paragraph{Training Details.}
We trained our HVAEs for $\simeq$1M steps using a batch size of 32 and the AdamW optimizer~\cite{loshchilov2017decoupled}. We used an initial learning rate of 1e-3 with 100 linear warmup steps, $\beta_1=0.9$, $\beta_2=0.9$ and a weight decay of 0.01. We set gradient clipping to 350 and set a gradient update skipping threshold of 500 (based on $L2$ norm). No significant training instability was observed. The final artefact is an exponential moving average of the model parameters with a rate of 0.999 which we use at inference time. For data-augmentation, we applied zero-padding of $4$ on all borders and random cropped to $32{\times}32$ resolution. Pixel intensities we rescaled to $[-1,1]$ for and validation/test images were zero-padded to $32{\times}32$.
\begin{figure}[!b]
    \centering
    \hfill
    \begin{subfigure}{.69\textwidth}
        \centering
        \includegraphics[width=\textwidth]{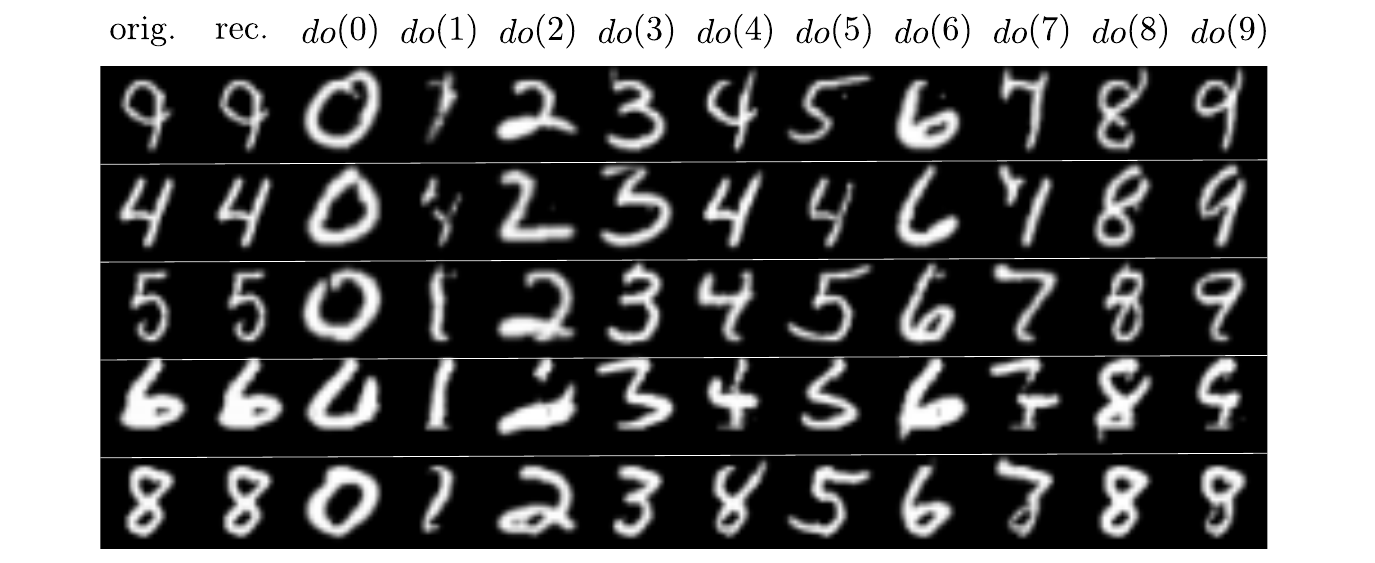}
    \end{subfigure}
    \hfill
    \begin{subfigure}{.29\textwidth}
        \centering
        \includegraphics[width=\textwidth]{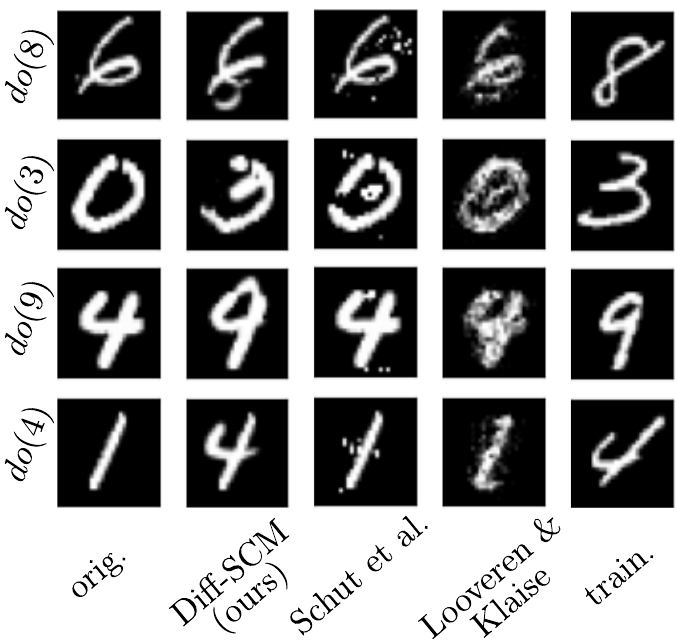}
    \end{subfigure}
    \hfill
    \caption{Figure credit belongs solely to~\citet{sanchez2021diffusion} -- reused here for qualitative comparison purposes only. From visual inspection alone, we can observe that the counterfactuals generated from our proposed deep SCM (e.g. see Appendix~\ref{app:morpho_extra}) are far superior in both counterfactual \textit{effectiveness} and \textit{composition} compared to the previous methods shown~\cite{sanchez2021diffusion,schut2021generating,van2021interpretable}.}
    \label{fig:sanchez}
\end{figure}
\newpage
\subsection{Extra Results}
\label{app:morpho_extra}
\begin{figure}[!ht]
    \centering
    \hfill
    \\[10pt]
    \begin{subfigure}{.49\textwidth}
        \includegraphics[width=\textwidth]{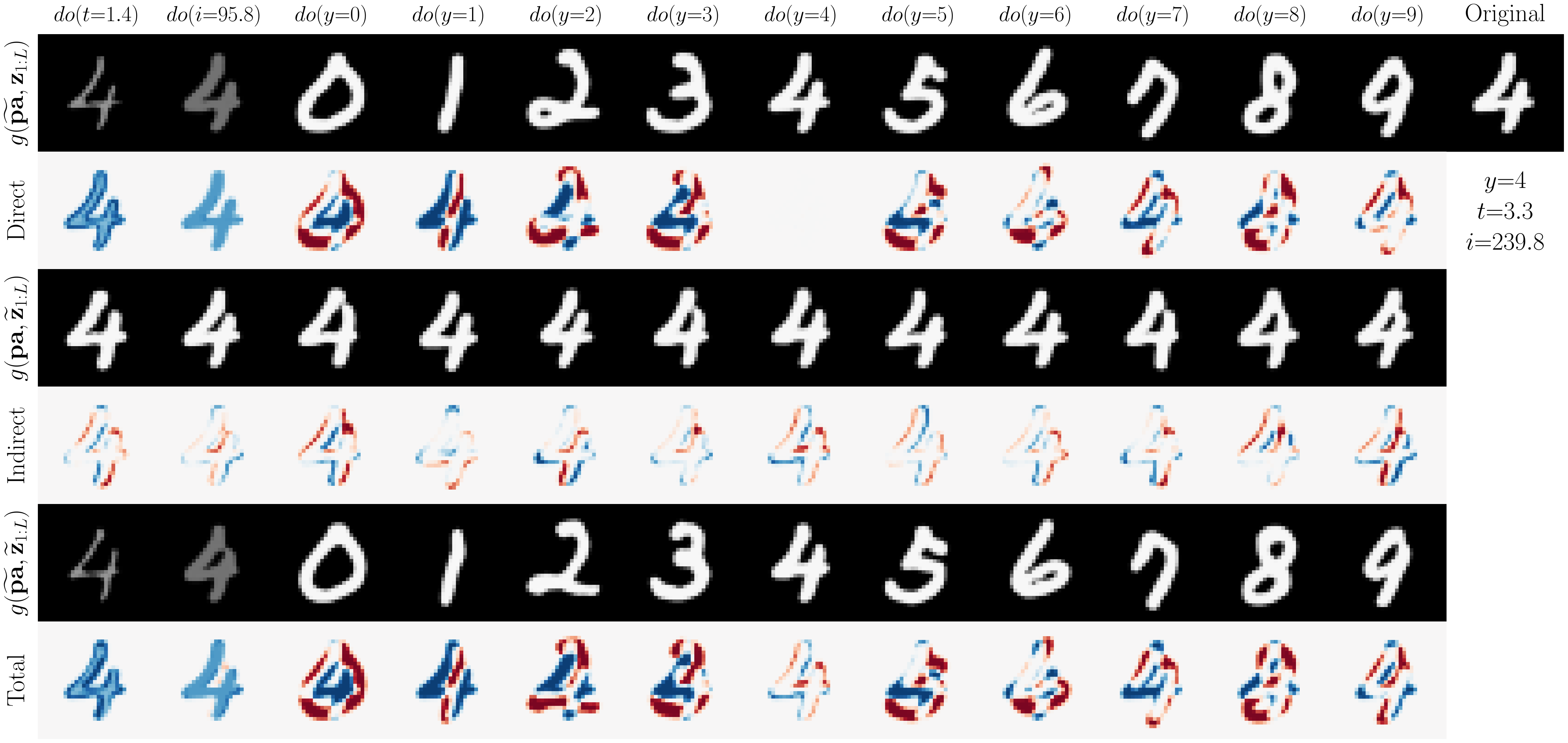}
    \end{subfigure}
    \hfill
    \begin{subfigure}{.49\textwidth}
        \includegraphics[width=\textwidth]{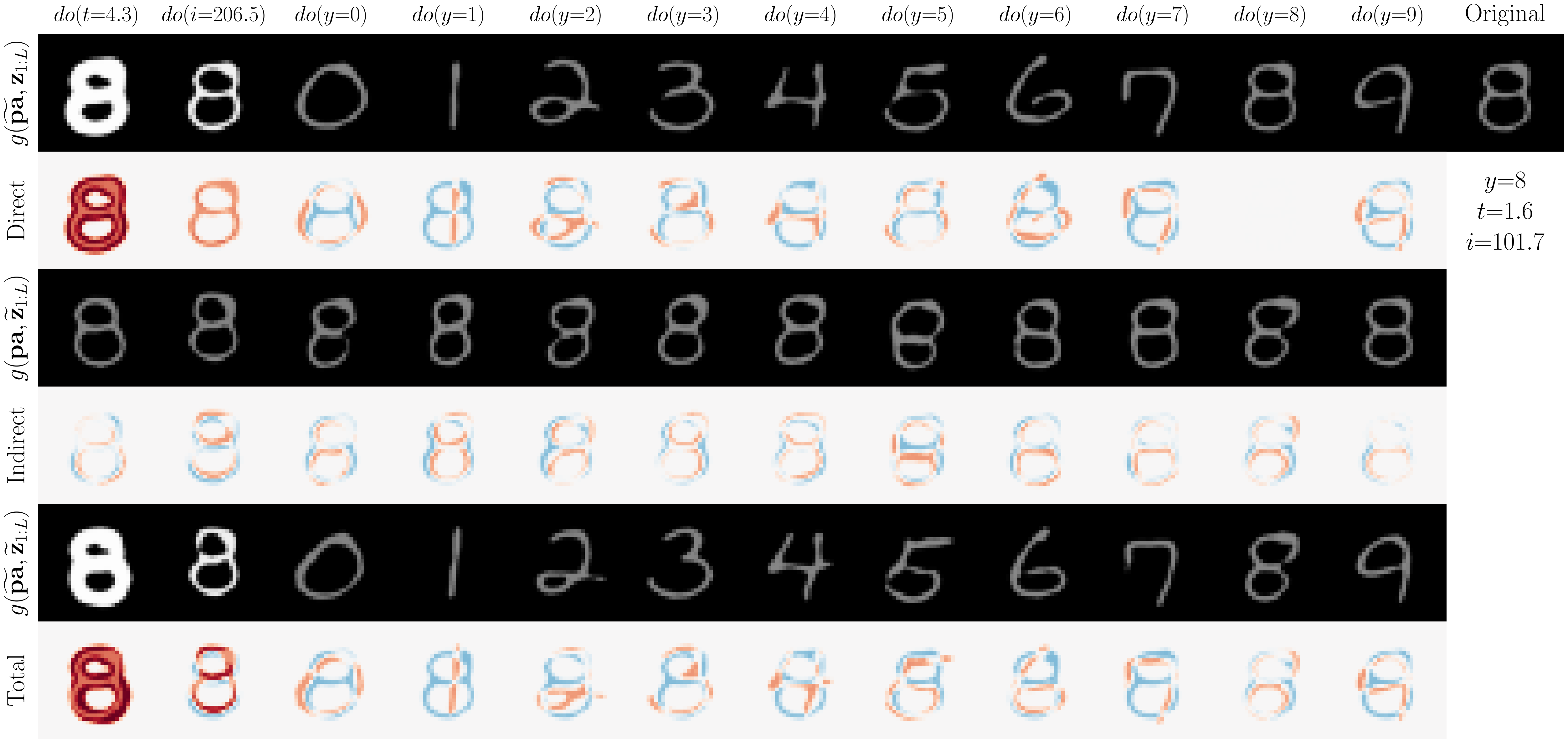}
    \end{subfigure}
    \hfill
    \\[20pt]
    \begin{subfigure}{.49\textwidth}
        \includegraphics[width=\textwidth]{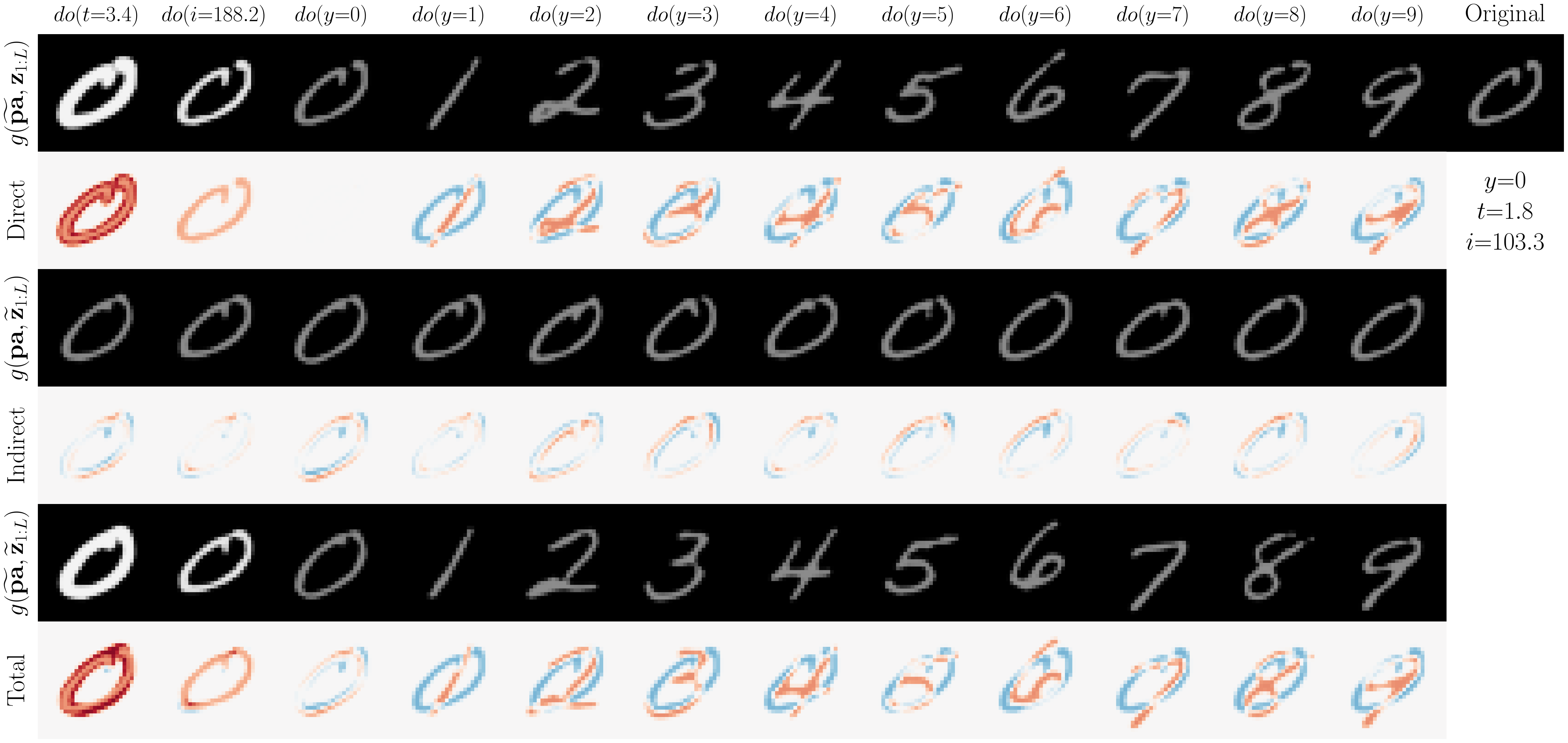}
    \end{subfigure}
    \hfill
    \begin{subfigure}{.49\textwidth}
        \includegraphics[width=\textwidth]{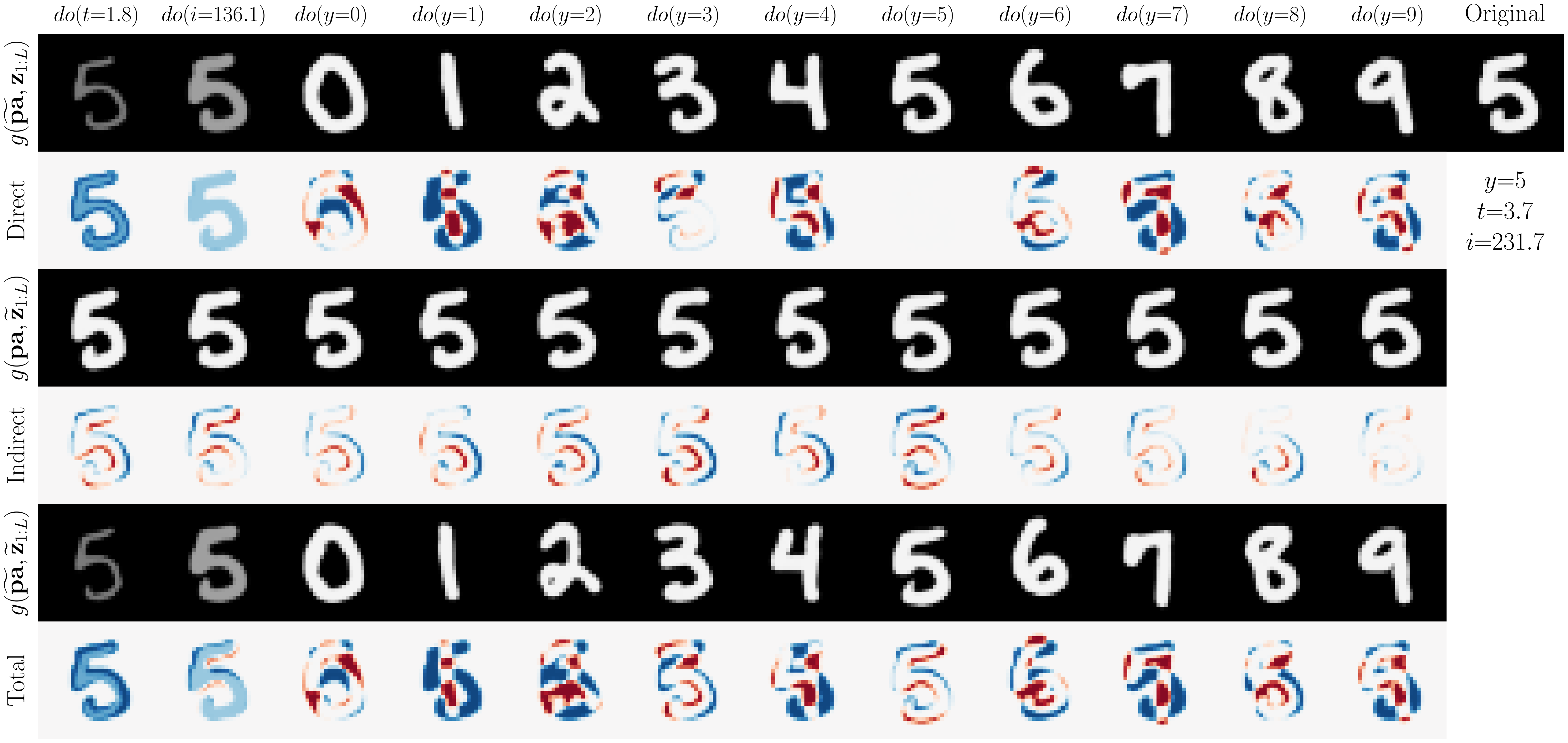}
    \end{subfigure}
    \hfill
    \\[20pt]
    \begin{subfigure}{.49\textwidth}
        \includegraphics[width=\textwidth]{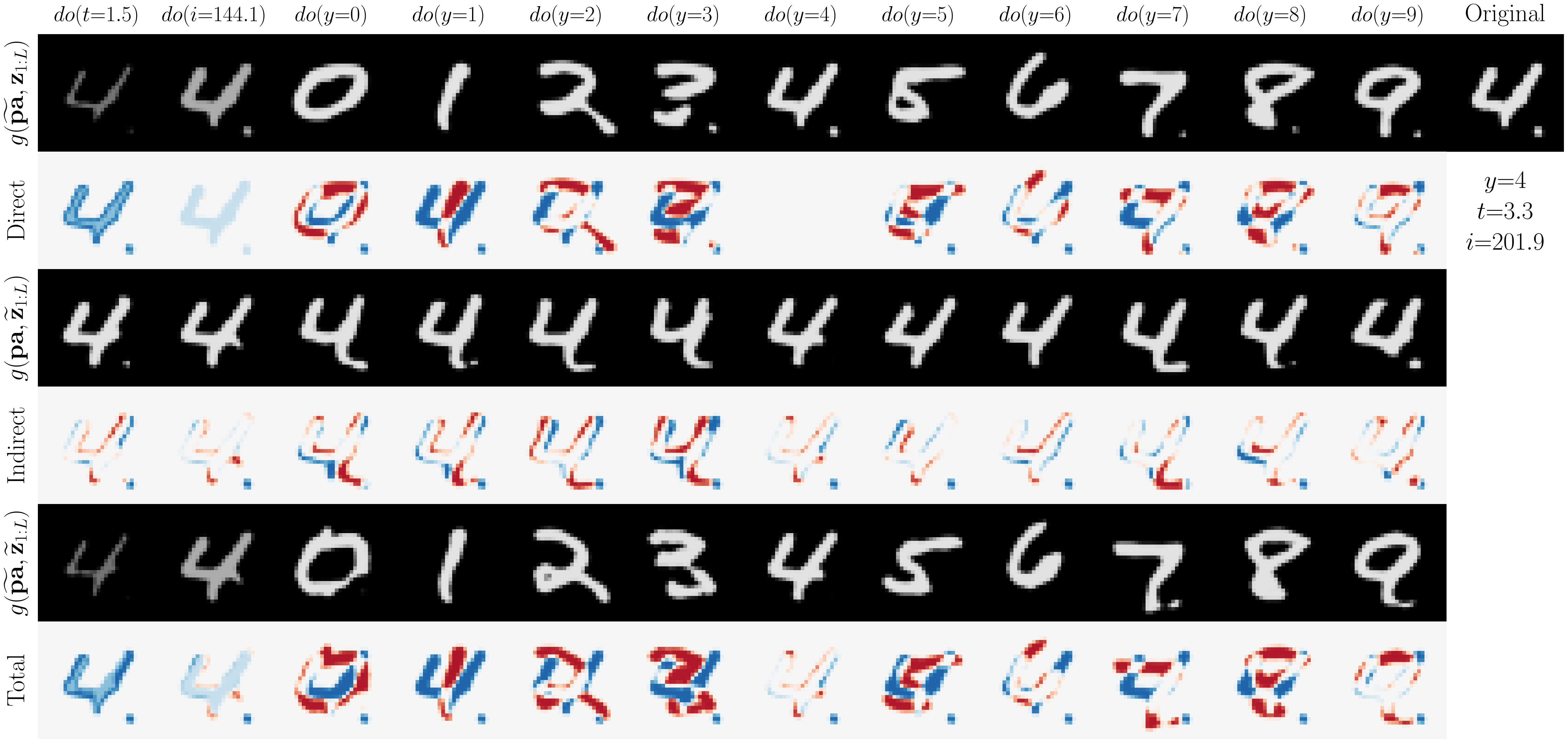}
    \end{subfigure}
    \hfill
    \begin{subfigure}{.49\textwidth}
        \includegraphics[width=\textwidth]{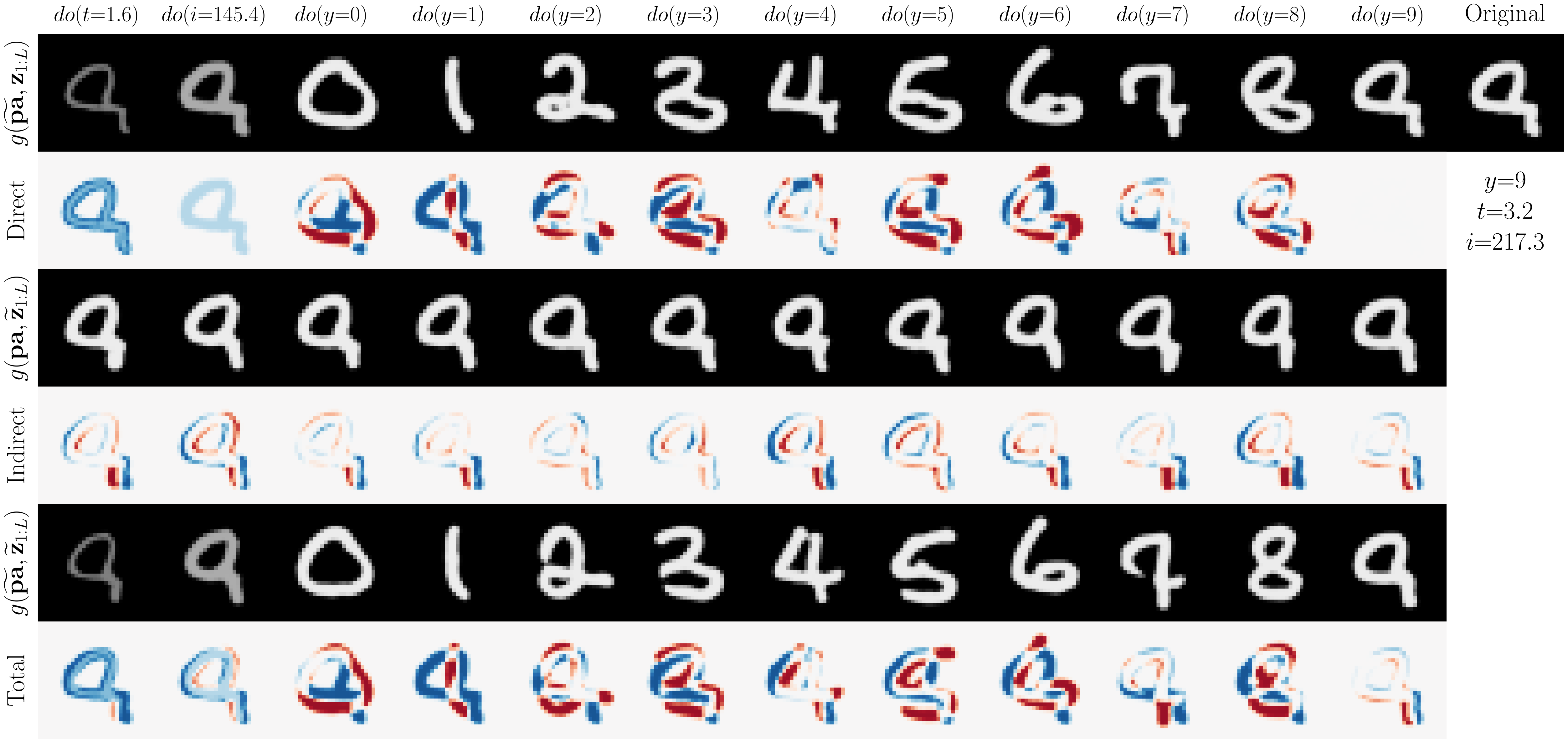}
    \end{subfigure}
    \hfill
    \\[20pt]
    \begin{subfigure}{.49\textwidth}
        \includegraphics[width=\textwidth]{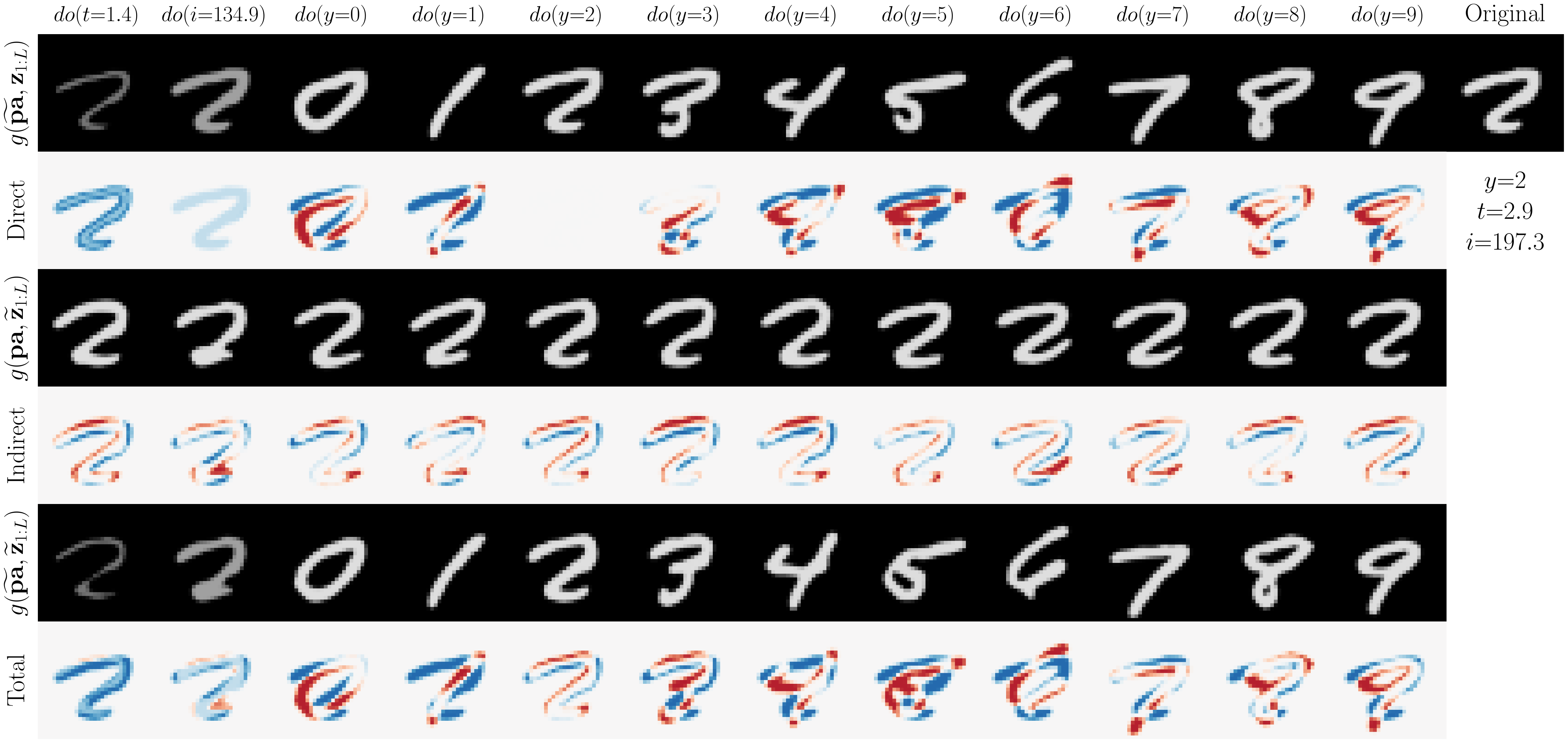}
    \end{subfigure}
    \hfill
    \begin{subfigure}{.49\textwidth}
        \includegraphics[width=\textwidth]{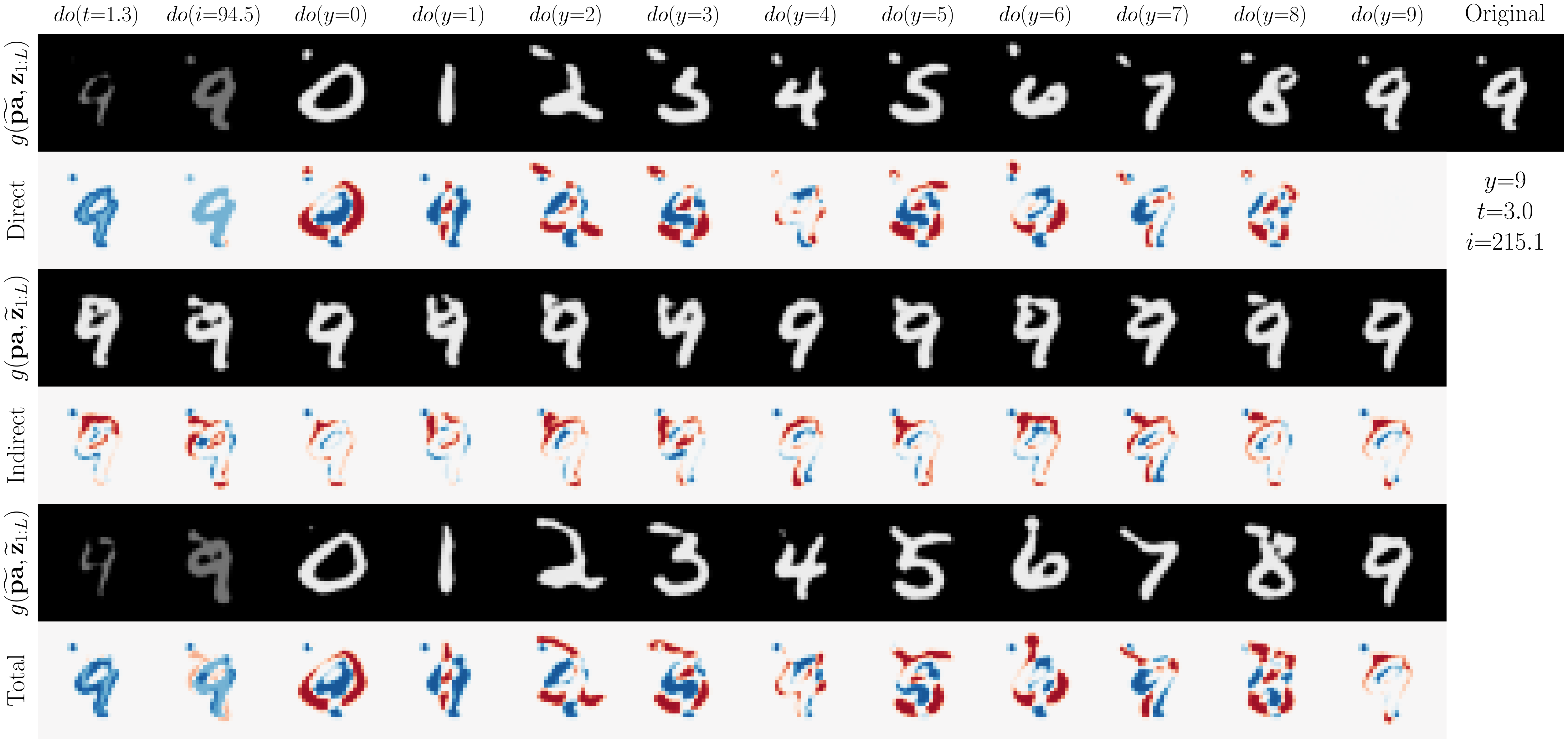}
    \end{subfigure}
    \hfill
    \caption{Visualizing additional random Morpho-MNIST test set counterfactuals from our latent mediator model ($\beta{=}1$). Direct, indirect and total effects are also shown. We observe counterfactuals which are faithful to interventions and the associated causal graph (e.g. intervening on thickness causes intensity to increase/decrease.). Intervening on the parents preserves observational identity (e.g. writing style). Recall that cross-world (indirect effect, row(s) 3) counterfactuals $\widetilde{\mathbf{x}}_{\text{IE}} \sim g(\mathbf{pa}_\mathbf{x}, \widetilde{\mathbf{z}}_{1:L})$ are the outcome of $\mathbf{x}$ given observed $\mathbf{pa}_\mathbf{x}$ and the counterfactual mediator we would have observed $\widetilde{\mathbf{z}}_{1:L}$ had $\mathbf{pa}_\mathbf{x}$ been $\widetilde{\mathbf{pa}}_\mathbf{x}$. We can see that the effect of the parents is mediated by $\mathbf{z}_{1:L}$, which mostly affects the writing style, and changes according to which counterfactual parent caused the (inferred) counterfactual mediator $\widetilde{\mathbf{z}}_{1:L}$.}
\end{figure}
\newpage
\begin{figure*}[!h]
    \centering
    \includegraphics[trim={0 0 150 0}, clip, width=\textwidth]{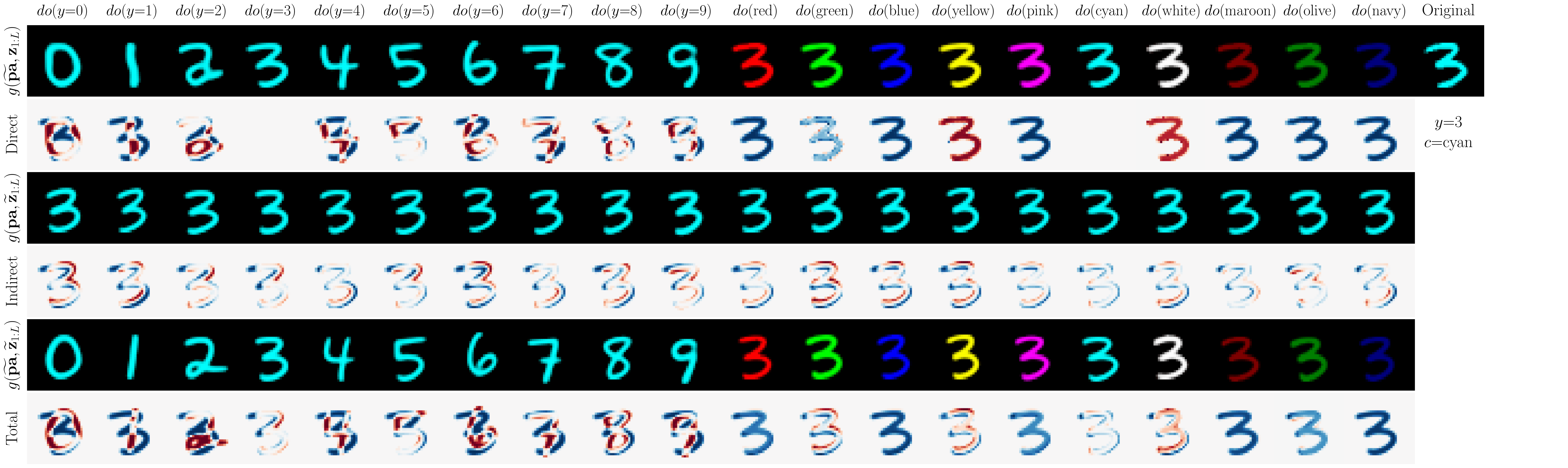}
    \caption{Test set counterfactuals produced by our latent mediator model using the Colour-MNIST dataset from~\cite{monteiro2023measuring}. In this case, the SCM is very simple and consists of only 3 variables: digit $y$, colour $c$ and image $\mathbf{x}$. There are 10 different discrete colours, there is no assumed causal relationship between $y$ \& $c$, and both $y$ and $c$ cause $\mathbf{x}$. We can observe clear localised changes based on the intervened upon parent variable whilst the identity of the observation is preserved. There is also no (visible) causal influence from $y$ to $c$ or vice-versa like in the thickness $t$ and intensity $i$ scenario from Morpho-MNIST where $t$ caused $i$. Direct, indirect and total causal effects are also shown.}
\end{figure*}
\newpage
\begin{figure}[t]
    \centering
    \includegraphics[width=.95\textwidth]{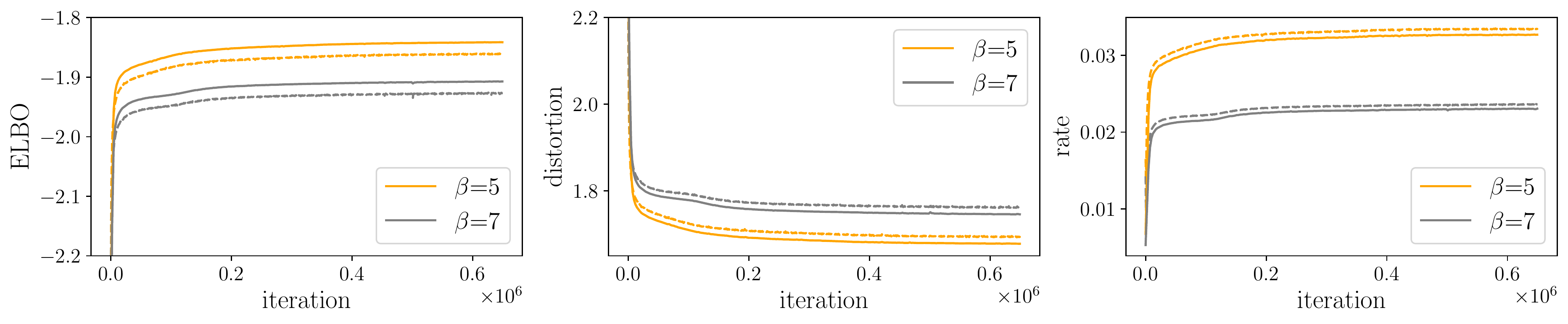}
    \caption{Evidence lower bound, distortion and rate throughout training of our exogenous prior HVAE mechanism on UK Biobank (shown in nats per dimension, top row). Dashed and solid lines correspond to training and validation respectively.}
    \label{fig:brains_elbo}
\end{figure}
\begin{figure}[t]
    \centering
    \includegraphics[width=.95\textwidth]{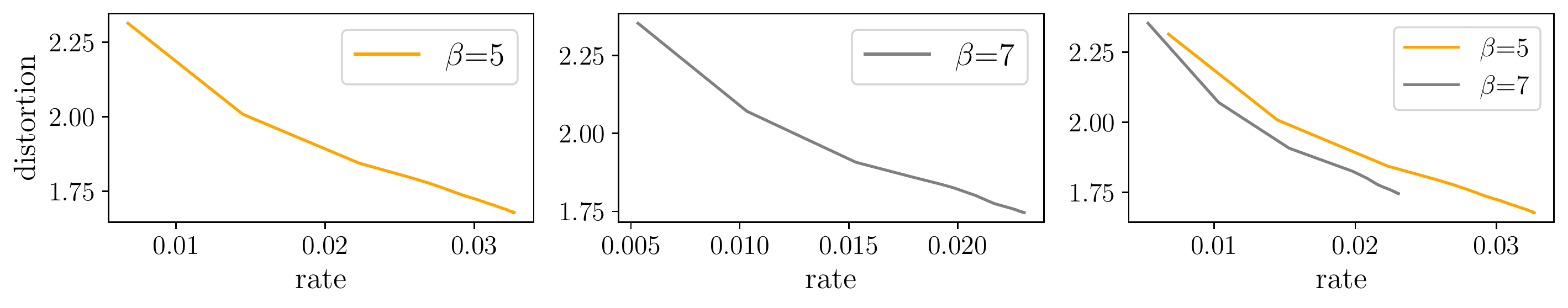}
    \label{fig:brains_rate_distortion}
    \caption{Rate-distortion curves of our exogenous prior HVAE mechanism on UK Biobank. We can again observe higher $\beta$ penalty encourages higher semantic compression, which leads to improvements in counterfactual conditioning (in terms of counterfactual effectiveness) at the cost of worse distortion (counterfactual composition).}
\end{figure}
\section{Brain MRI (UK Biobank)}
\label{app:brains}
\subsection{Dataset Details} In terms of data generation and pre-processing, we follow the original pipeline used by~\citet{alfaro2018image} and~\citet{pawlowski2020deep}. The pre-processing entails skull removal, bias field correction, segmentation of brain structures, and registration. Mid-axial 2D slices were then extracted and max-min normalised to $[0,255]$ inside the brain mask, whereas background pixels were set to zero. The attributes for each subject (age, sex, brain/ventricle volume) were retrieved from the UK Biobank dataset. In addition, we use both T1-weighted and T2-FLAIR brain MRI scans (when available) and include a binary indicator variable ($m$) for the scan modality in our structural causal models. We randomly split the full dataset into subsets of 19466 training, 3500 validation and 3500 test samples. Further, we ensure no overlapping subjects between the training and evaluation datasets exist.

\subsection{Experiment Setup}
\label{sec:brain_exp_setup}
\paragraph{Architecture.} For the Brain MRI experiments, we used a scaled-up version of our exogenous prior HVAE for $\mathbf{x}$'s mechanism to accommodate the higher resolution of $192{\times}192$ (see details in Appendix C). The stochastic latent variables in our HVAE span 5 resolution scales up to $\frac{1}{2}$ the input resolution: $\{1^2, 6^2, 12^2, 24^2, 48^2, 96^2\}$, and each respective resolution scale contains the following number of residual blocks: $\{2,4,8,12,8,4\}$. Each latent variable has 16 channels and the feature map widths at each resolution scale are: 
$\{512, 192, 160, 128, 96, 64, 32\}$, where $32$ refers to the width of the final (deterministic) $192{\times}192$ upsampling residual block. The resulting architecture comprises a total of 38 stochastic latent variables layers and $\simeq$17M trainable parameters. Conditioning this HVAE on the parents $\mathbf{pa}_\mathbf{x}$ follows the same expansion/concatenation strategy as for the Morpho-MNIST experiments. It is likely that using a more sophisticated conditioning strategy involving spatial/cross attention would perform better, but the one we used is simple and performed well enough in our experiments so we leave further exploration to future work.

\paragraph{Training Details.} We trained our HVAEs for $\simeq$650K iterations with a batch size of 32 and the AdamW optimizer. We used an initial learning rate of 1e-3 with 100 iterations of linear warmup, $\beta_1=0.9$, $\beta_2=0.9$ and a weight decay of $0.05$. We set gradient clipping to 350 and used a gradient update skipping threshold of 500 based on the $L2$ norm of the gradients. The final model is an exponential moving average of the parameters with a rate of $0.999$ which we use for inference. For data-augmentation, we apply a zero-padding of $9$ to all borders and perform random horizontal flips with probability $0.5$. Pixel intensities were normalised to a range of $[-1,1]$. As explained in Section~\ref{subsec: Exogenous Prior HVAE}, rather than using the (invertible) continuous likelihood mechanism proposed by~\citet{pawlowski2020deep} which requires dequantization of discrete pixel intensities and inversion of the sampling mechanism $h(\cdot)$ during training, we used a discretized Gaussian likelihood as is commonly used in Diffusion models~\cite{ho2020denoising} and infer the exogenous sampling noise for counterfactuals at inference time only. We found this to be beneficial in terms of training stability and final performance. Following ~\citet{ho2020denoising}, we obtain discrete log likelihoods as follows:
\begin{align}
\begin{split}
p_\theta(\mathbf{x} \mid \mathbf{z}_{1:L}, \mathbf{pa}_{\mathbf{x}}) &= \prod_{i=1}^D \int_{\delta_{-}(\mathbf{x}^i)}^{\delta_{+}(\mathbf{x}^i)} \mathcal{N}(\mathbf{x} \mid \boldsymbol{\mu}_\theta^i(\mathbf{z}_{1:L}, \mathbf{pa}_{\mathbf{x}}), \boldsymbol{\sigma}_\theta^i(\mathbf{z}_{1:L}, \mathbf{pa}_{\mathbf{x}})) \, d\mathbf{x} \label{eq:discrete_decoder} \\
\delta_{+}(\mathbf{x}) &= \begin{cases}
\infty & \text{if}\ \mathbf{x}=1 \\
\mathbf{x}+\frac{1}{255} & \text{if}\ \mathbf{x} < 1
\end{cases}
\qquad \delta_{-}(\mathbf{x}) = \begin{cases}
-\infty & \text{if}\ \mathbf{x}=-1 \\
\mathbf{x}-\frac{1}{255} & \text{if}\ \mathbf{x} > -1
\end{cases}
\end{split}
\end{align}
where $D$ is the data dimensionality and the $i$ superscript denotes a single coordinate.

\paragraph{Alternative Mechanisms.} As our focus is on high fidelity counterfactual generation, we elected not to compare directly with a simple VAE baseline for $\mathbf{x}$'s mechanism in these experiments (e.g.~\cite{pawlowski2020deep,monteiro2023measuring}), as simple VAEs are known to perform poorly in these scenarios. We felt that the comparisons would not be apple-to-apples or particularly meaningful. Early attempts to train Normalizing Flow based causal mechanisms (which \textit{are} directly amenable to abduction) revealed prohibitory training instabilities in large-scale high resolution settings, as also discussed in~\citet{pawlowski2020deep}. Furthermore, alternative deep generative models like GANs and Diffusion models are not directly amenable to explicit abduction like HVAEs, so we leave the required practical/theoretical modifications to future work. Promising avenues include variational diffusion models~\cite{kingma2021variational}, GAN inversion~\cite{xia2022gan}, and the diffusion-based approach studied by~\citet{sanchez2021diffusion}(Diff-SCMs), albeit in simplistic two variable causal models.

Notably, counterfactuals from Diff-SCMs~\cite{sanchez2021diffusion} can be susceptible to progressive loss of the observation's identity. This is partly because the abducted exogenous noise at time $T$ from the DDIM~\cite{song2021denoising} forward diffusion process (using the learned model) is not guaranteed to be semantically meaningful~\cite{preechakul2022diffusion}, or identity-preserving as one iteratively reverses diffusion towards the counterfactual parent conditioning. \citet{preechakul2022diffusion} attempt to address this lack of semantic meaning in diffusion model latents by introducing a two-part latent code inspired by StyleGAN~\cite{karras2019style}. The first part is a semantically meaningful code vector inferred from an additional trained encoder, and the second part captures stochastic details via a diffusion model conditioned on the first part. Nonetheless, they explain that certain image reasoning tasks may require more precise local latent variables, for which 2D latent variable maps can be beneficial. This view validates our HVAE-based approach. Further, our HVAE-based mechanisms were designed to adhere to structural equation modelling by explicitly attempting to disentangle the role of the exogenous noise from the parent conditioning: $\mathbf{x} = f_\theta(\mathbf{pa}_\mathbf{x}, \mathbf{u}_{\mathbf{x}})$, where $p(\mathbf{u}_{\mathbf{x}}) = p_\theta(\mathbf{z}_{1:L})p(\boldsymbol{\epsilon})$. In this way, we leverage the exact same hierarchy of semantically meaningful abducted exogenous noise components for computing both factuals and counterfactuals, as stipulated by Pearl's theory of interventional counterfactuals~\cite{pearl2009causality}.
\onecolumn
\subsection{Extra Results}
\begin{figure}[!ht]
    \centering
    \hfill
    \begin{subfigure}{\textwidth}
        \centering
        \includegraphics[width=.7\textwidth]{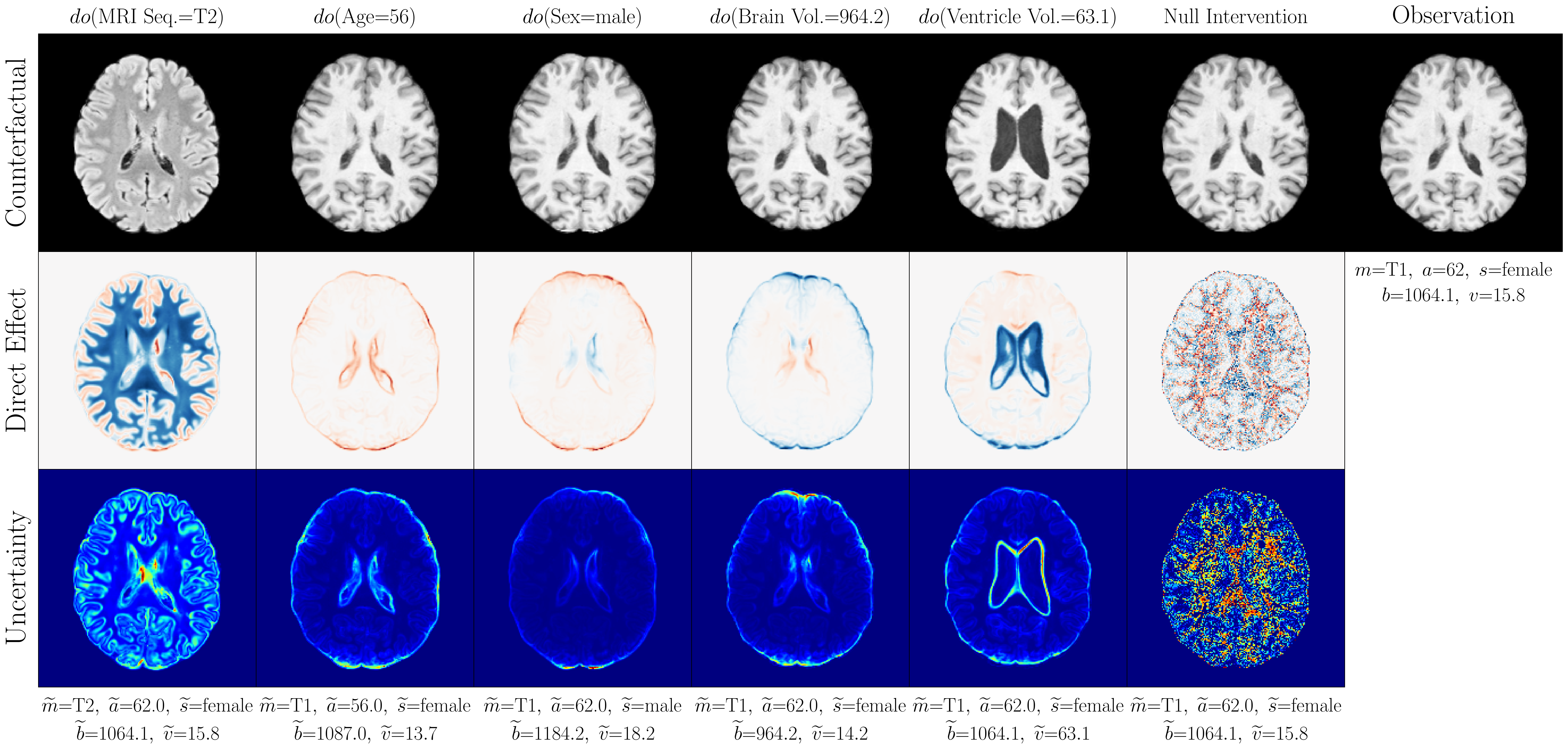}
    \end{subfigure}
    \hfill
    \\[10pt]
    \begin{subfigure}{\textwidth}
        \centering
        \includegraphics[width=.7\textwidth]{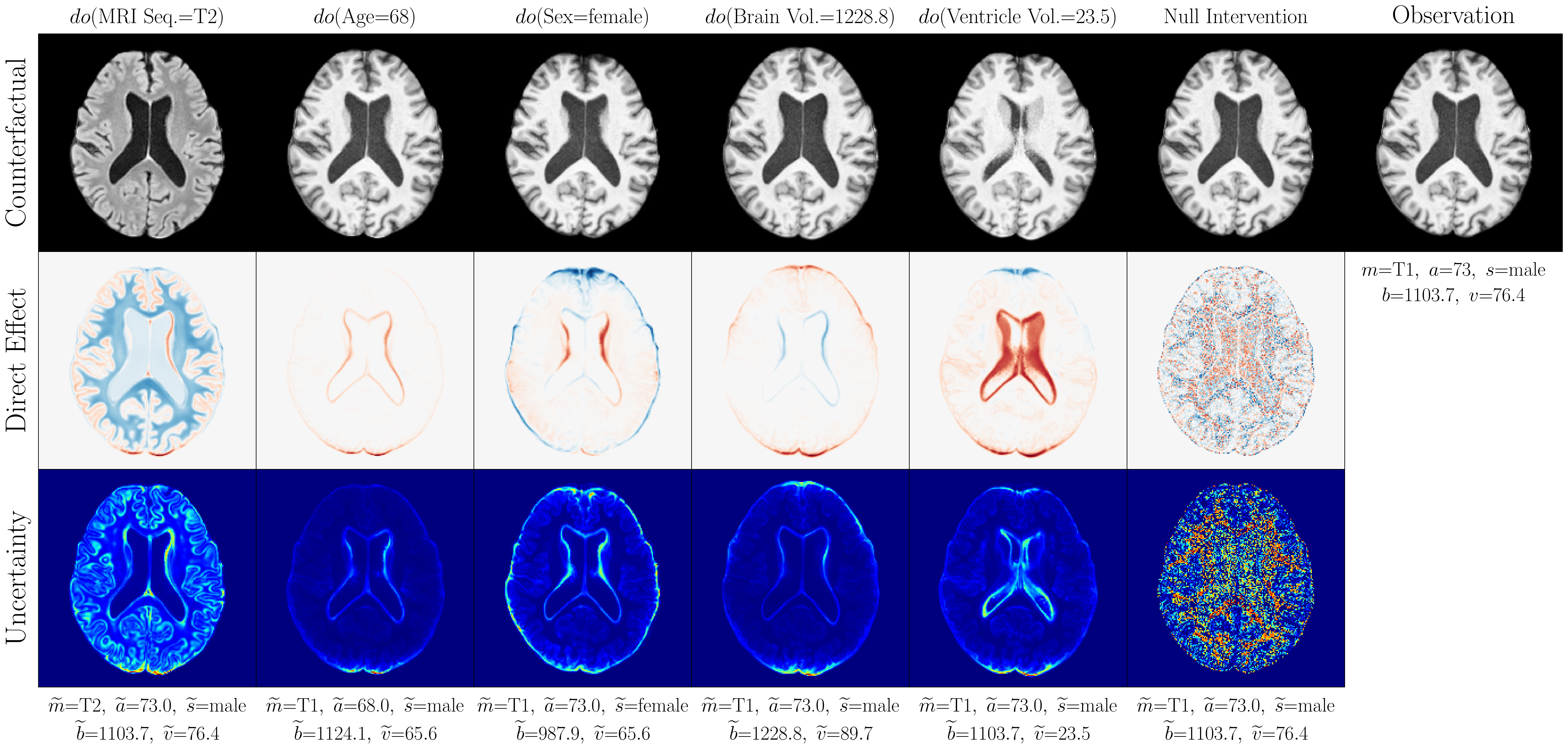}
    \end{subfigure}
    \hfill
    \\[10pt]
    \begin{subfigure}{\textwidth}
        \centering
        \includegraphics[width=.7\textwidth]{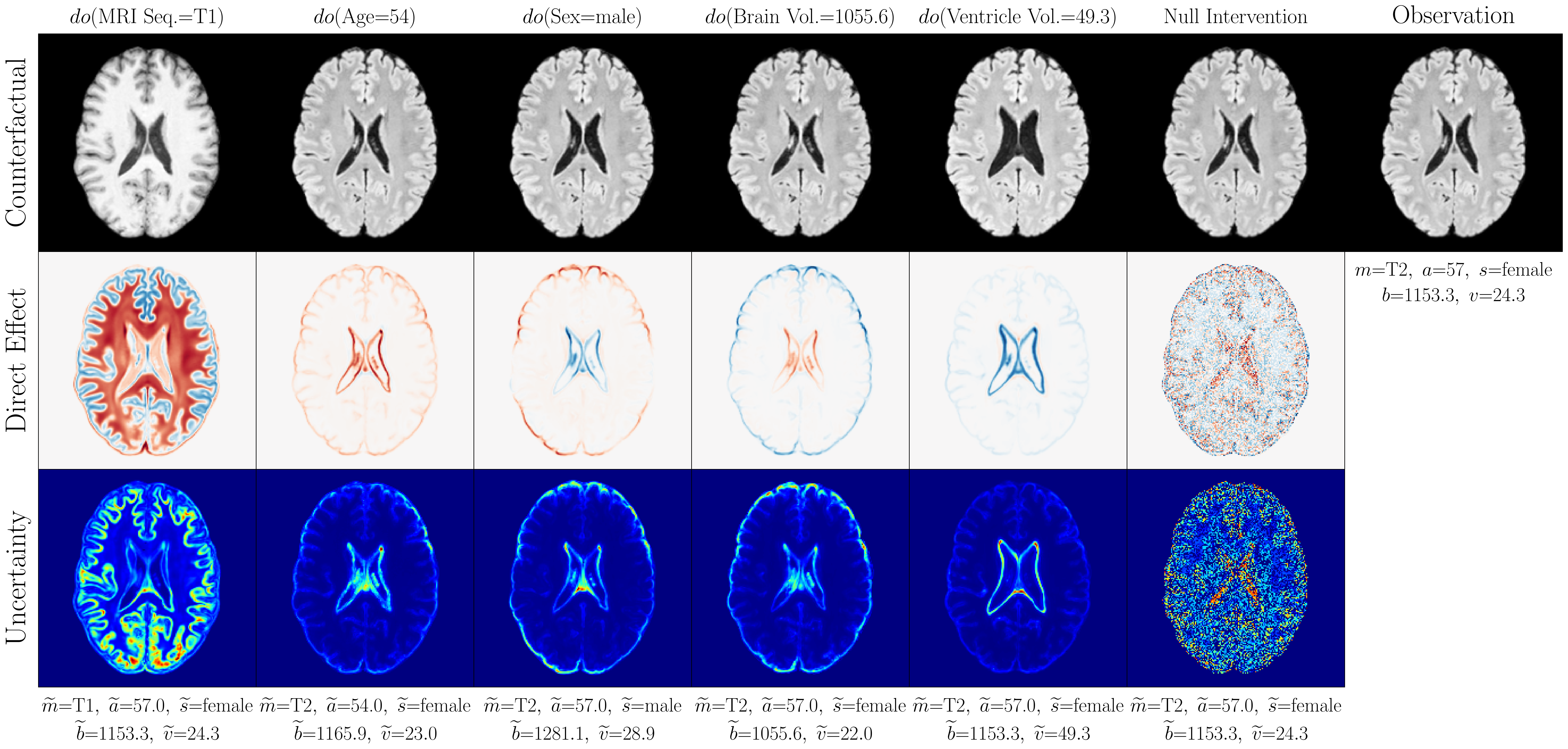}
    \end{subfigure}
    \hfill
    \caption{Additional counterfactual results from performing random interventions of each attribute (non-cherrypicked). We observed localized changes in accordance with the interventions performed and the associated assumed causal graph. Crucially, the identity of the original observation is well preserved, indicating that the approximate abduction step is fairly accurate even at high resolutions. As explained in the main text, the direct treatment effect is the difference between the generated counterfactual and the null-intervened (reconstructed) outcome. The per-pixel counterfactual uncertainty maps relate to the uncertainty in the abduction of the $\mathbf{x}$'s exogenous noise, namely $\mathbf{z}_{1:L}$ and $\boldsymbol{\epsilon}$, and were calculated as described in Appendix A.2. using 50 samples from the posterior distribution of the exogenous noise.}
\end{figure}
\begin{figure*}
    \centering
    {\large `MRI Seq.' counterfactuals $do(m)$}
    \\[5pt]
    \includegraphics[trim={0 0 0 0}, clip, width=\textwidth]
    {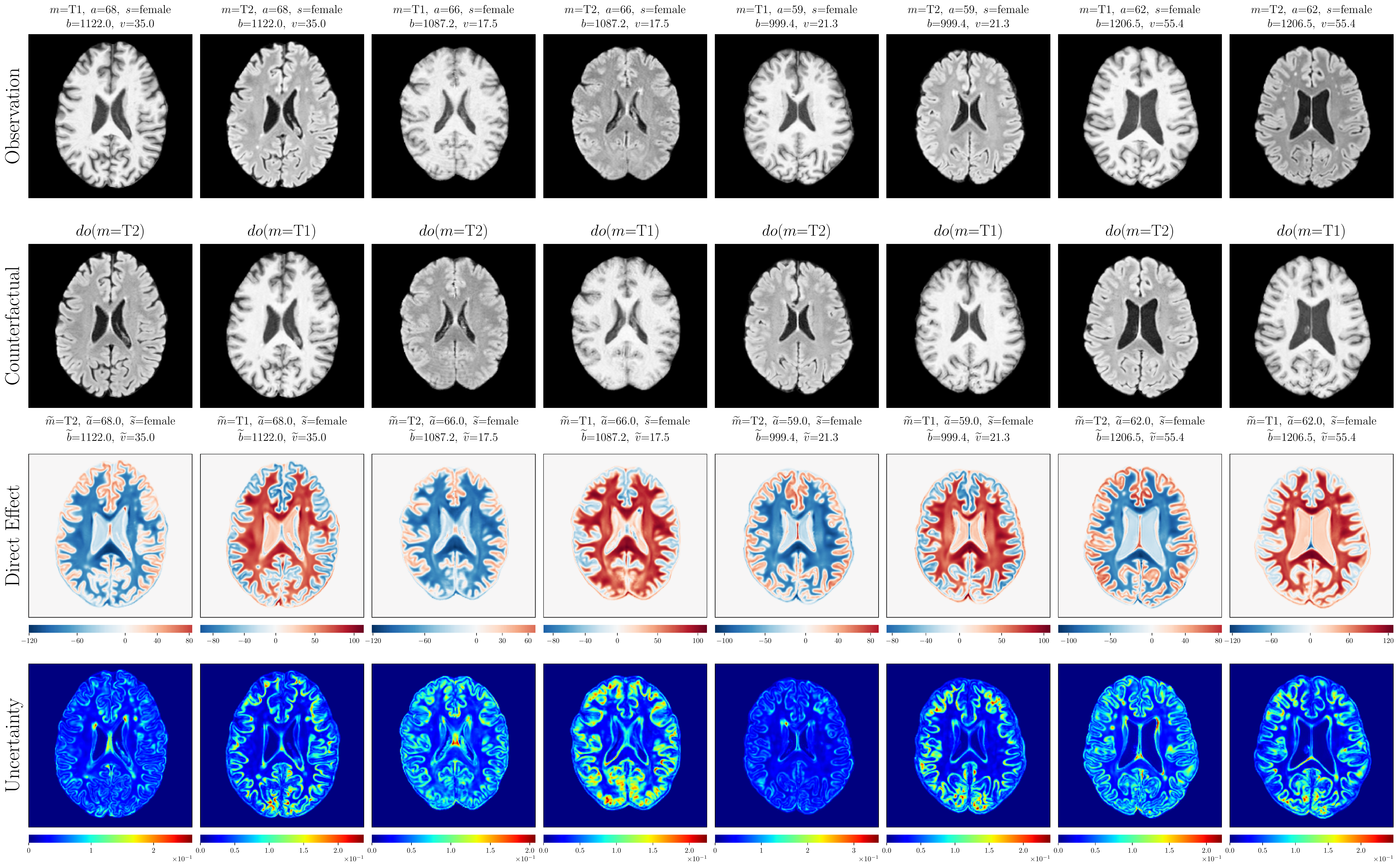}
    \\[5pt]
    \raggedright 
    \quad Post counterfactual training:
    \\[5pt]
    \centering
    \includegraphics[trim={0 0 0 0}, clip, width=\textwidth]{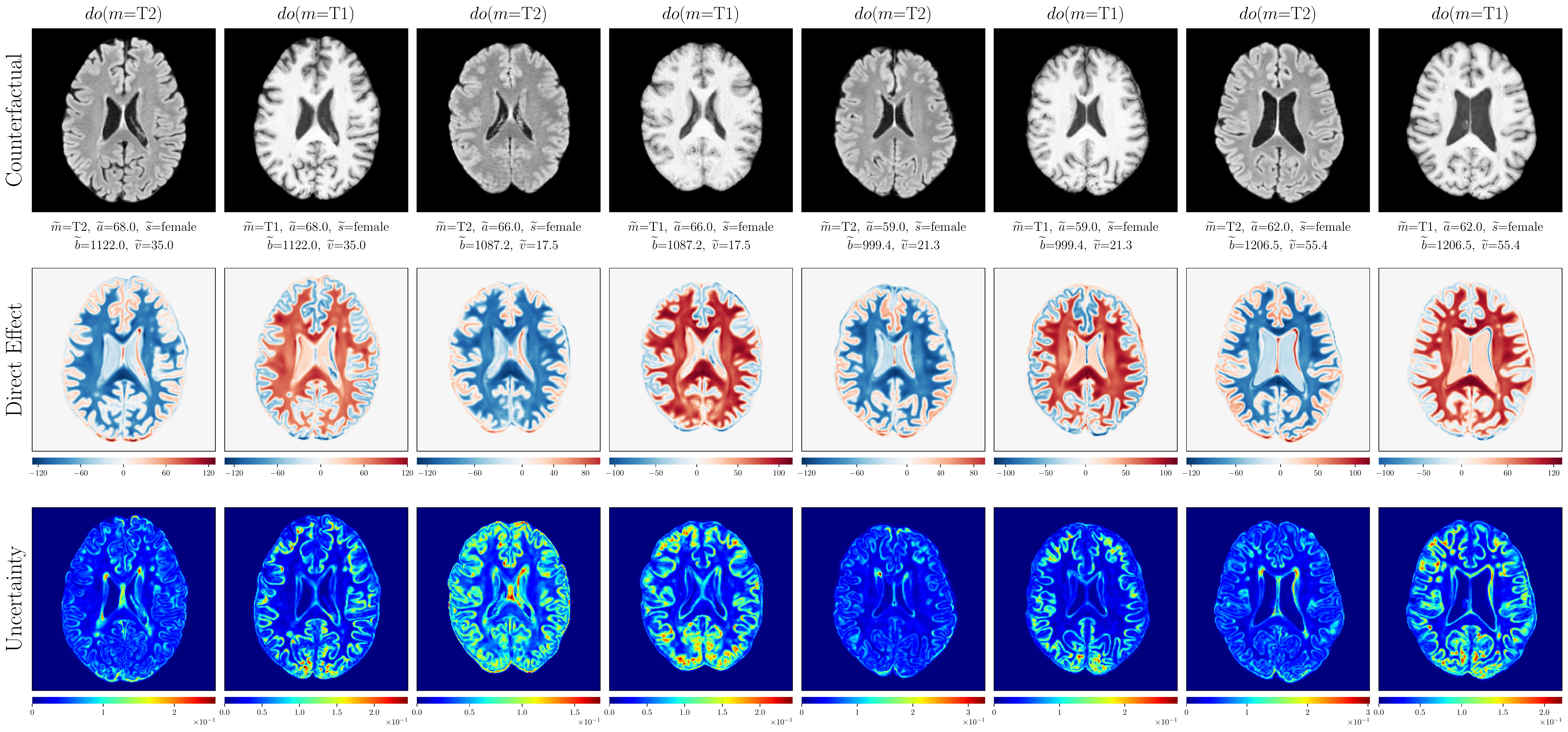}
    \caption{
    Visualizing brain MRI Seq. ($m$) test set counterfactuals from our Deep SCM using the proposed exogenous prior HVAE for $\mathbf{x}$'s mechanism (non-cherrypicked). Direct treatment effects and per-pixel counterfactual uncertainty maps are also shown. Counterfactuals were obtained from random interventions on T1/T2 ($do(m)$) before (rows 2 to 4) and after counterfactual training (rows 5 to 7). T1 and T2 scans belonging to the same subject are shown side-by-side in pairs.
    }
\end{figure*}
\begin{figure*}[!ht]
    \centering
    {\large `Sex' counterfactuals $do(s)$}
    \\[5pt]
    \includegraphics[trim={0 0 0 0}, clip, width=\textwidth]{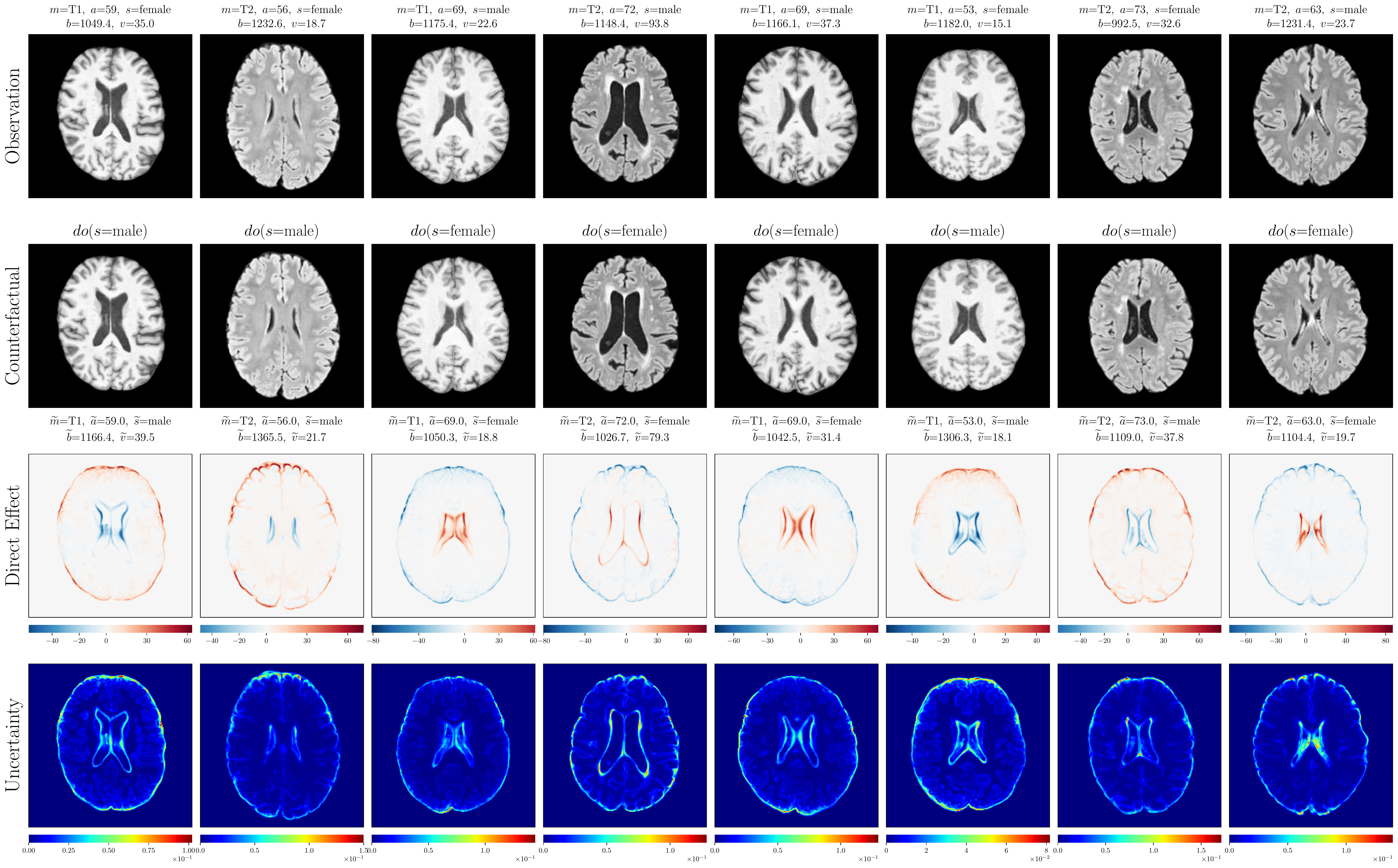}
    \\[5pt]
    \raggedright
    \ \ Post counterfactual training:
    \\[5pt]
    \includegraphics[trim={0 0 0 0}, clip, width=\textwidth]{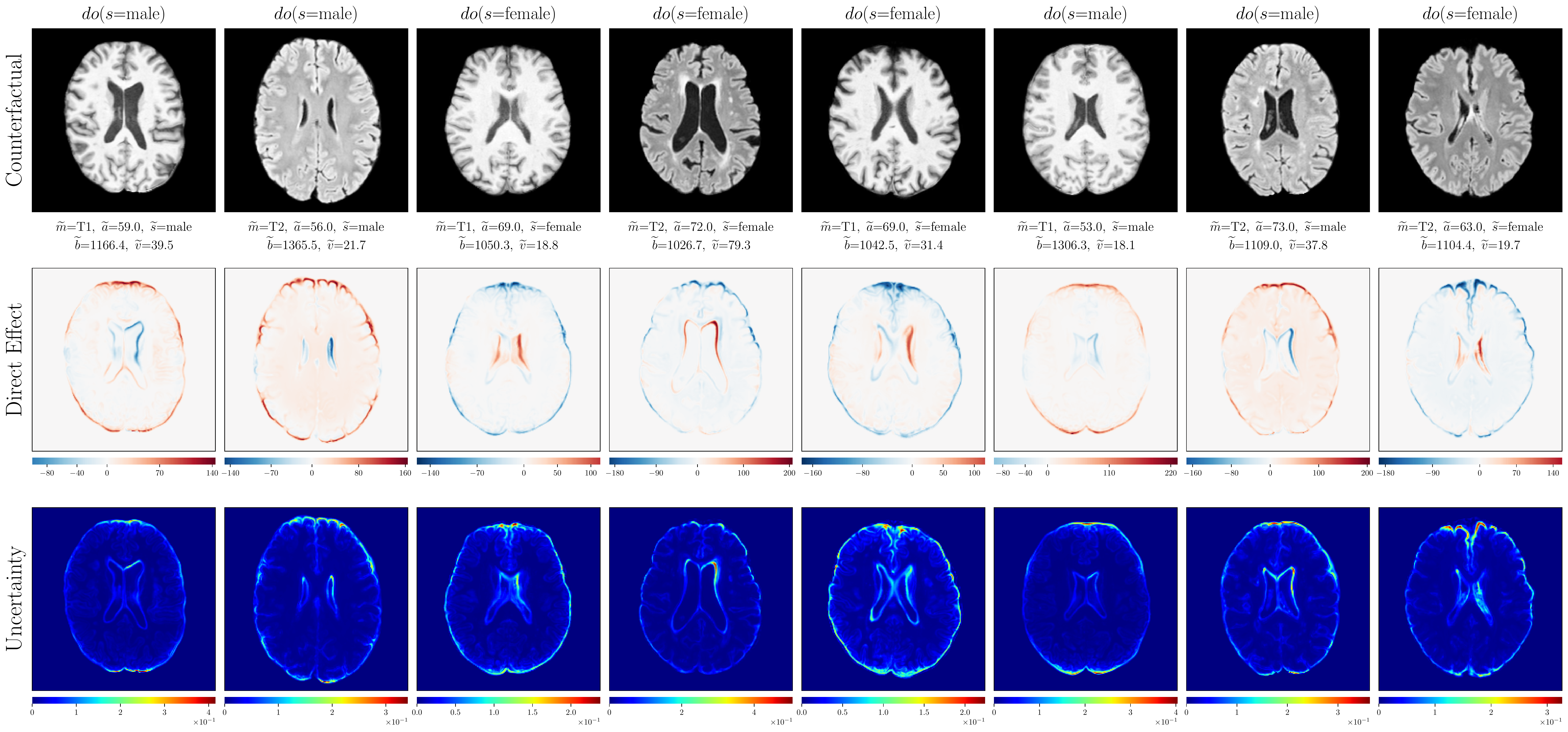}
    \caption{Visualizing brain sex ($s$) test set counterfactuals from our Deep SCM using the proposed exogenous prior HVAE for $\mathbf{x}$'s mechanism (non-cherrypicked). Direct treatment effects and per-pixel counterfactual uncertainty maps are also shown. Counterfactuals were obtained from random interventions on sex ($do(s)$) before (rows 2 to 4) and after counterfactual training (rows 5 to 7). 
    }
\end{figure*}
\begin{figure*}[!ht]
    \centering
    {\large `Age' counterfactuals $do(a)$}
    \\[5pt]
    \includegraphics[trim={0 0 0 0}, clip, width=\textwidth]{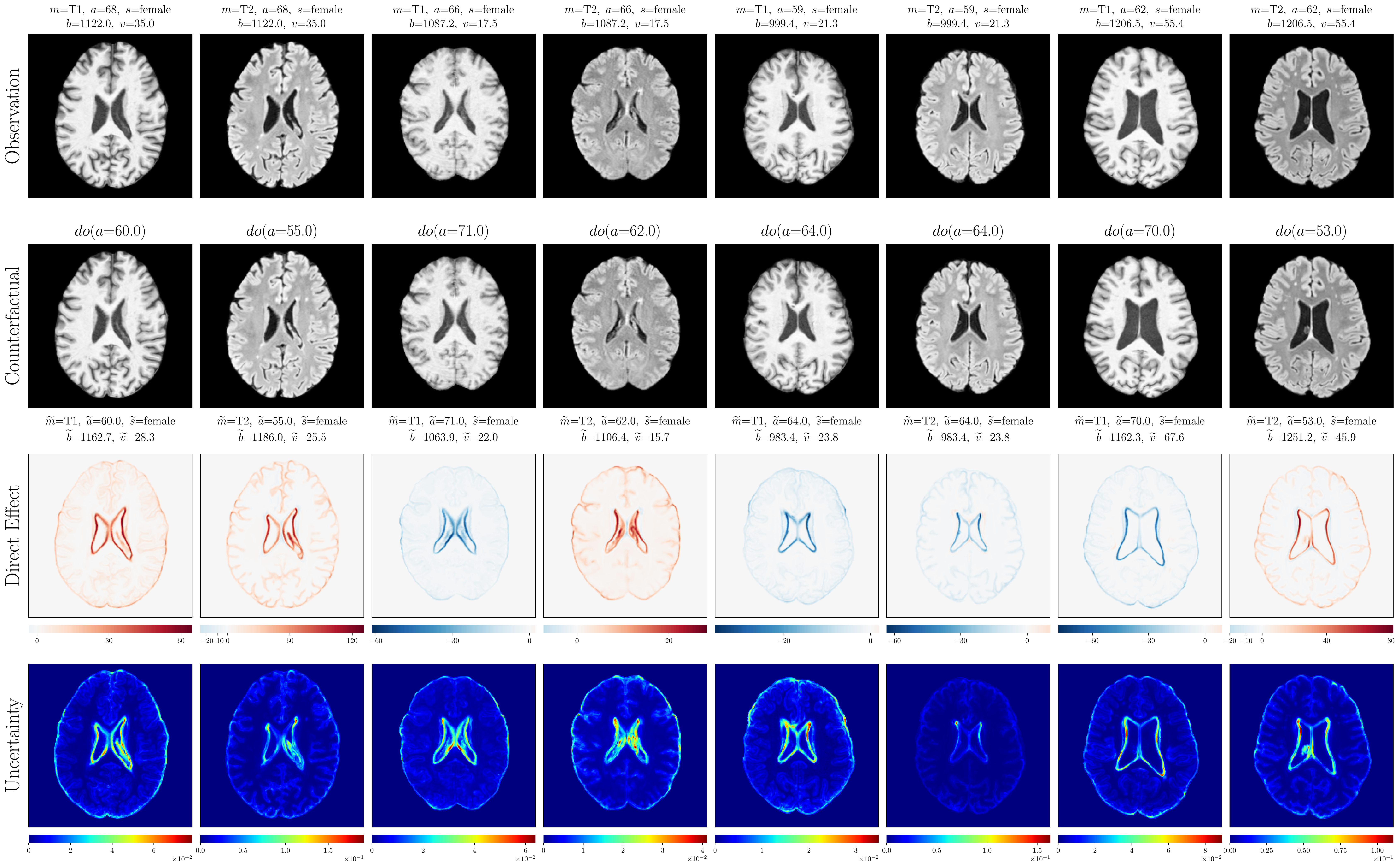}
    \\[5pt]
    \raggedright
    \ \ Post counterfactual training:
    \\[5pt]
    \includegraphics[trim={0 0 0 0}, clip, width=\textwidth]{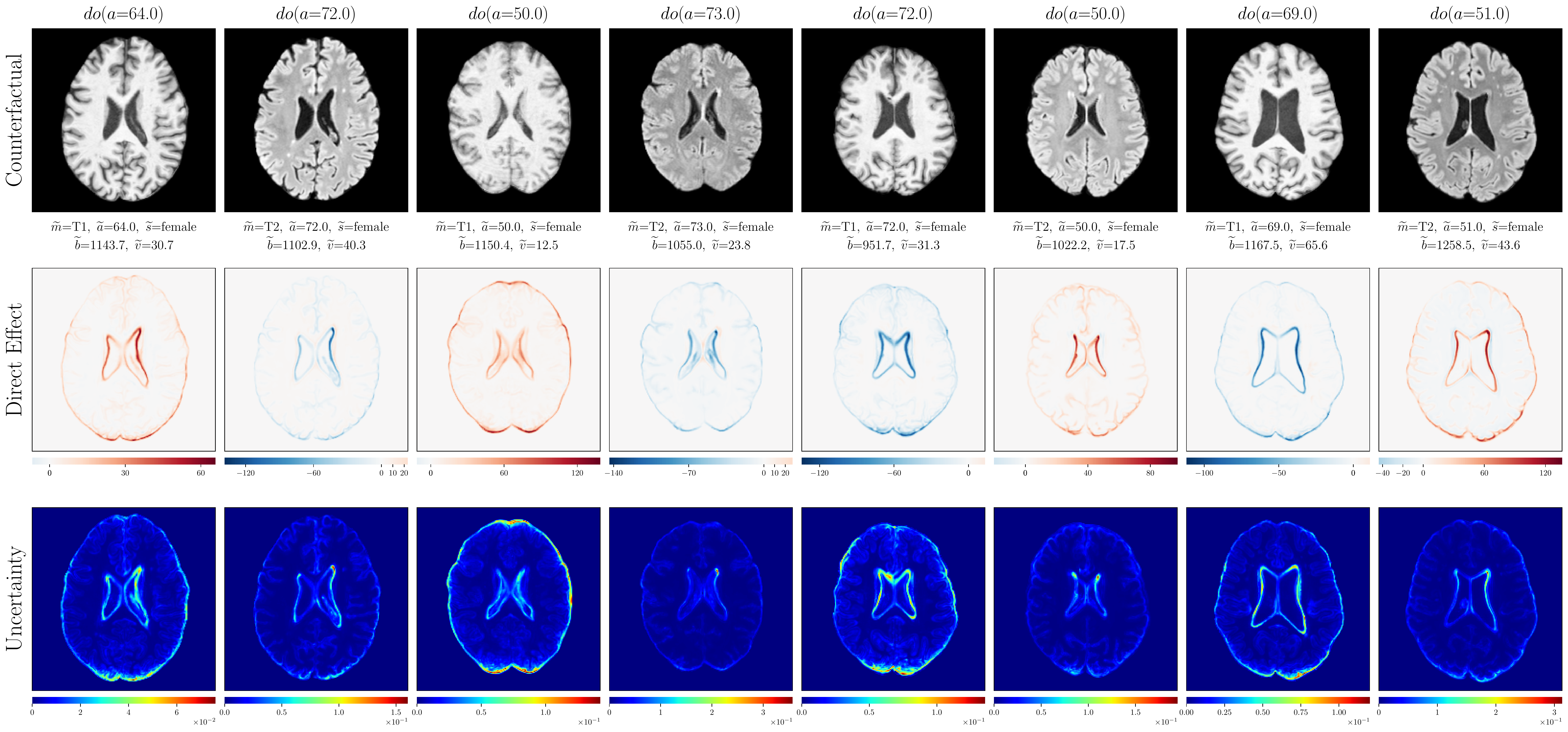}
    \caption{Visualizing brain age ($a$) test set counterfactuals from our Deep SCM using the proposed exogenous prior HVAE for $\mathbf{x}$'s mechanism (non-cherrypicked). Direct treatment effects and per-pixel counterfactual uncertainty maps are also shown. Counterfactuals were obtained from random interventions on age ($do(a)$) before (rows 2 to 4) and after counterfactual training (rows 5 to 7). 
    }
\end{figure*}
\begin{figure*}[!ht]
    \centering
    {\large `Brain Volume' counterfactuals $do(b)$}
    \\[5pt]
    \includegraphics[trim={0 0 0 0}, clip, width=\textwidth]{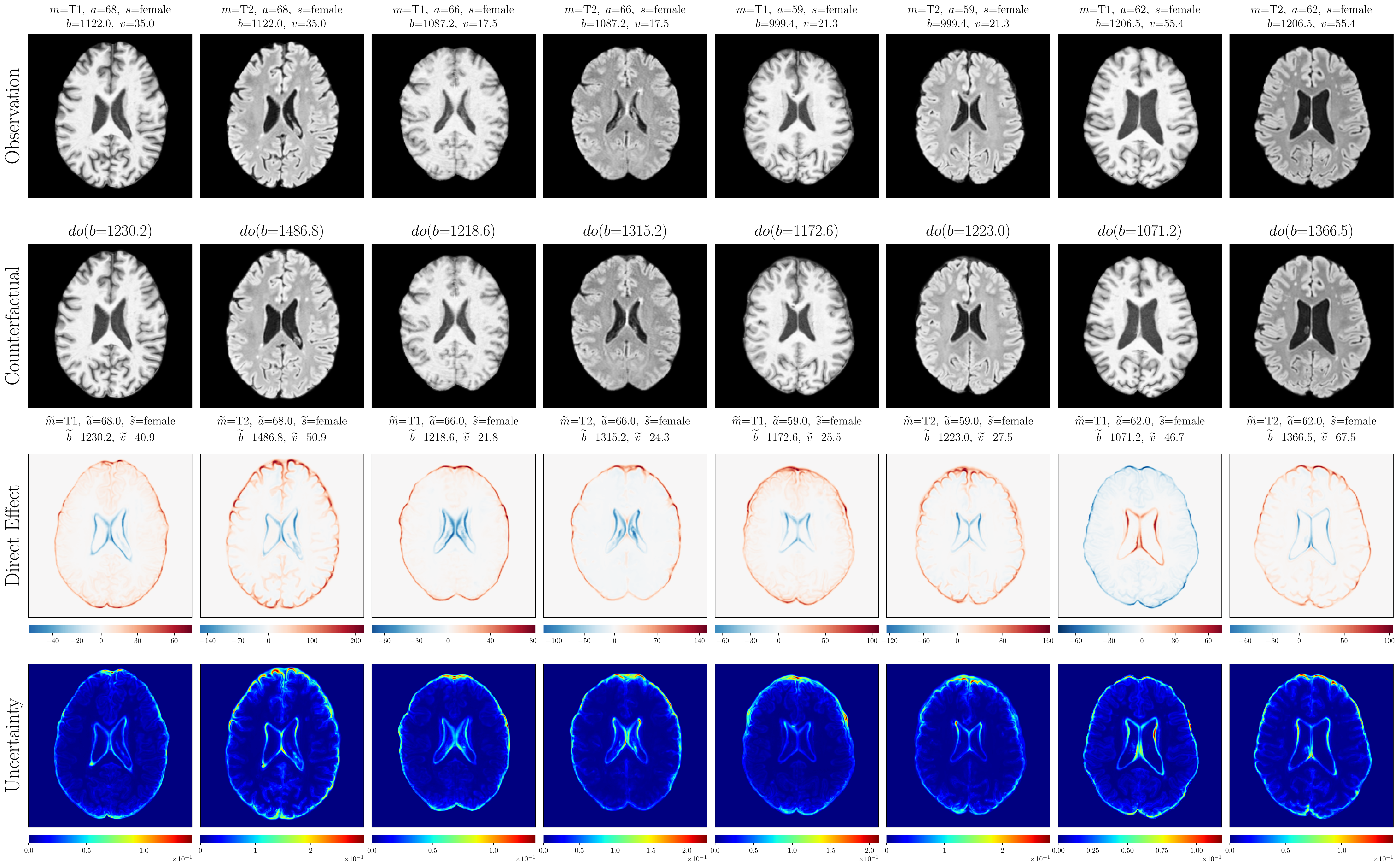}
    \\[5pt]
    \raggedright
    \ \ Post counterfactual training:
    \\[5pt]
    \includegraphics[trim={0 0 0 0}, clip, width=\textwidth]{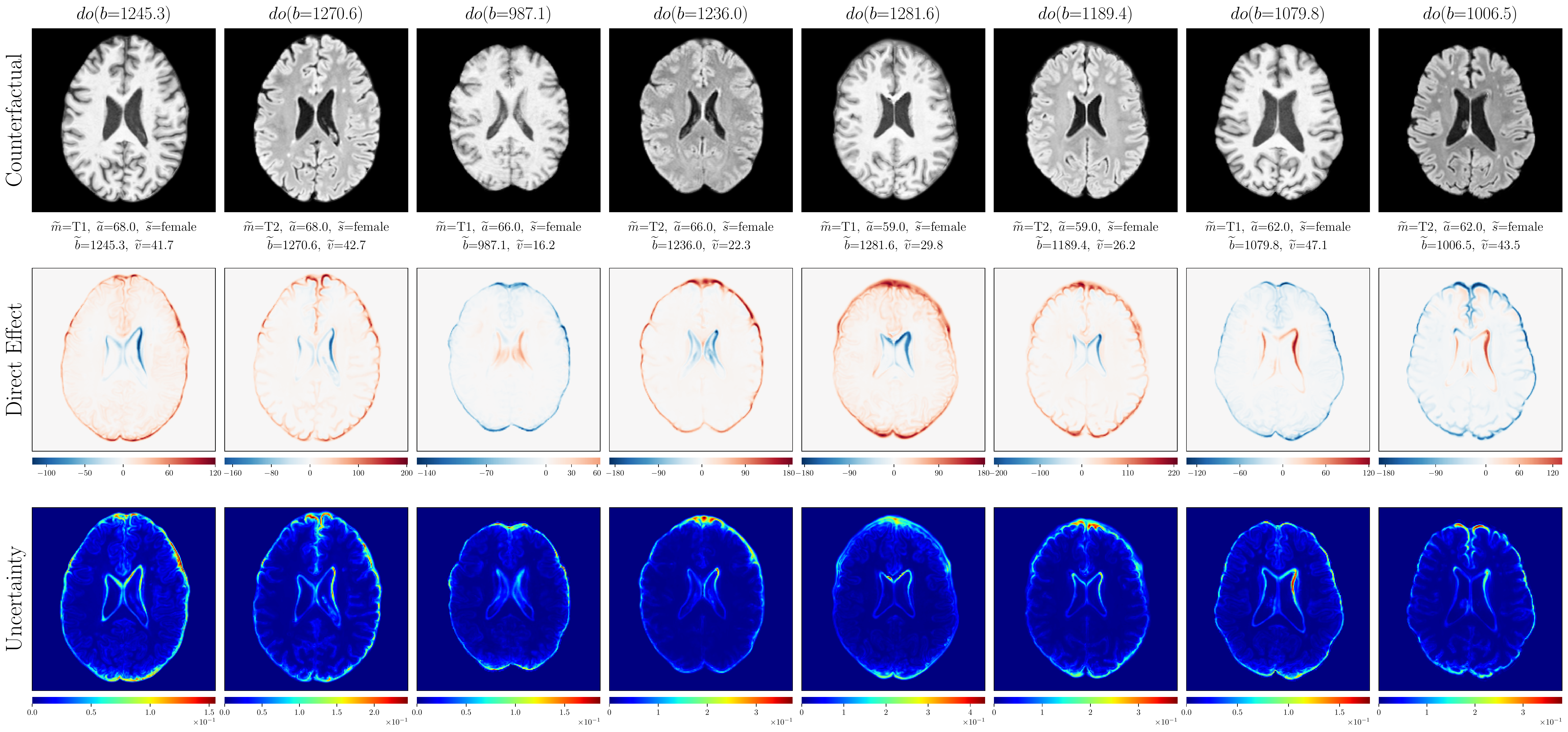}
    \caption{Visualizing brain volume ($b$) test set counterfactuals from our Deep SCM using the proposed exogenous prior HVAE for $\mathbf{x}$'s mechanism (non-cherrypicked). Direct treatment effects and per-pixel counterfactual uncertainty maps are also shown. Counterfactuals were obtained from random interventions on brain vol. ($do(b)$) before (rows 2 to 4) and after counterfactual training (rows 5 to 7). 
    }
\end{figure*}
\begin{figure*}[!ht]
    \centering
    {\large `Ventricle Volume' counterfactuals $do(v)$}
    \\[5pt]
    \includegraphics[trim={0 0 0 0}, clip, width=\textwidth]{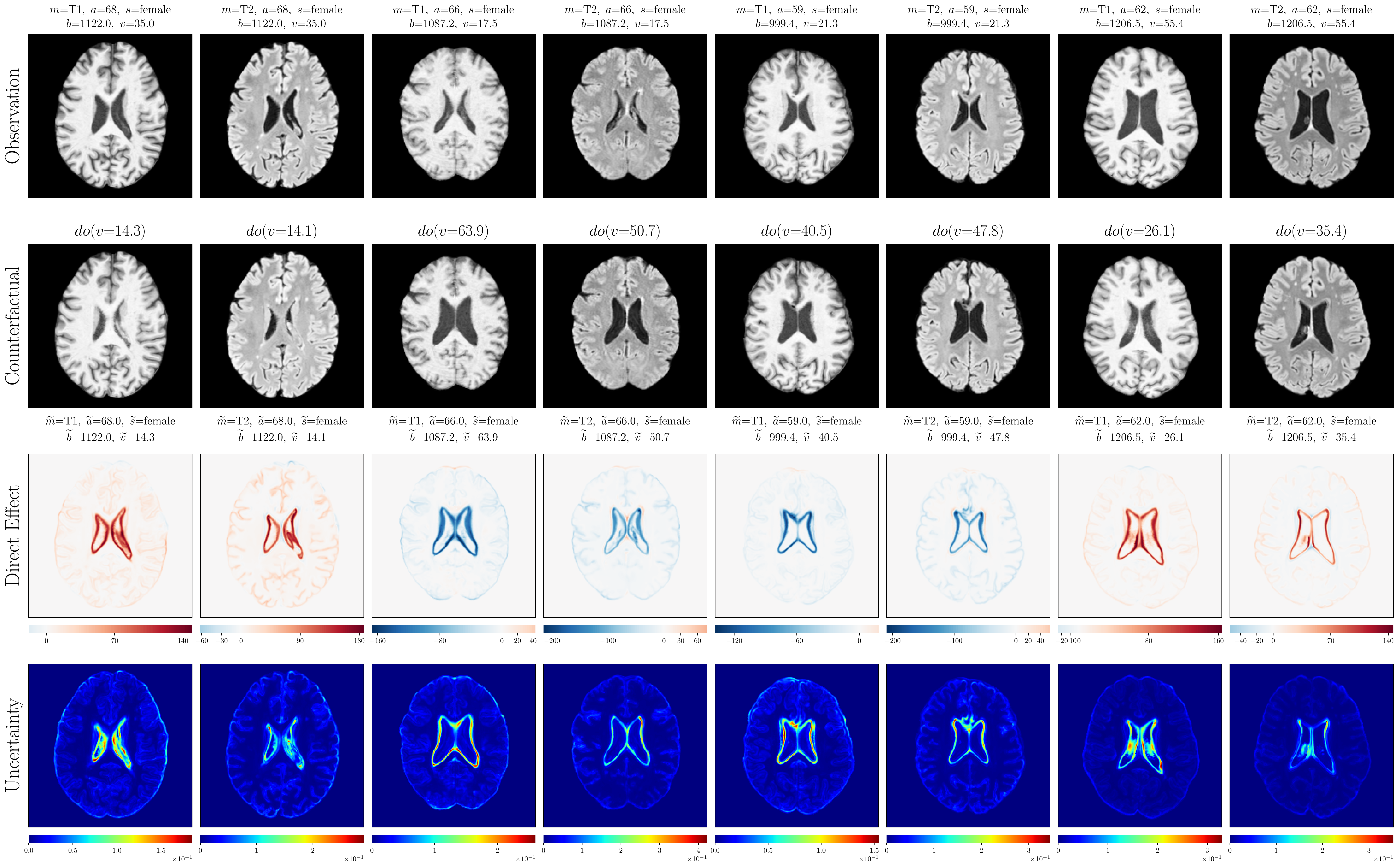}
    \\[5pt]
    \raggedright
    \ \ Post counterfactual training:
    \\[5pt]
    \includegraphics[trim={0 0 0 0}, clip, width=\textwidth]{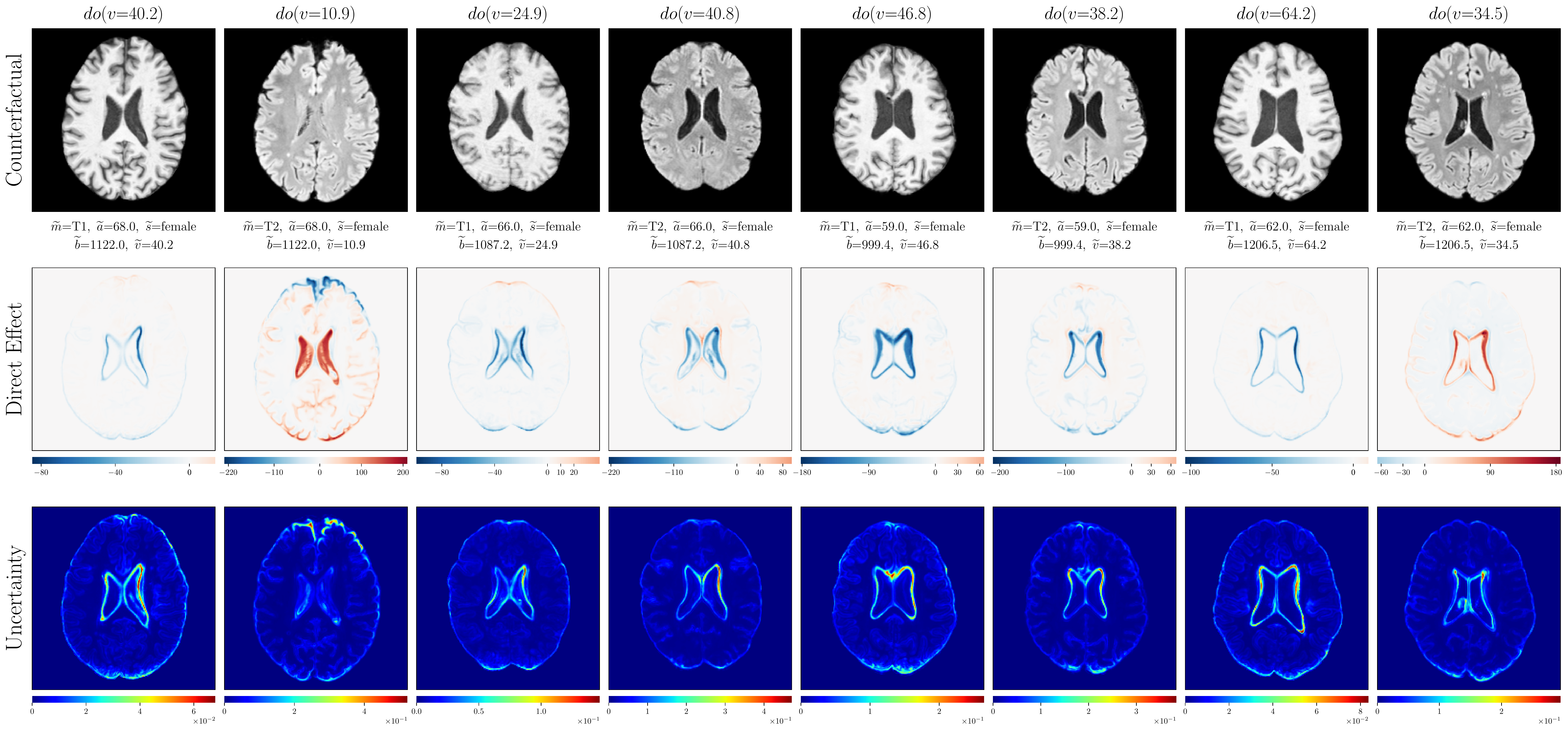}
    \caption{Visualizing ventricle volume ($v$) test set counterfactuals from our Deep SCM using the proposed exogenous prior HVAE for $\mathbf{x}$'s mechanism (non-cherrypicked). Direct treatment effects and per-pixel counterfactual uncertainty maps are also shown. Counterfactuals were obtained from random interventions on ventricle vol. ($do(v)$) before (rows 2 to 4) and after counterfactual training (rows 5 to 7). 
    }
\end{figure*}
\begin{figure}
    \centering
    {\large Random Samples from full SCM}
    \\[5pt]
    \includegraphics[width=.95\textwidth]{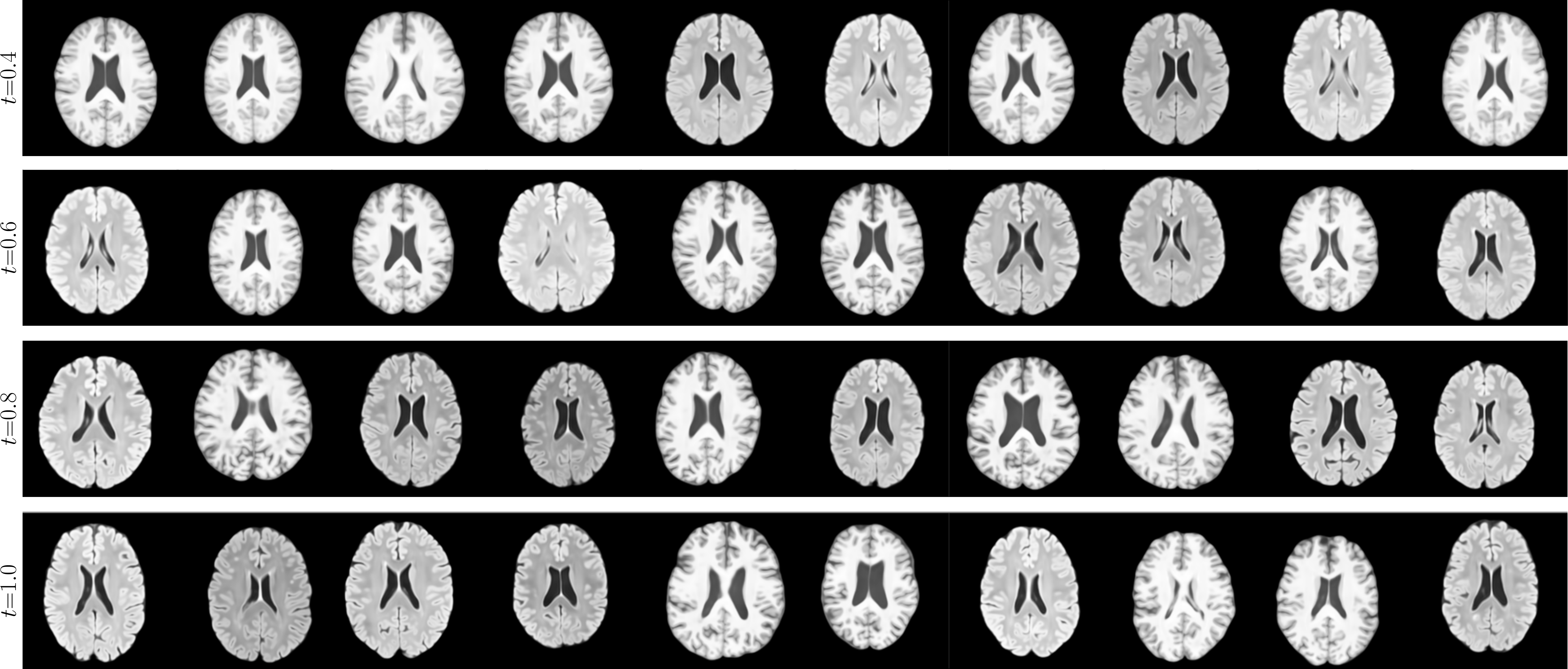}
    \caption{Random samples from our full structural causal model at different temperatures $t$ per row ($192{\times}192$ resolution, non-cherrypicked). Lower temperature samples ($t<1$) are typically used for more complex datasets. Samples above may sometimes be translated because $\mathbf{x}$'s HVAE mechanism was trained using random cropping data augmentation.}
\end{figure}
\begin{figure}
    \centering
    {\large Null-Interventions on full SCM}
    \\[5pt]
    \includegraphics[width=.95\textwidth]{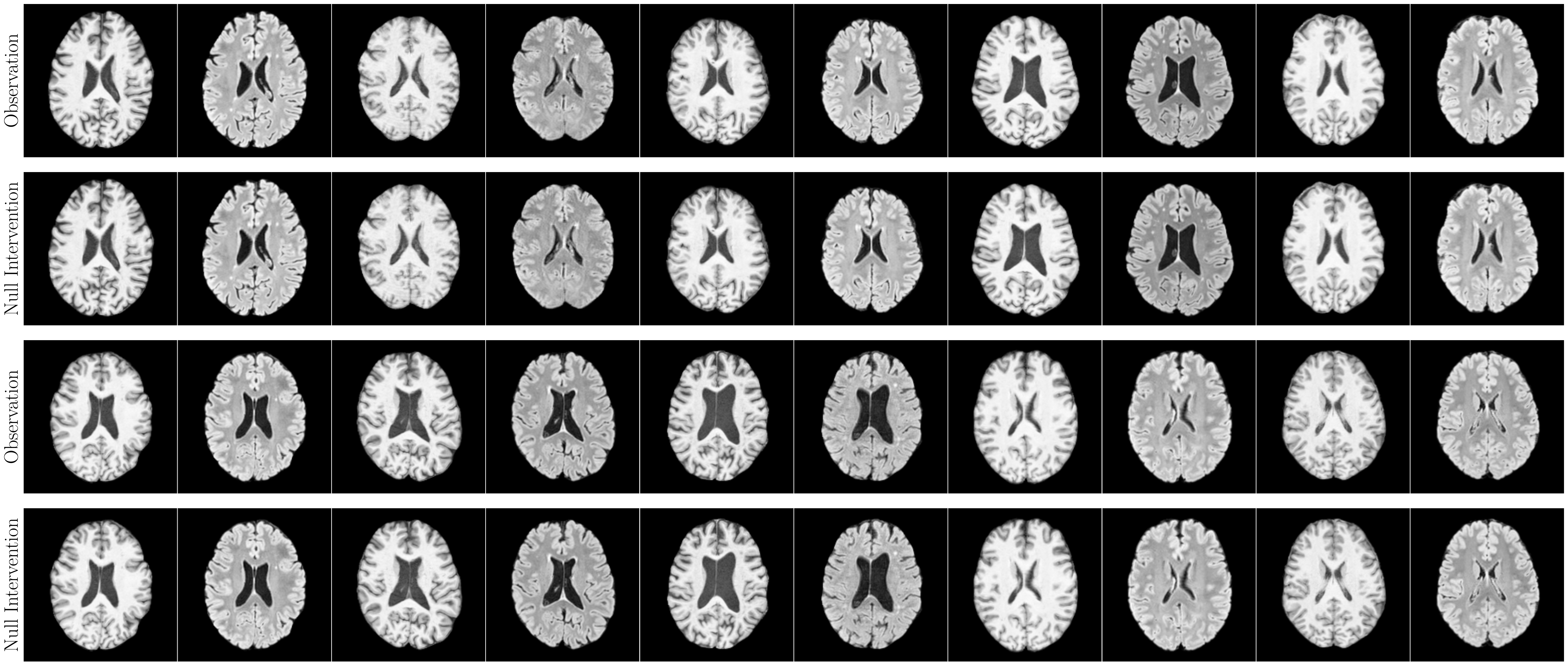}
    \caption{Null intervention results on random test set examples from UK Biobank (non-cherrypicked, using our exogenous prior HVAE as $\mathbf{x}$'s mechanism). Recall that a null intervention is simply the act of intervening on our SCM using values of the parents $\mathbf{pa}_\mathbf{x}$ of an observation $\mathbf{x}$. In effect, it is a \textit{reconstruction} of the input and it measures the composition property.}
\end{figure}
\onecolumn
\onecolumn
\newpage
\section{Chest X-ray (MIMIC-CXR)}
\label{app:chest}
\subsection{Dataset details}
We resized all the MIMIC-CXR chest X-ray images to $192{\times}192$ resolution and selected four attributes of interest from the meta-data, namely: \textit{sex}, \textit{race}, \textit{age} and \textit{disease}. The assumed causal graph is presented in Figure~\ref{fig:chest}. Notably, for disease we only considered Pleural Effusion and filtered the dataset of other diseases. Therefore, our resulting dataset only contains subjects that were either diagnosed as healthy (no finding) or with Pleural Effusion. Finally, we split the dataset into 62,336 subjects for training, 9,968 for validation and 30,535 for testing.

\subsection{Experiment setup}
\textbf{Architecture.} We used the same exogenous prior HVAE architecture as in the Brain MRI experiments (see Section~\ref{sec:brain_exp_setup}).

\textbf{Training details.} We trained our HVAEs for $\simeq$210K steps using a batch size of 32 and the AdamW optimizer~\cite{loshchilov2017decoupled}. The rest of the training settings are the same as in Section~\ref{sec:brain_exp_setup}.

\subsection{Discrete counterfactuals} 
For the MIMIC-CXR chest X-ray dataset, we assumed a causal model as shown in Figure~\ref{fig:chest}. In this causal structure, age $a$ was the parent of disease $d$ which represents the existence of Pleural Effusion. Since $d$ is not a continuous variable, normalizing flows could not be directly employed for modelling $d$'s (invertible) mechanism. To solve this, we adopted the discrete mechanisms with the Gumbel-max parametrisation as suggested in \citet{pawlowski2020deep}, Appendix C. More mathematical details can be found in \citet{Maddison2017Gumbel,oberst2019counterfactual}.

The Gumbel-max trick is a method to draw a sample for a discrete distribution, given its probabilities over categories. Suppose we have a discrete random variable $y$ over $K$ categories, with likelihood represented by logits $\bm{\lambda}$:
\begin{align}
p(y=k) = \frac{e^{\lambda{k}}}{\sum_{n=1}^{K}e^{\lambda{n}}},
\end{align}

Due to a special property of the Gumbel distribution, if we sample $\hat{y}$ by: 
\begin{align}
    \label{eq:gumbel arg max}
    &\hat{y}:={\arg\max}_{k'\leq K}(\epsilon_{k'}+\lambda_{k'}), &\epsilon_{k'} \sim \text{Gumbel}(0,1),
\end{align}
the resulting $\hat{y}$ has exactly the same distribution as $y$. Furthermore, if we were to observe $y=k$, then we can infer the values of $\bm{\epsilon}$ by sampling from the exact posterior as follows:
\begin{align}
\label{eq:gumbel infer e}
    \hat{\epsilon}_{k} &= G_{k}+\log \sum_{l=1}^{K}e^{\lambda_{l}} -\lambda_{k},  
    && G_{k} \sim \text{Gumbel}(0,1),\\ \hat{\epsilon}_{i} &= -\log (e^{-G_{i}-\lambda_{i}}+e^{-\hat{\epsilon}}_{k}-\lambda_{k})-\lambda_{i},  
    && G_{i} \sim \text{Gumbel}(0,1), \ i\neq k. 
\end{align}

We can then formulate the (approximately) invertible mechanism for a discrete attribute $y$ with parents $\mathbf{pa}_{{y}}$ by making $\bm{\lambda}$ a function of $\mathbf{pa_{{y}}}$ via a neural network $g_{y}$.
Thus, the forward mechanism $f_{y}$ to generate $y$ given its parents $\mathbf{pa}_{y}$ consists of first computing the logits $\bm{\lambda}$, then sampling via Eq.~\ref{eq:gumbel arg max}:
\begin{align}
    &&y:=f_{y}(\bm{\epsilon}, \mathbf{pa}_{y}) ={\arg\max}_{k'\leq K}(\epsilon_{k'}+\lambda_{k'}), && \bm{\lambda} =g_{y}(\mathbf{pa}_{y}),&&\epsilon_{k'} \sim \text{Gumbel}(0,1).
\end{align}
Moreover, when we perform an upstream intervention on $\widetilde{\mathbf{pa}}_{y}$ yielding: $\bm{\widetilde{\lambda}}=g(\mathbf{\widetilde{pa}}_{y})$, we can (non-deterministically) compute the counterfactual outcome $\widetilde{y}$ by first inferring $\bm{\hat{\epsilon}}$ from the exact posterior via Eq.~\ref{eq:gumbel infer e} using the original (observational) logits $\bm{\lambda}$, and then computing $\widetilde{y}$ via Eq.~\ref{eq:gumbel arg max} using $\bm{\widetilde{\lambda}}$ and the inferred $\bm{\hat{\epsilon}}$. 

\onecolumn
\subsection{Extra results}
\label{app:chest extra results}
\begin{figure}[!ht]
    \hfill
    \begin{subfigure}{\textwidth}
        \centering
        \includegraphics[width=.625\textwidth]{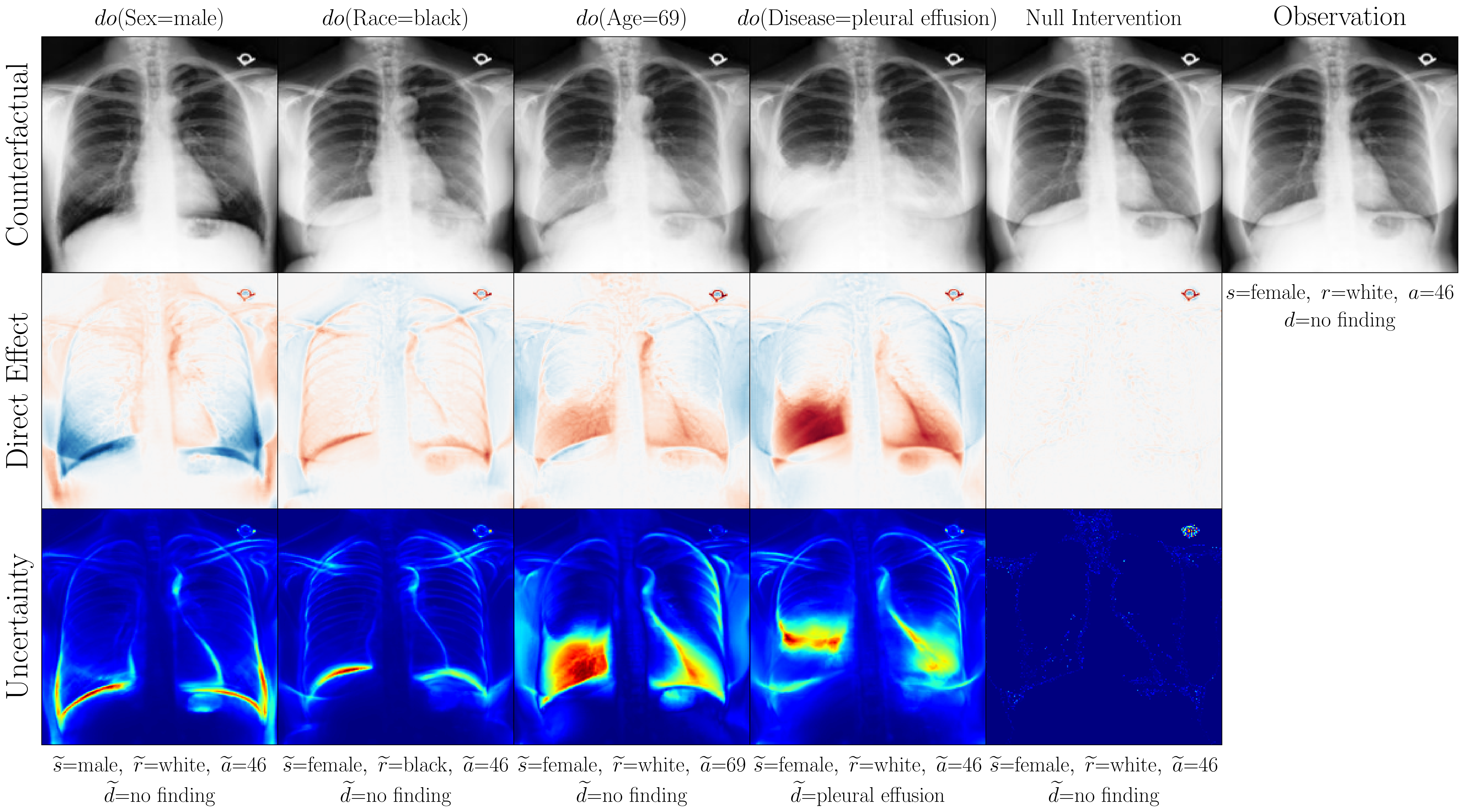}
    \end{subfigure}
    \hfill
    \\[0pt]
    \begin{subfigure}{\textwidth}
        \centering
        \includegraphics[width=.625\textwidth]{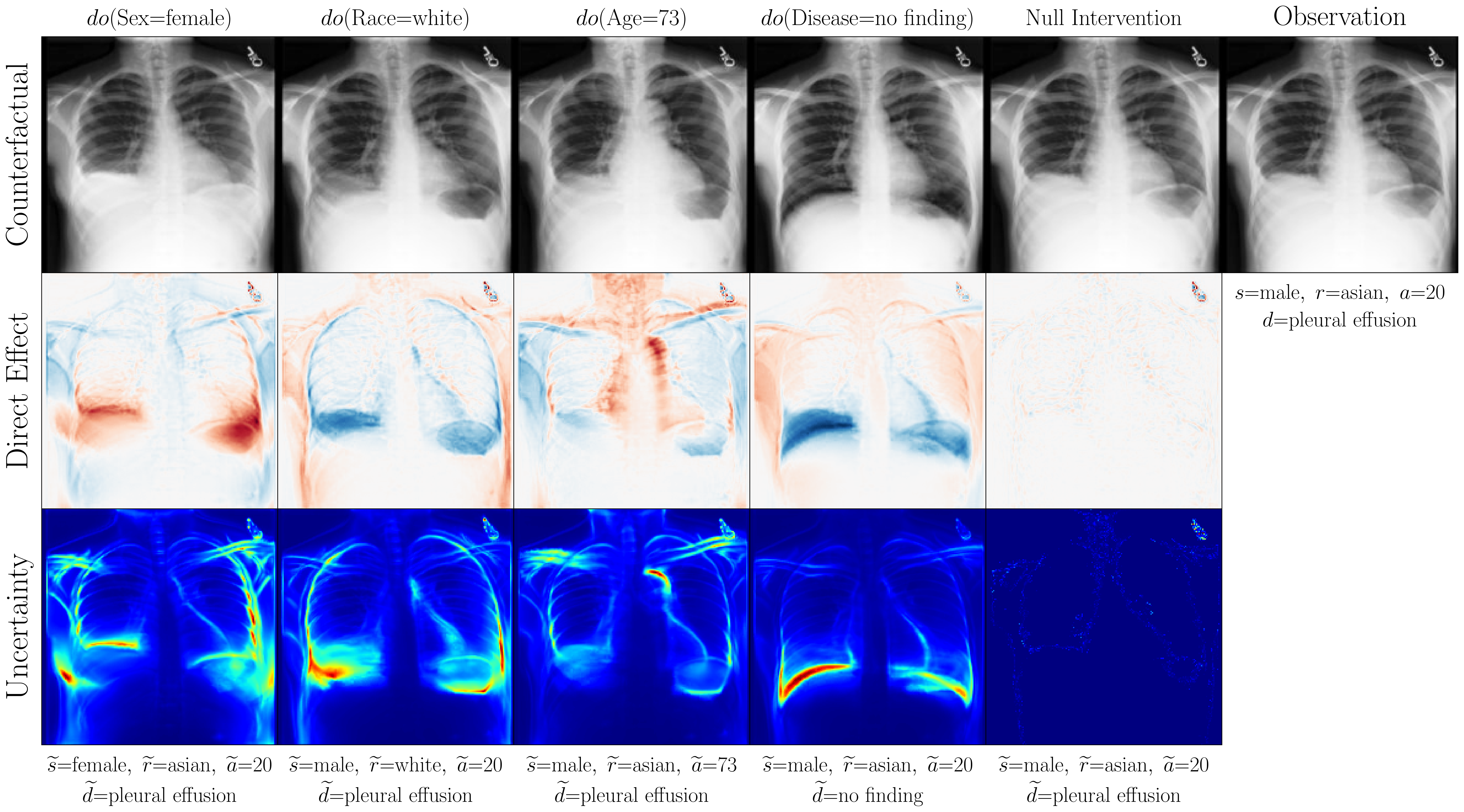}
    \end{subfigure}
    \hfill
    \\[0pt]
    \begin{subfigure}{\textwidth}
        \centering
        \includegraphics[width=.625\textwidth]{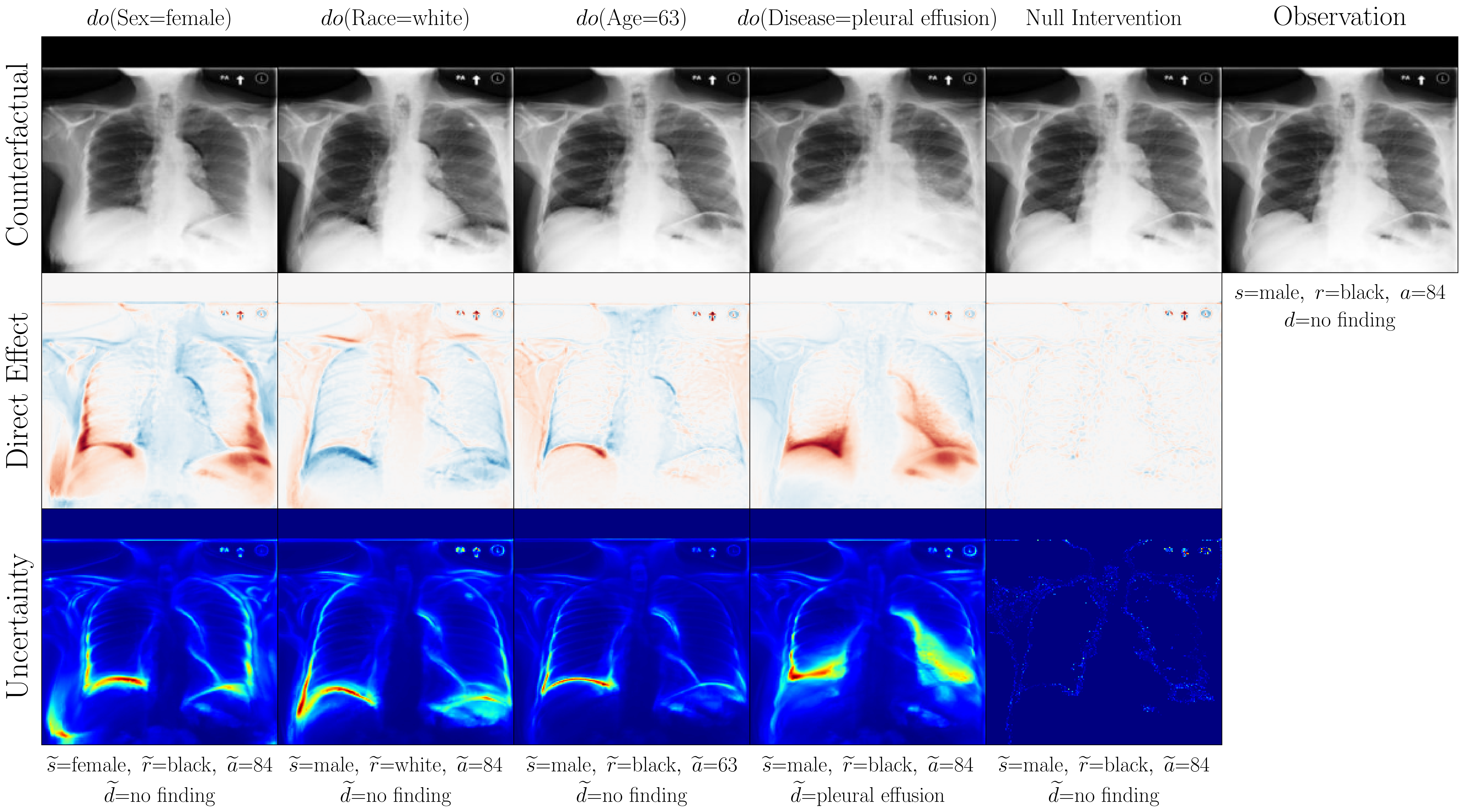}
    \end{subfigure}
    \hfill
    \caption{Additional counterfactuals from random interventions of each attribute on the chest X-ray test set. We observed localized changes in accordance with the interventions performed and the associated assumed causal graph. Crucially, the identity of the original observation is well preserved, indicating that the approximate abduction step is fairly accurate even at high resolutions. The direct causal effect is the difference between the generated counterfactual and the null-intervened (reconstructed) outcome. The per-pixel counterfactual uncertainty maps relate to the uncertainty in the abduction of the $\mathbf{x}$'s exogenous noise and were calculated as described in Appendix~\ref{appsec:distribution over causal worlds} using 32 samples from the posterior distribution.}
\end{figure}
\begin{figure*}[!ht]
    \centering
    {\large `Sex' counterfactuals $do(s)$}
    \\[5pt]
    \includegraphics[trim={0 0 0 0}, clip, width=\textwidth]{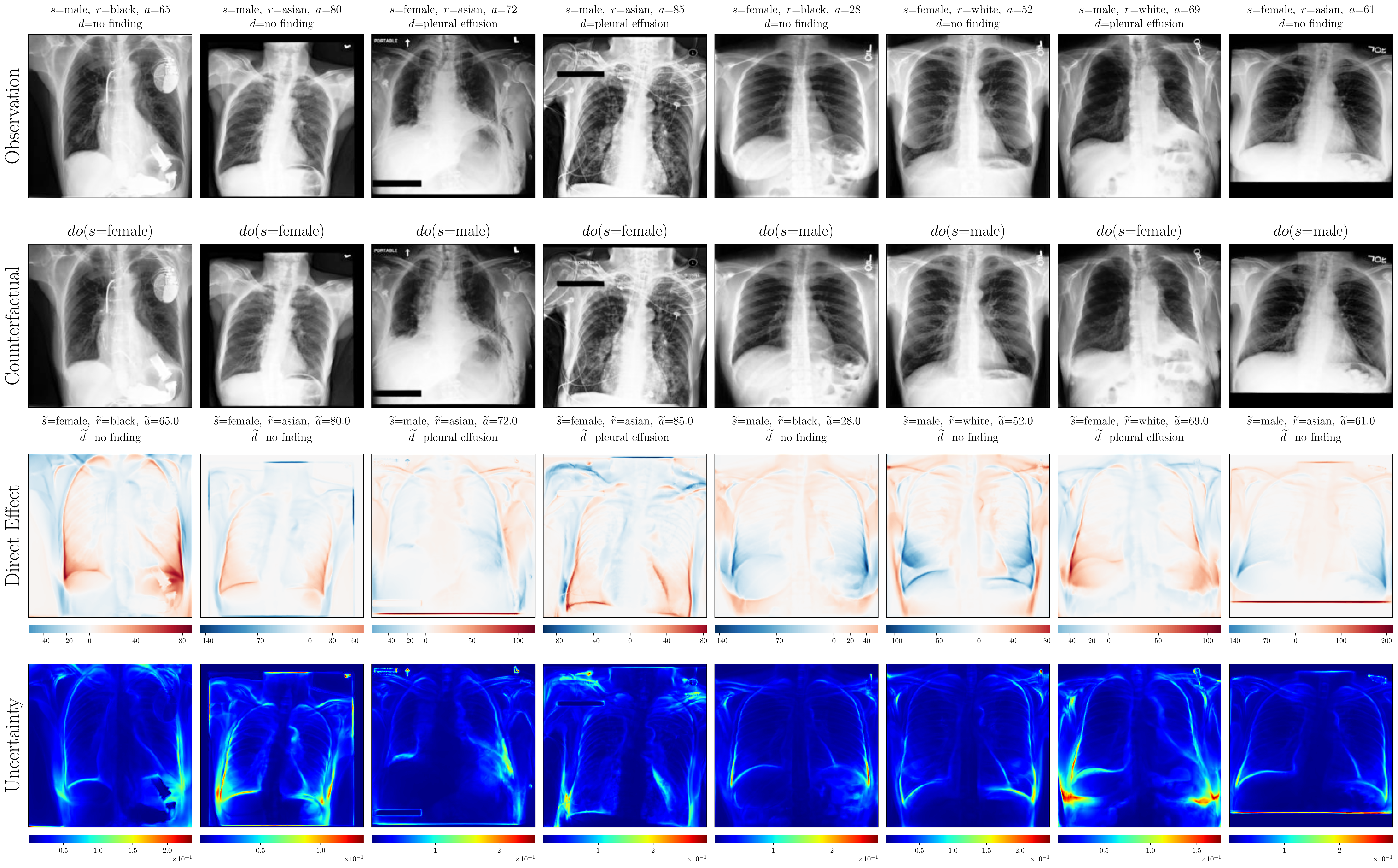}
    \\[5pt]
    \raggedright
    \ \ Post counterfactual training:
    \\[5pt]
    \includegraphics[trim={0 0 0 0}, clip, width=\textwidth]{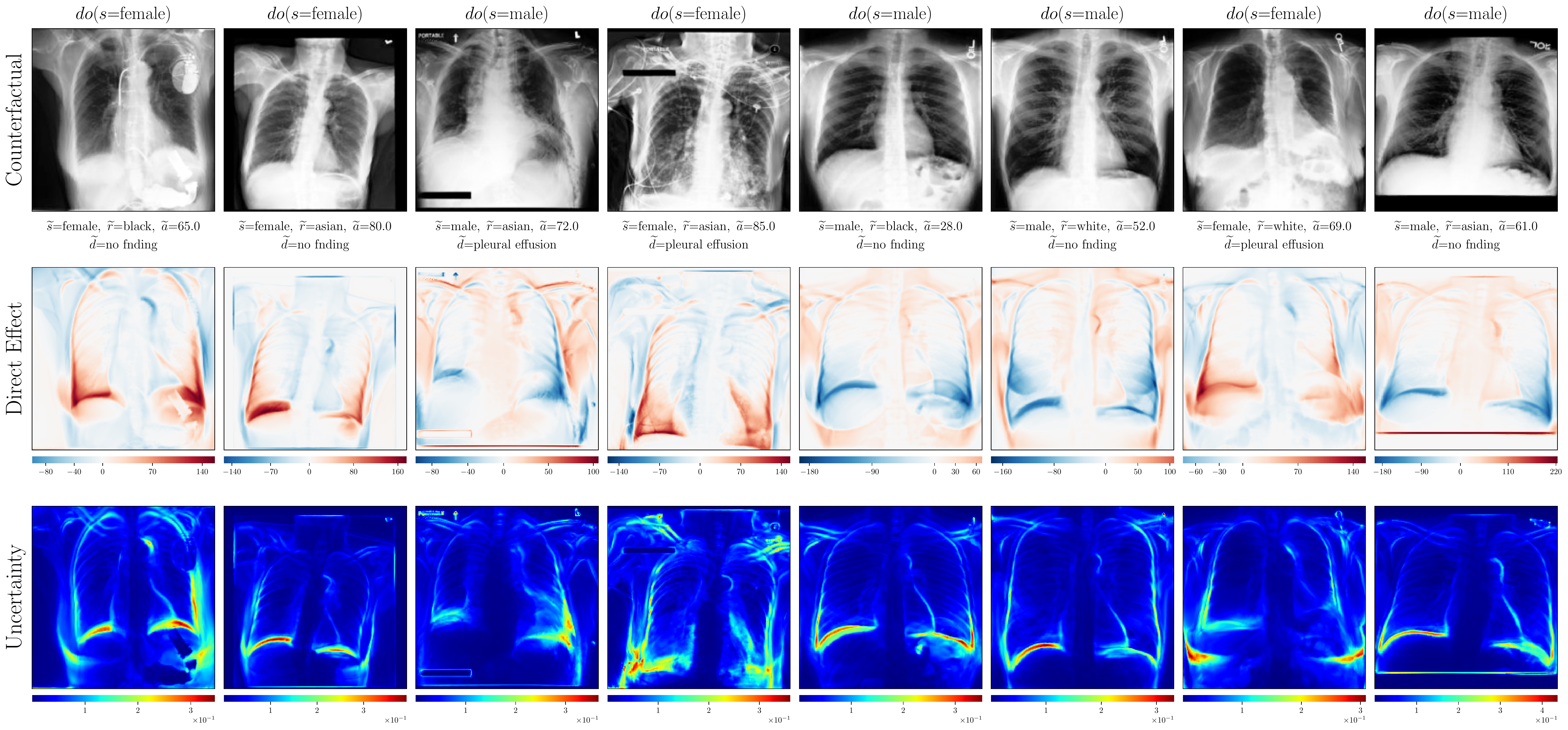}
    \caption{Visualizing chest sex ($s$) test set counterfactuals from our Deep SCM using the proposed exogenous prior HVAE for $\mathbf{x}$'s mechanism (non-cherrypicked). Direct treatment effects and per-pixel counterfactual uncertainty maps are also shown. Counterfactuals were obtained from random interventions on sex ($do(s)$) before (rows 2 to 4) and after counterfactual training (rows 5 to 7). 
    }
\end{figure*}
\begin{figure*}[!ht]
    \centering
    {\large `Race' counterfactuals $do(r)$}
    \\[5pt]
    \includegraphics[trim={0 0 0 0}, clip, width=\textwidth]{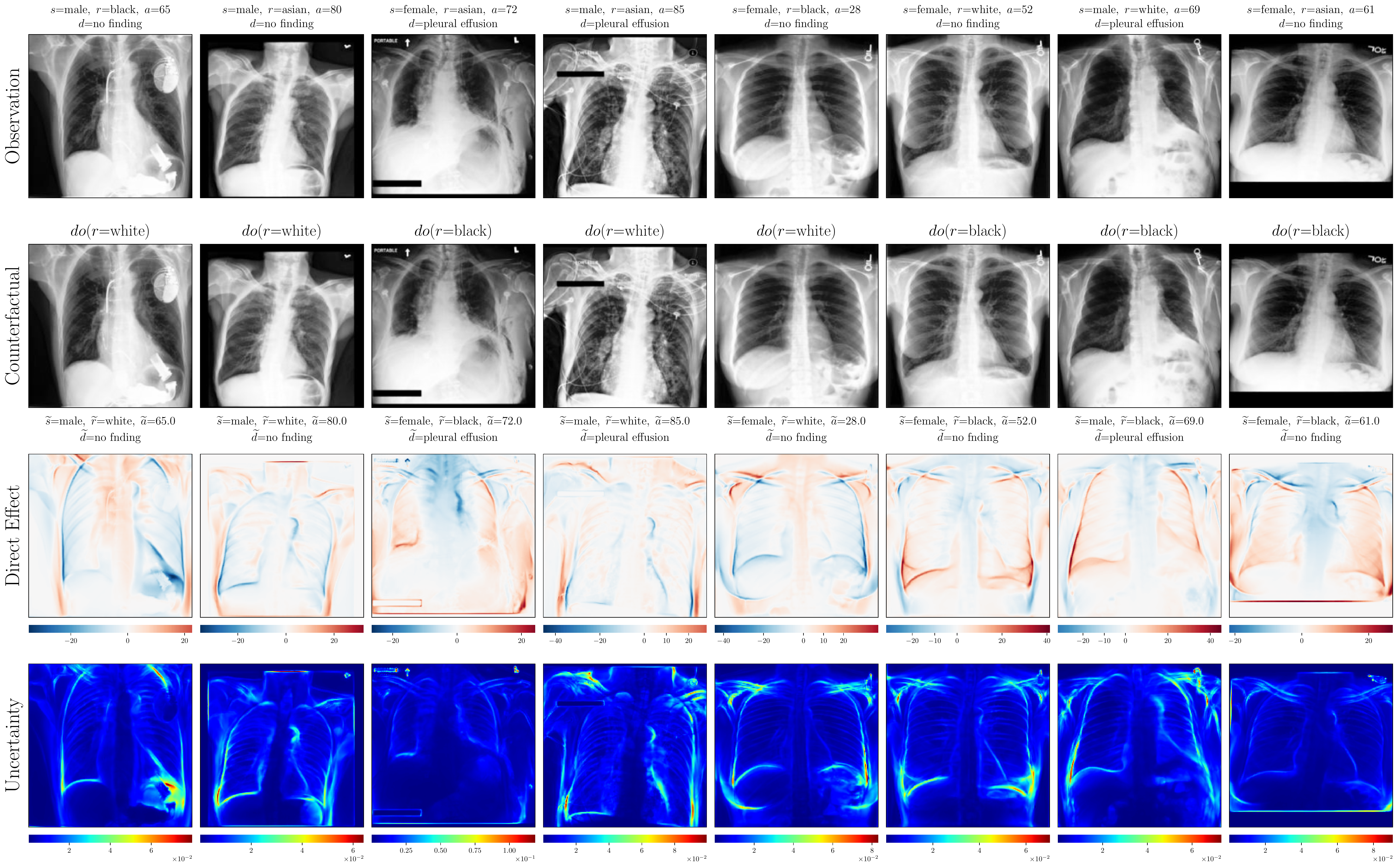}
    \\[5pt]
    \raggedright
    \ \ Post counterfactual training:
    \\[5pt]
    \includegraphics[trim={0 0 0 0}, clip, width=\textwidth]
    {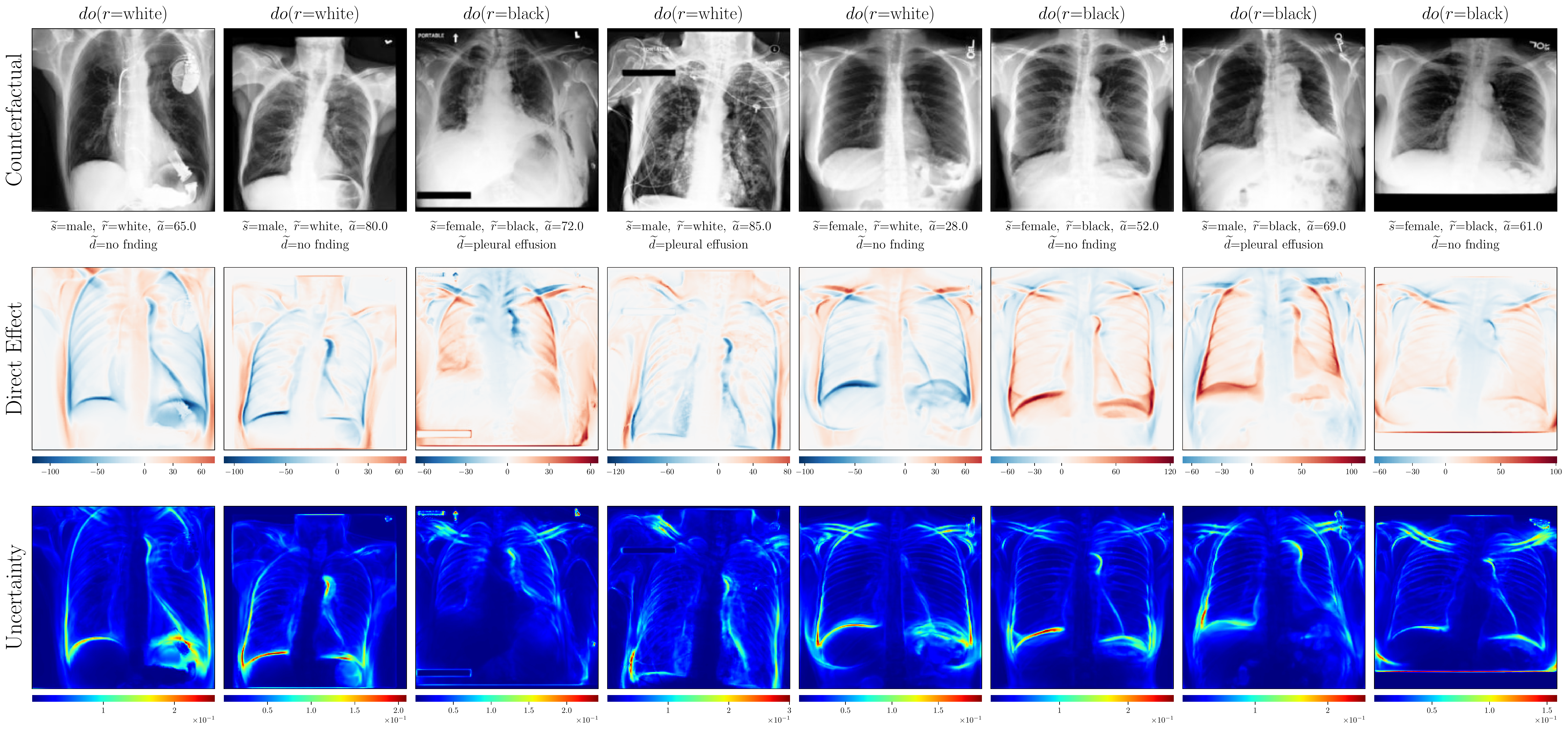}
    \caption{Visualizing chest race ($r$) test set counterfactuals from our Deep SCM using the proposed exogenous prior HVAE for $\mathbf{x}$'s mechanism (non-cherrypicked). Direct treatment effects and per-pixel counterfactual uncertainty maps are also shown. Counterfactuals were obtained from random interventions on race ($do(r)$) before (rows 2 to 4) and after counterfactual training (rows 5 to 7). 
    }
\end{figure*}
\begin{figure*}[!ht]
    \centering
    {\large `Age' counterfactuals $do(a)$}
    \\[5pt]
    \includegraphics[trim={0 0 0 0}, clip, width=\textwidth]
    {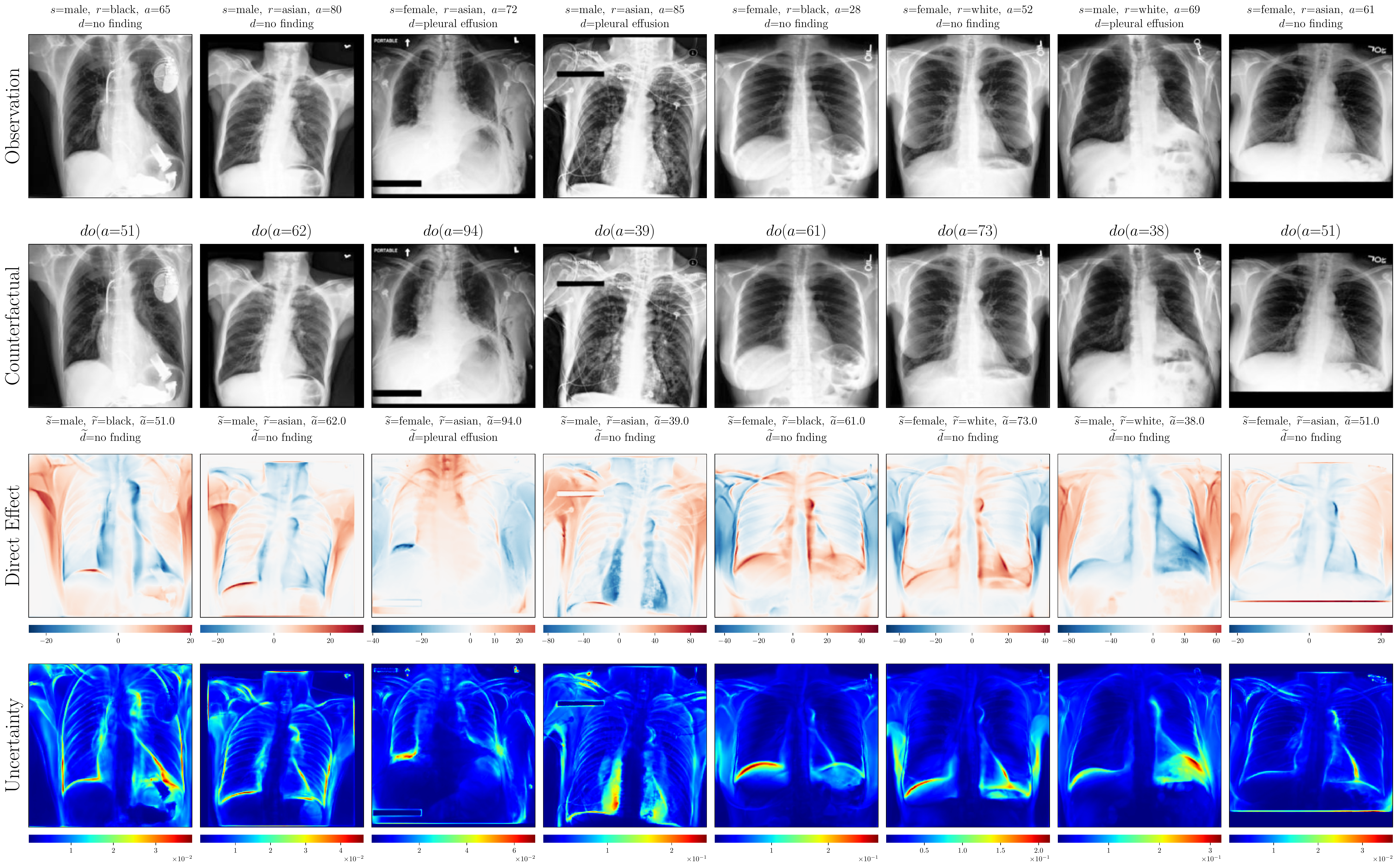}
    \\[5pt]
    \raggedright
    \ \ Post counterfactual training:
    \\[5pt]
    \includegraphics[trim={0 0 0 0}, clip, width=\textwidth]
    {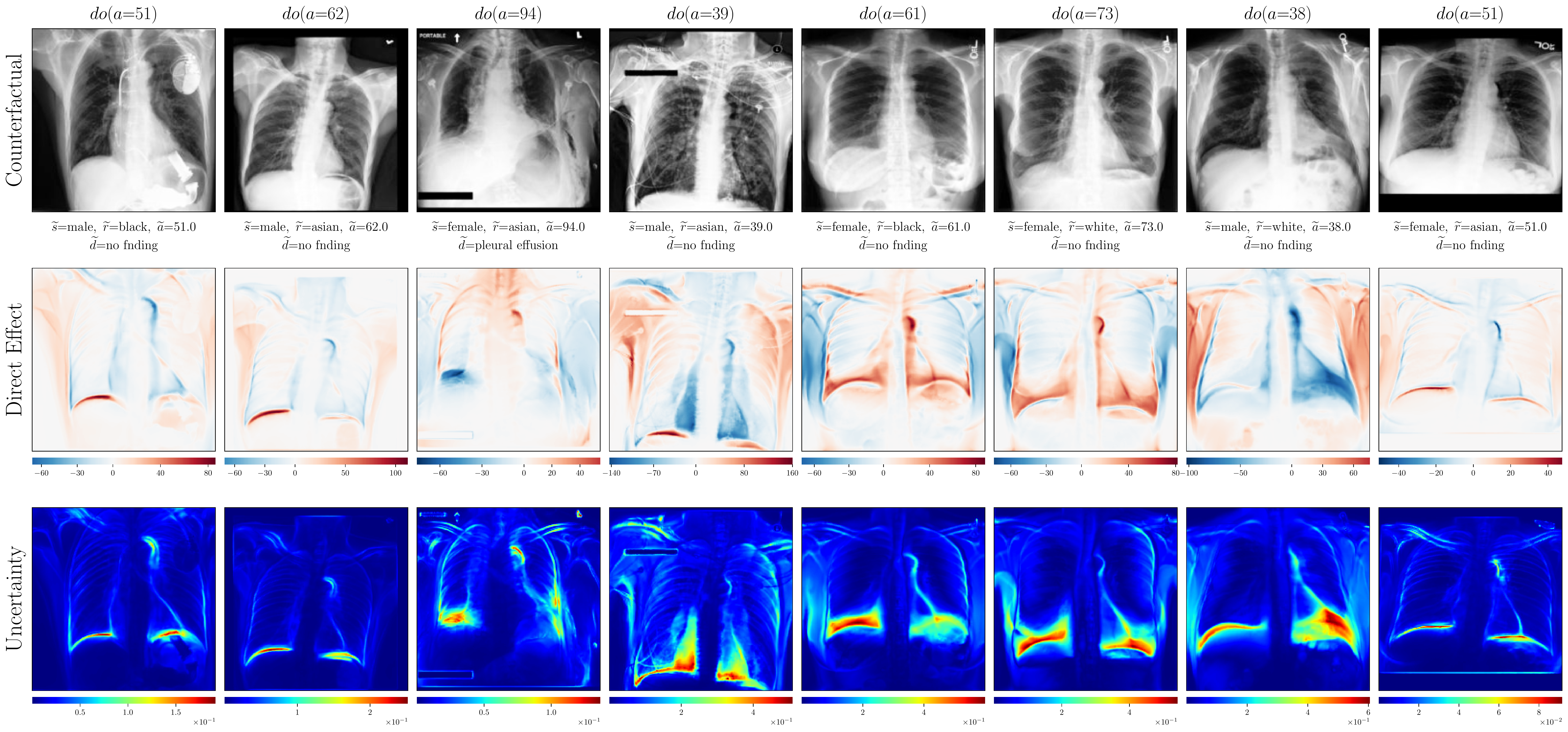}
    \caption{Visualizing chest age ($age$) test set counterfactuals from our Deep SCM using the proposed exogenous prior HVAE for $\mathbf{x}$'s mechanism (non-cherrypicked). Direct treatment effects and per-pixel counterfactual uncertainty maps are also shown. Counterfactuals were obtained from random interventions on age ($do(age)$) before (rows 2 to 4) and after counterfactual training (rows 5 to 7). 
    }
\end{figure*}
\begin{figure*}[!ht]
    \centering
    {\large `Disease' counterfactuals $do(d)$}
    \\[5pt]
    \includegraphics[trim={0 0 0 0}, clip, width=\textwidth]
    {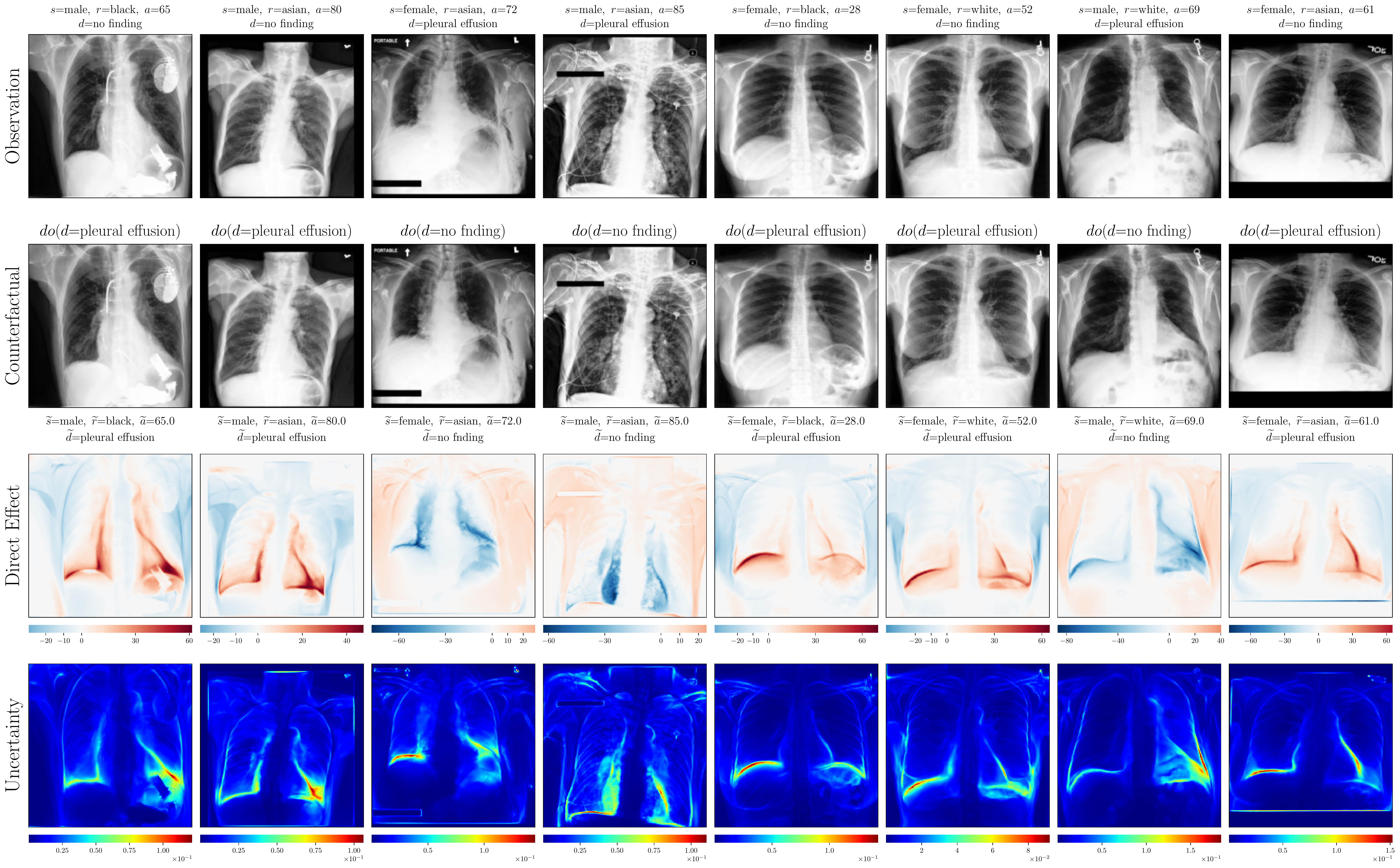}
    \\[5pt]
    \raggedright
    \ \ Post counterfactual training:
    \\[5pt]
    \includegraphics[trim={0 0 0 0}, clip, width=\textwidth]
    {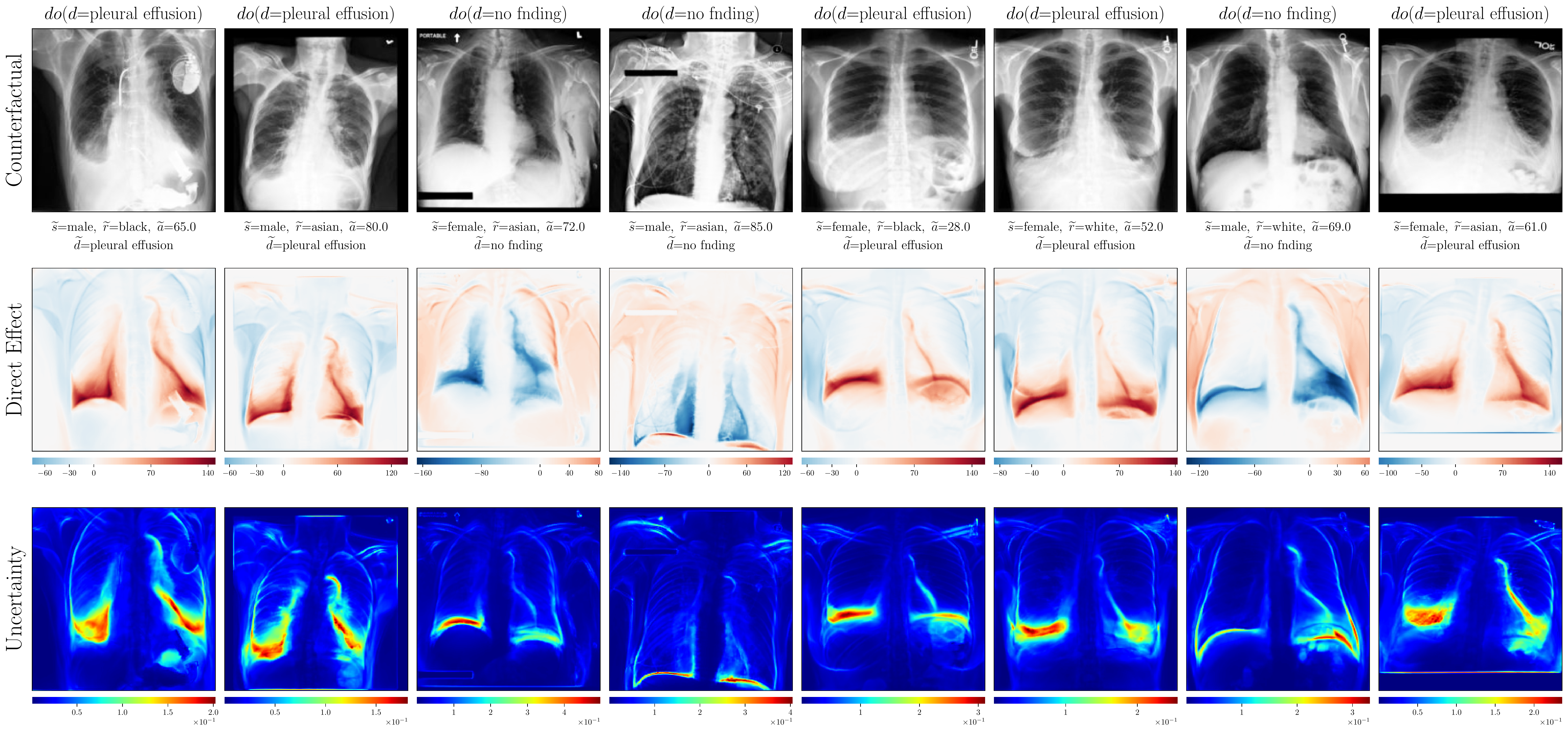}
    \caption{Visualizing chest disease ($d$) test set counterfactuals from our Deep SCM using the proposed exogenous prior HVAE for $\mathbf{x}$'s mechanism (non-cherrypicked). Direct treatment effects and per-pixel counterfactual uncertainty maps are also shown. Counterfactuals were obtained from random interventions on disease ($do(d)$) before (rows 2 to 4) and after counterfactual training (rows 5 to 7). 
    }
\end{figure*}
\begin{figure}
    \centering
    {\large Null-Interventions on full SCM}
    \\[5pt]
    \includegraphics[width=.95\textwidth]{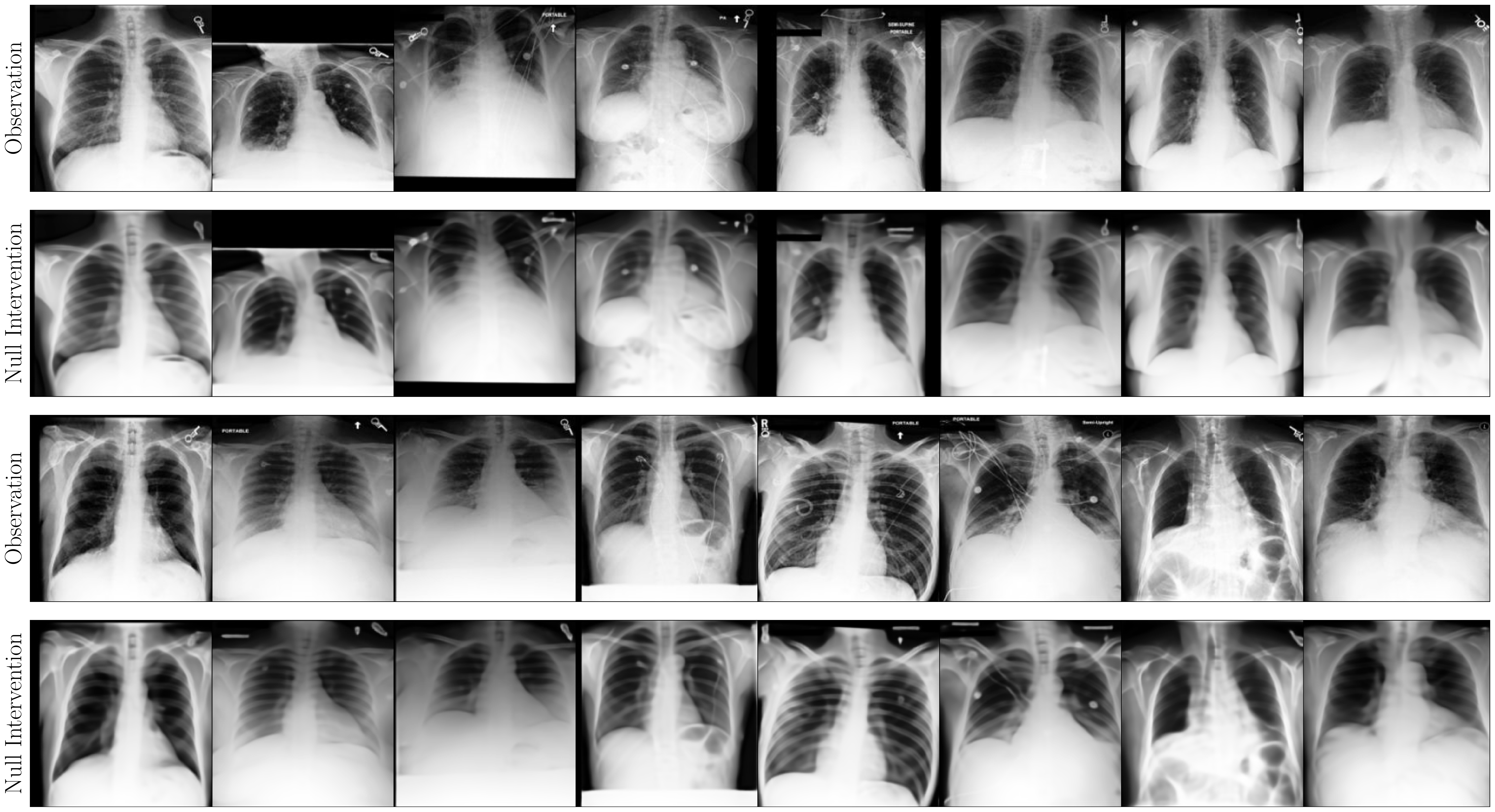}
    \caption{Null intervention results on random test set examples from MIMIC-CXR (non-cherrypicked, using our exogenous prior HVAE as $\mathbf{x}$'s mechanism). Recall that a null intervention is simply the act of intervening on our SCM using values of the parents $\mathbf{pa}_\mathbf{x}$ of an observation $\mathbf{x}$. In effect, it is a \textit{reconstruction} of the input and it measures the composition property.}
\end{figure}
\onecolumn
\section{Anticausal Predictors $q_{\psi}(k \mid \mathbf{pa}_k)$}
\label{app:Anticausal Predictors}
\begin{figure*}[!ht]
    \centering
    \begin{minted}[fontsize=\footnotesize]{python}
        class MLP(nn.Module):
            def __init__(self, num_inputs=1, width=32, num_outputs=1):
                super().__init__()
                activation = nn.LeakyReLU()
                self.mlp = nn.Sequential(
                    nn.Linear(num_inputs, width, bias=False),
                    nn.BatchNorm1d(width),
                    activation,
                    nn.Linear(width, width, bias=False),
                    nn.BatchNorm1d(width),
                    activation,
                    nn.Linear(width, num_outputs),
                )
        
            def forward(self, x):
                return self.mlp(x)
    \end{minted}
    \centering
    \begin{minted}[fontsize=\footnotesize]{python}
        class CNN(nn.Module):
            def __init__(self, in_shape, width=16, num_outputs=1, context_dim=0):
                super().__init__()
                in_channels = in_shape[0]
                res = in_shape[1]
                s = 2 if res > 64 else 1
                activation = nn.LeakyReLU()
                self.cnn = nn.Sequential(
                    nn.Conv2d(in_channels, width, 7, s, 3, bias=False),
                    nn.BatchNorm2d(width),
                    activation,
                    (nn.MaxPool2d(2, 2) if res > 32 else nn.Identity()),
                    nn.Conv2d(width, 2*width, 3, 2, 1, bias=False),
                    nn.BatchNorm2d(2*width),
                    activation,
                    nn.Conv2d(2*width, 2*width, 3, 1, 1, bias=False),
                    nn.BatchNorm2d(2*width),
                    activation,
                    nn.Conv2d(2*width, 4*width, 3, 2, 1, bias=False),
                    nn.BatchNorm2d(4*width),
                    activation,
                    nn.Conv2d(4*width, 4*width, 3, 1, 1, bias=False),
                    nn.BatchNorm2d(4*width),
                    activation,
                    nn.Conv2d(4*width, 8*width, 3, 2, 1, bias=False),
                    nn.BatchNorm2d(8*width),
                    activation
                )
                self.fc = nn.Sequential(
                    nn.Linear(8*width + context_dim, 8*width, bias=False),
                    nn.BatchNorm1d(8*width),
                    activation,
                    nn.Linear(8*width, num_outputs)
                )
        
            def forward(self, x, y=None):
                x = self.cnn(x)
                x = x.mean(dim=(-2, -1))  # avg pool
                if y is not None:
                    x = torch.cat([x, y], dim=-1)
                return self.fc(x)
    \end{minted}
\end{figure*}
\newpage
\paragraph{Architecture.} The parent predictors (classifiers/regressors) shown in the PyTorch code above were used for both Morpho-MNIST and the Brain MRI dataset and were trained using Pyro. For MIMIC-CXR dataset, we adopted the standard ResNet-18~\cite{he2016deep} architecture pre-defined in \textit{Torchvision}~\cite{paszke2019pytorch} for the parent predictors. These predictors play two roles in our approach: (i) to serve as pseudo-oracles in evaluating the effectiveness of generated counterfactuals; (ii) to provide guidance during our proposed counterfactual training technique. For both purposes, our parent predictors are trained on observational data and in the \textit{anticausal} direction with respect to each variable in the assumed SCM. That is, each variable is predicted from its children. When a variable in the SCM is not a direct parent of the image $\mathbf{x}$, we use the MLP architecture for its predictor, otherwise, we use the CNN. As a side note, missing values for some of the parents in our observed data can restrict the applicability of SCMs. In order to handle missing values, we can use variational predictors to infer parent attributes in the anticausal direction. That is, when a certain parent is not present in an observed datum, we can infer it given its observed children (imputation). The inferred parent may then be used downstream as if it was observed to, e.g. compute approximate counterfactuals.

\begin{table*}[!t]
    \small 
    \centering
    \caption{Attribute predictor performance on UK Biobank observational data.}
    \label{tab:classifiers}
    \begin{tabular}{ccccc}
        \toprule
         \textsc{Sex} $(s)$ & \textsc{MRI} $(m)$ & \textsc{Age} $(a)$ & \textsc{Brain Vol.} $(b)$ & \textsc{Ventricle Vol.} $(v)$ \\[2pt]
         ROCAUC $\uparrow$ & ROCAUC $\uparrow$ & MAE (years) $\downarrow$ & MAE (ml) $\downarrow$ & MAE (ml) $\downarrow$ \\
        \midrule
         0.9764 $\pm$ 2e-3 & 1.0 & 4.847 $\pm$ 7e-4 & 26.77 $\pm$ 0.39 & 1.958 $\pm$ 3e-2 \\
        \bottomrule
    \end{tabular}
\end{table*}

\begin{table*}[!t]
    \small 
    \centering
    \caption{Attribute predictor performance on MIMIC-CXR observational data.}
    \label{tab:chest_classifiers}
    \begin{tabular}{ccccc}
        \toprule
         \textsc{Sex} $(s)$ & \textsc{Race} $(m)$ & \textsc{Age} $(a)$ & \textsc{Disease} $(d)$ \\[2pt]
         ROCAUC $\uparrow$ & ROCAUC $\uparrow$ & MAE (years) $\downarrow$ & ROCAUC $\uparrow$  \\
        \midrule 0.9950
        & 0.7496
        & 6.219
        & 0.9419
         \\
        \bottomrule
    \end{tabular}
\end{table*}

\paragraph{Predictor Training Details.} For each dataset, we train the predictors for all parents simultaneously until convergence, where the total loss is simply the sum of all the individual predictor losses. We use a batch size of 32, and use the AdamW optimizer with a learning rate of 1e-4 and weight decay of 0.1 for UK Biobank, 0.01 for Morpho-MNIST and 0.05 for MIMIC-CXR. The final artefacts are an exponential moving average of the predictor's parameters with a rate of 0.999, which we use at inference time. For data augmentation, we random crop with an all-border zero-padding of 4 for Morpho-MNIST 9 for UK Biobank and MIMIC-CXR. We further perform random horizontal flips with probability 0.5 for UK Biobank. Pixel intensities were rescaled to $[-1,1]$ for all datasets.

\paragraph{Counterfactual Training Details.} As described in the main text, once all the mechanisms in the SCM of interest have been trained, they are combined into one PyTorch module which we use for counterfactual training/inference. For the Brain MRI and Chext X-ray datasets, we found it most effective to freeze all other mechanisms (including parent predictors) in the SCM during constrained counterfactual fine-tuning of $\mathbf{x}$'s HVAE mechanism -- since we care mostly about the quality/faithfulness of $\widetilde{\mathbf{x}}$ counterfactuals. For counterfactual fine-tuning, we start with the exponential moving average of the pre-trained HVAE parameters, and use the AdamW optimizer with a learning rate 10${\times}$ smaller than what was used to originally train the HVAE mechanism, 1e-4. All else remain the same as the original HVAE training setup except for the batch size which was decreased to 24 due to GPU memory constraints. To train the Lagrange multiplier parameter $\lambda$, we used a separate AdamW optimizer with a learning rate of 0.1, and initialised $\lambda$ at 10. We optimize the Lagrangian in Eq.~\eqref{eq:lagrangian} by performing stochastic gradient descent on the HVAE's parameters and \textit{ascent} on the Lagrange multiplier. We observed that training for around 10K iterations was sufficient to achieve the desired effect on the brain MRI data, and training for 6.5K iterations led to desirable performance on the chest X-ray data. Note that the efficacy of this strategy is entirely dependent on the quality of the initial parent predictors used for counterfactual training. Moreover, in order to help identify whether the model learned shortcuts in producing \textit{effective} interventions, we ensured that we used separately trained anticausal predictors for counterfactual training and for the subsequent counterfactual \textit{effectiveness} evaluation. As is the case with all types of classifier-based guidance, it is important to keep the possibility of shortcut learning in mind when using counterfactual training, and to take precautions to mitigate it. Using more robust classifiers with smoother loss surfaces (e.g. Bayesian predictors, adversarial training techniques etc) may prove particularly helpful.

\end{document}